\pdfoutput=1

\documentclass[11pt]{article}

\usepackage[preprint]{acl}

\usepackage{times}
\usepackage{latexsym}

\usepackage[T1]{fontenc}

\usepackage[utf8]{inputenc}

\usepackage{microtype}
\usepackage{amsmath} 
\usepackage{inconsolata}
\usepackage{longtable}             
\usepackage{graphicx}
\usepackage{url}            
\usepackage{booktabs}       
  
\usepackage{amsfonts}       
\usepackage{nicefrac}       
\usepackage{color,xcolor,xspace}         
\usepackage{bigstrut}
\definecolor{green}{RGB}{47,110,200} 
\usepackage{titlesec}
\usepackage{textcomp}
\usepackage{colortbl}
\usepackage{graphics}
\usepackage{multirow}
\usepackage{pdfpages}
\usepackage{float} 
\usepackage{bm}
\usepackage{subfigure}
\usepackage{enumitem}
\usepackage{wrapfig}
\usepackage{bbm}
\usepackage{tcolorbox}
\usepackage{makecell}
\title{Explaining  Length Bias in LLM-Based Preference Evaluations}
\def \OURS {AdapAlpaca}

\newtheorem{assumption}{Assumption}
\newtheorem{definition}{Definition}

\tcolorboxenvironment{datageneration}{
  colback=rgb{0.7765,0.9373,0.8078}
  boxrule=1pt, 
  boxsep=1pt,
  left=2pt,right=2pt,top=2pt,bottom=2pt,
  oversize=2pt,
  sharp corners,
  before skip=\topsep,
  after skip=\topsep,
}

\tcbset{
  mybox/.style={
    colback=blue!5,
    boxrule=1pt, 
    boxsep=1pt,
    left=10pt, 
    right=10pt, 
    top=10pt, 
    bottom=10pt, 
    oversize=2pt,
    arc=4pt, 
    before skip=10pt, 
    after skip=10pt, 
    baseline=3.5ex 
  }
}

\author{
Zhengyu Hu$^{1}$,
~~Linxin Song$^{3}$,
~~Jieyu Zhang$^4$,
~~Zheyuan Xiao$^{1,7}$,
~~\textbf{Tianfu Wang}$^{1}$,\\
~~\textbf{Zhenyu Chen}$^{5}$,
~~\textbf{Jianxun Lian}$^{6}$,
~~\textbf{Nicholas Jing Yuan}$^{6}$,
~~\textbf{Kaize Ding}$^{8}$,
~~\textbf{Hui Xiong}$^{1,2}$\\
$^1$ Thrust of Artificial Intelligence, HKUST (Guangzhou), China 
\\
$^2$ Department of Computer Science and Engineering, HKUST,  Hong Kong SAR, China
\\
$^3$ University of Southern California  \quad $^4$ University of Washington 
\quad $^5$ MeiTuan   \\
\quad $^6$ Microsoft 
\quad $^7$ Resideo 
\quad $^8$ Northwestern University
}

\begin{document}
\maketitle

\begin{abstract}

The use of large language models (LLMs) as judges, particularly in preference comparisons
has become widespread, but this reveals a notable bias towards longer responses, undermining the reliability of such evaluations.
To better understand such bias, we propose to decompose the preference evaluation metric, specifically the \emph{win rate}, into two key components: 
\emph{desirability} and \emph{information mass}, where the former is length-independent and related to trustworthiness such as correctness, toxicity, and consistency, and the latter is length-dependent 
and  represents the amount of information
in the response.
We empirically demonstrated the decomposition through controlled experiments and found that response length impacts evaluations by influencing information mass. 
To derive a reliable evaluation metric that assesses content quality without being confounded by response length, we propose \OURS{}, a simple yet effective adjustment to win rate measurement.
Specifically, \OURS{} ensures a fair comparison of response quality by aligning the lengths of reference and test model responses under equivalent length intervals.

\end{abstract}

\section{Introduction}
\label{sec:intro}

As LLMs are increasingly deployed across various domains of artificial intelligence, from natural language processing to complex decision-making systems~\citep{ song2024adaptive, song2025coact, zhang2024offline,hu2025unveiling,liuyue_GuardReasoner}, ensuring their performance, reliability, and fairness has become a critical challenge~\citep{hu2024let,liuyue_GuardReasoner-VL,wang2023pandalm, song2023nlpbench, wang2024template, li2025treble, song2025discovering}.
LLM-based auto-evaluators have emerged as a crucial tool in this context, offering a cost-effective and scalable alternative to labor-intensive human evaluations~\citep{li2024generation,li2024llms,dubois2024alpacafarm, song2024better, ijcai-2025-coder-agent}. 
Despite their advantages, these automated systems are not without their shortcomings, particularly concerning the introduction and perpetuation of biases~\citep{wang2023large,ma2025judgingmindsperspectivesmean,www-2025-gen-mentor, song2025hallucination}. One of the important biases observed in LLM-based evaluations is the preference for longer textual responses. Previous empirical studies have explored a strong correlation between the length of response and its perceived quality represented by win rate~\citep{zhao2024long,dubois2024length,park2024disentangling}.
However, it is not reasonable to simply attribute the preference to length since length is only the surface factor for the quality of a sentence.
Therefore, in this work, we investigate the following question: 
\textit{what are the major factors contributing to the win rate?}

To solve this problem, we propose a new framework that decomposes the \emph{quality} of a response, as measured by its win rate in pairwise comparisons, into two distinct components:
(1) \textit{desirability}, which is independent of length and reflects the trustworthiness of the response, encompassing factors such as correctness, toxicity, and consistency; and 
(2) \textit{information mass}, which is dependent on length and represents the amount of information in the response, measurable through conditional entropy.
We validate our hypothesis by testing win rates in two different scenarios: (i) comparing normal responses with those differing in desirability (e.g., Logical to be desired and Biased not desired), and 
(ii) comparing normal responses with concise and detailed responses, which vary in information mass. 
Our experiments demonstrate that responses with negative desirability significantly decrease the win rate, whereas information mass, when not negatively influenced by desirability, is positively correlated with the win rate, thus confirming the effectiveness of our 
metric.
Following this finding, we design a new prompt called "Quality Enhancement" to improve information mass with positive desirability. 
This prompt enables GPT-4 to achieve state-of-the-art results on AlpacaEval, increasing the win rate from 50.00\% to 70.16\%.

\begin{figure}[t] 

\small
\includegraphics[width=0.45\textwidth]{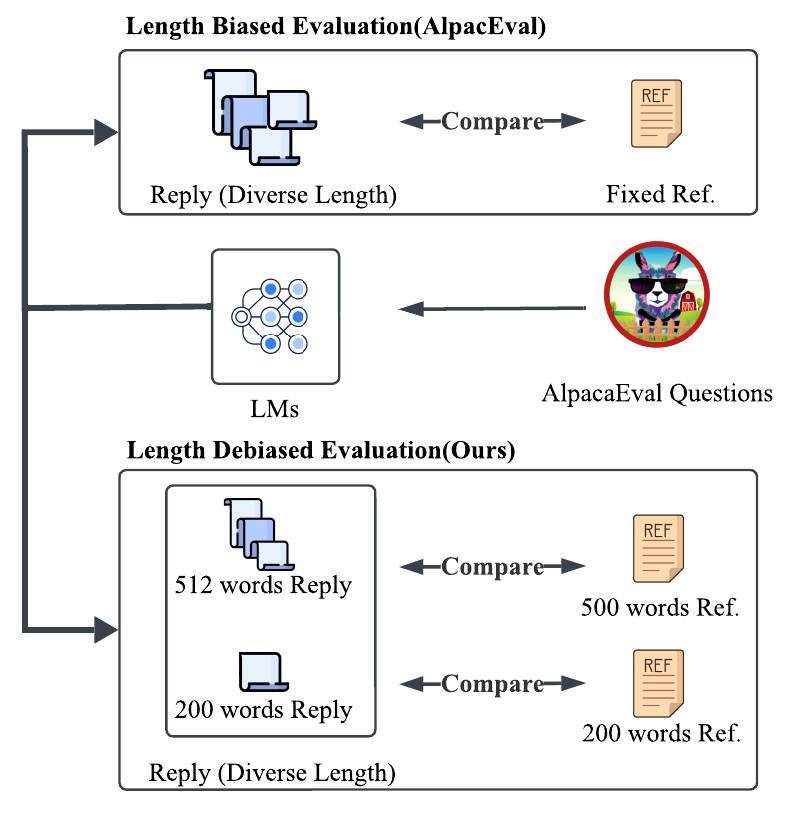}
\caption{Comparison between AlpacaEval and \OURS{} (Ours). In AlpacaEval, the reference answer has a fixed length, regardless of the length of the test model's answer. In contrast, \OURS{} dynamically selects a reference answer that matches the length of the test model's answer.
}
\label{fig:AdapAlpaca}
\end{figure}

Through our decomposition of the quality of a response, we observe that response length impacts evaluations primarily by influencing information mass. 
However, a reliable evaluation
metric should assess content quality without being confounded by extraneous factors such as response length~\citep{koo2023benchmarking,ye2024justice, dubois2024length}, we further propose \OURS{}, a benchmark designed to improve evaluation fairness. 
By ensuring that responses are compared at the same length intervals, \OURS{} effectively mitigates length bias, enabling accurate content quality assessments
(see Figure~\ref{fig:AdapAlpaca}). 
With \OURS{}, we further analyze length bias in Direct Preference Optimization (DPO)~\cite{Rafailov2023DirectPO}  to examine the findings in prior work~\citep{gu2024minillm,ivison2023camels,liu2024length} that DPO lengthens model responses. 
Specifically, we test TÜLU2~\citep{ivison2023camels} and TÜLU2-dpo models at 7B, 13B, and 70B scales on AlpacaEval and \OURS{}. 
Our results indicate that DPO leads to higher human preference, but this gain is amplified by response length, with AlpacaEval showing higher win rates gain than \OURS{}.
Our major findings and contributions are as follows:

\begin{itemize}[leftmargin=0.4cm]

    \item We propose a novel interpretation of win rate, emphasizing desirability and information mass, offering a more precise LLM performance measure.  
    Based on this interpretation, we develop the "Quality Enhancement" prompt, which improves win rates by boosting information mass with positive desirability.
    This prompt improves win rates across multiple LLMs, with average increases of 23.44\% for GPT-3.5, 16.48\% for GPT-4, 22.28\% for LLAMA3-70b, and 20.40\% for Qwen1.5 72B.

    \item To mitigate length bias, we introduce \OURS{}, a method that aligns the response lengths of the reference and test model, enabling a fair comparison of desirability and information mass under the  same length intervals.

    \item Using both AlpacaEval and \OURS{}, we analyze the impact of length bias in DPO. Our experiments with TÜLU2 and TÜLU2-dpo models at 7B, 13B, and 70B scales show that  DPO leads to higher human preference, but this gain is amplified by response length, with AlpacaEval showing higher win rates gain than \OURS{}.

\end{itemize}

\section{Related Work}

\textbf{Reference-free Evaluation Metrics.}
Reference-free evaluation metrics have a long history~\citep{Louis2013AutomaticallyAM},  which evaluates the generated text based on intrinsic properties and coherence with the context. Although they achieve high accuracy on matching inner-annotator,  the achievement suffers from spurious correlations such as perplexity and length~\citep{Durmus2022SpuriousCI}. Recently,  people have started using a strong model (e.g.,  GPT-4) as an evaluator to perform a zero-shot reference-free evaluation on the weak models~\citep{Shen2023LooseLS, dubois2024alpacafarm, chen2023alpagasus,park2024disentangling}. 
However,  leveraging a strong model's intrinsic knowledge to perform reference-free evaluation ignores the prompt preference of the strong model,  for example,  the prompt's length.

\vspace{1.5mm}

\noindent \textbf{Correlation Between Length and Win Rate.}
Previous research reveals that sentence length will influence the evaluation of trustworthiness. Specifically,  when using a GPT-4 to represent human preference,  it will prefer to choose a long sentence rather than a short sentence~\citep{dubois2024length, ivison2023camels, gu2024minillm, hu2024rethinking, koo2023benchmarking, wang2023large, wu2023style, dubois2024alpacafarm, chen2023alpagasus,hu2024language}. Such preference will introduce a length-correlated bias and help the model with long-generation sentences gain a high score on human preference evaluation.
Although these approaches show a high correlation to human preference,  debiasing such as automated evaluation is highly valuable. \citep{dubois2024length} proposes a length-controlled (LC) win rate by removing the length-correlated term in the win rate regression model. The new LC win rate shows an even performance between concise and verbose input and a higher correlation when compared with human preference.

\section{\hspace{-0.2cm}Understanding the Major Factors of Win Rate}
\label{sec: WRDe}
To interpret the correlation between length and win rate, we propose
a new framework based on \textit{quality}, which includes \emph{desirability} (length-independent, related to trustworthiness) and \emph{information mass} (length-dependent, represented by conditional entropy). 
We validate our hypothesis through two scenarios: 
(1) testing the impact of different desirability on win rate with the same information mass, and 
(2) testing the influence of different information mass on win rate with the same desirability.

\subsection{Preliminary}

\textbf{Evaluation protocol.}
We utilize the AlpacaEval dataset~\citep{alpaca_eval} to assess human preferences. AlpacaEval is a reference-free evaluation dataset for LLMs, encompassing 805 instructions that reflect human interactions on the Alpaca web demo. To ensure a comprehensive evaluation of human preferences, we extend our testing to additional datasets, including LIMA~\citep{zhou2023lima}, Vicuna~\citep{vicuna2023}, Koala~\citep{vu2023koala}, Wizardlm~\citep{xu2023wizardlm}, and Self-Instruct~\citep{wang2022self}, in line with previous studies~\citep{chen2023alpagasus,zhang2024recost,du2023mods,zhao2024long,li2023quantity}.

\vspace{1.5mm}

\noindent \textbf{Base Models.}
In our experiments, we follow the setup in the AlpacaEval Leaderboard~\footnote{\url{https://tatsu-lab.github.io/alpaca_eval}}, using the GPT-4 Preview (11/06) as \textit{Baseline} and the \textit{Annotator}.
The references to GPT-3.5, LLAMA3-70b, and Qwen1.5 72b in the main text denote gpt-3.5-turbo-0125, meta-llama/Meta-Llama-3-70B-Instruct~\citep{llama3modelcard}, and Qwen/Qwen1.5-72B-Chat~\citep{qwen}, respectively.
Following previous work~\citep{wei2024large},
we calculate conditional entropy using the method described in ~\citep{von1932mathematische}.
Further details are provided in Appendix~\ref{sec:conditional-entropy}.

\vspace{1.5mm}

\noindent \textbf{Win rate.}
Assume we have a set of instructions $x$. We prompt a test model $m$ to generate a response $z_m$ for each instruction. Similarly, we prompt a reference model $b$ (referred to as the "baseline" in AlpacaEval) to generate a response $z_b$ for each instruction.\footnote{In this context, $m$ stands for "model" and $b$ denotes "baseline", which in this paper follows the AlpacaEval Leaderboard's use of GPT-4 Preview (11/06).}
An annotator then evaluates these responses based on their quality and assigns a preference $y \in \{m, b\}$, indicating which model's response is superior.
To properly understand the concept of win rate, we first need to define what we mean by response quality:

\begin{definition}
(\textbf{Response Quality}), denoted as $Q_e(z|x)$, quantifies the effectiveness of the model's response $z$ in addressing the given instruction $x$, as evaluated by an annotator $e$. Annotator prefer responses with higher quality.
\end{definition}
By leveraging the definition of quality, we can now formulate the win rate as the comparison of sentence quality as follows:
\begin{equation}
    \text{WinRate}(m, b) = \mathbb{E}_{x}\left[\mathbbm{1}_{Q_e(z_m\mid x) > Q_e(z_b\mid x)}\right],
\end{equation}
where $\mathbbm{1}$ is an indicator function and $\mathbbm{1}_{Q_e(z_m \mid x) > Q_e(z_b \mid x)}$ represents the preference distribution for each individual.
Previous works~\citep{chen2023alpagasus,alpaca_eval,dubois2024length,dubois2024alpacafarm} utilize LLMs as zero-shot evaluators due to their exceptional performance on real-world tasks.
Our experimental setup adheres to the AlpacaEval Leaderboard~\citep{alpaca_eval} guidelines, employing the GPT-4 Preview (11/06)\footnote{In this paper, unless specified otherwise, GPT-4 refers to GPT-4 Preview (11/06).} as both the \textit{Baseline} $b$ and the \textit{Annotator} $e$.

\begin{figure*}[h]
\centering
\scalebox{0.85}{
    \subfigure[\small AlpacaEval]{
        \includegraphics[width=0.315\textwidth]{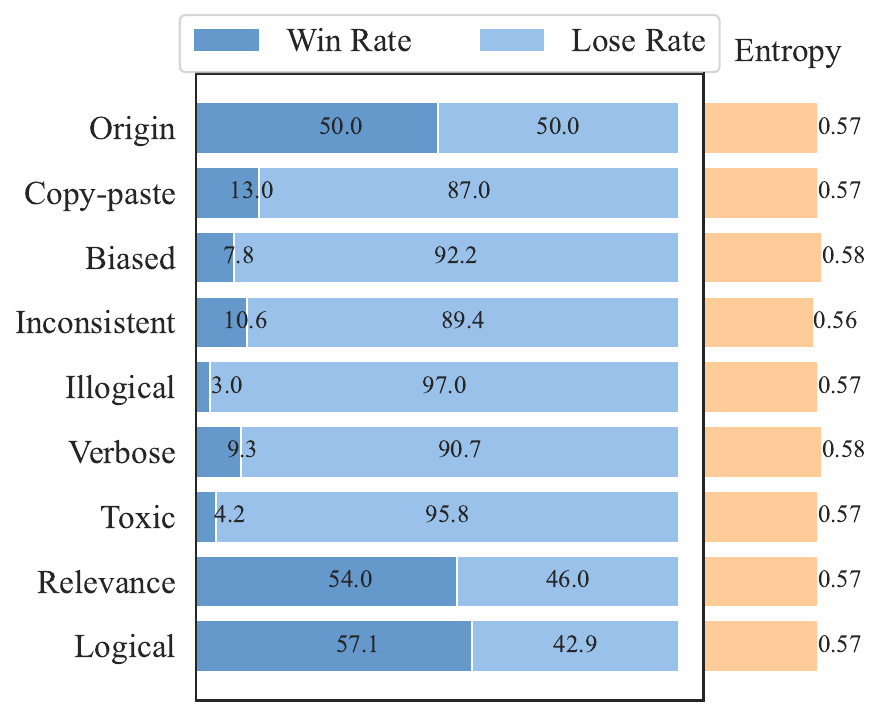}
    }
    \hspace{0.02\textwidth}
    \subfigure[\small LIMA]{
        \includegraphics[width=0.315\textwidth]{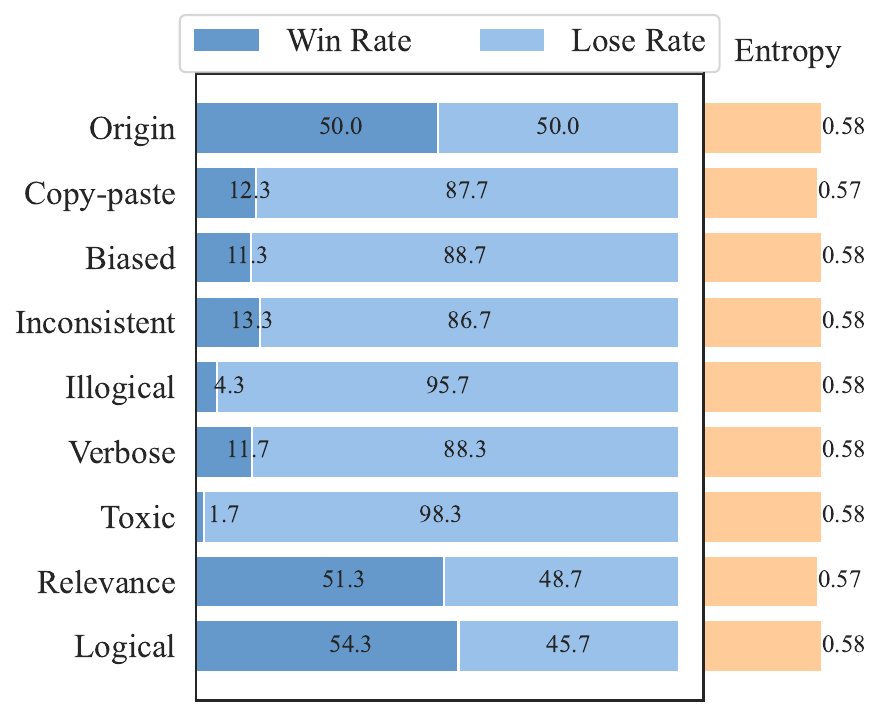}
    }
    \hspace{0.02\textwidth}
    \subfigure[\small Koala]{
        \includegraphics[width=0.315\textwidth]{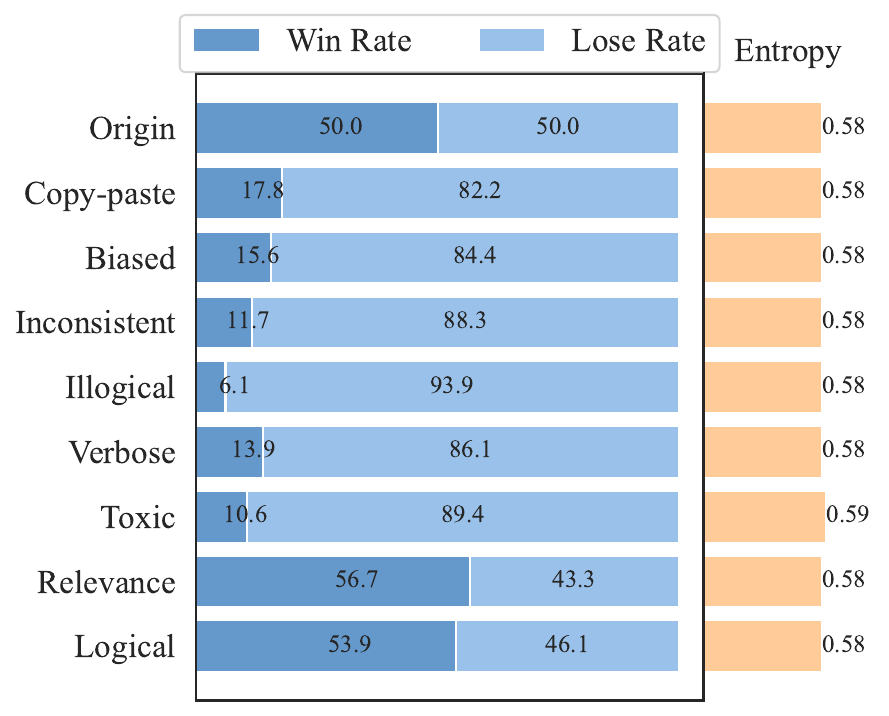}
    }
}

\vspace{10pt}

\scalebox{0.85}{
    \subfigure[\small Self-Instruct]{
        \includegraphics[width=0.315\textwidth]{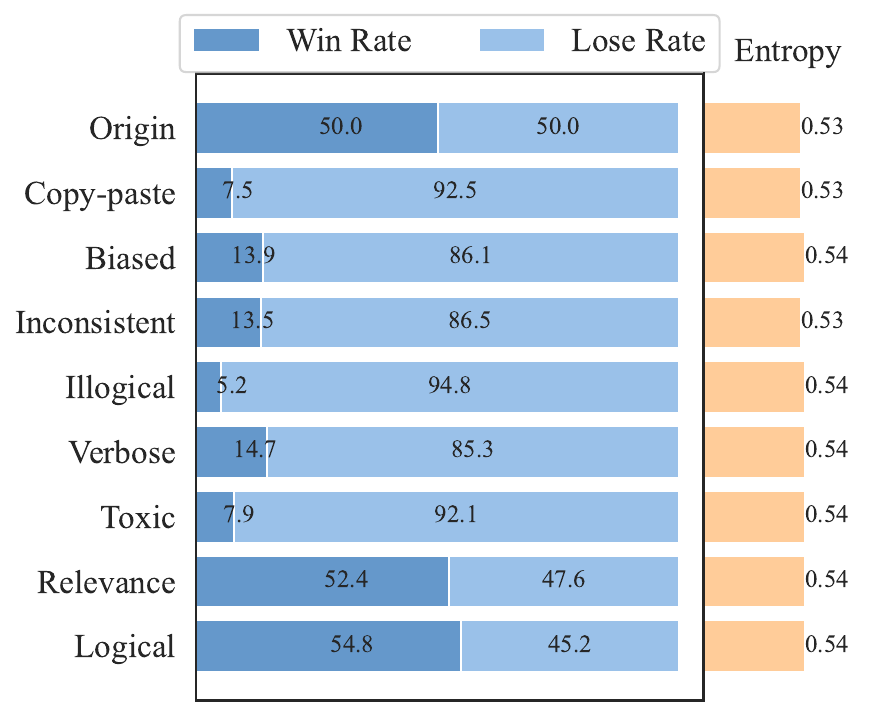}
    }
    \hspace{0.02\textwidth}
    \subfigure[\small Vicuna]{
        \includegraphics[width=0.315\textwidth]{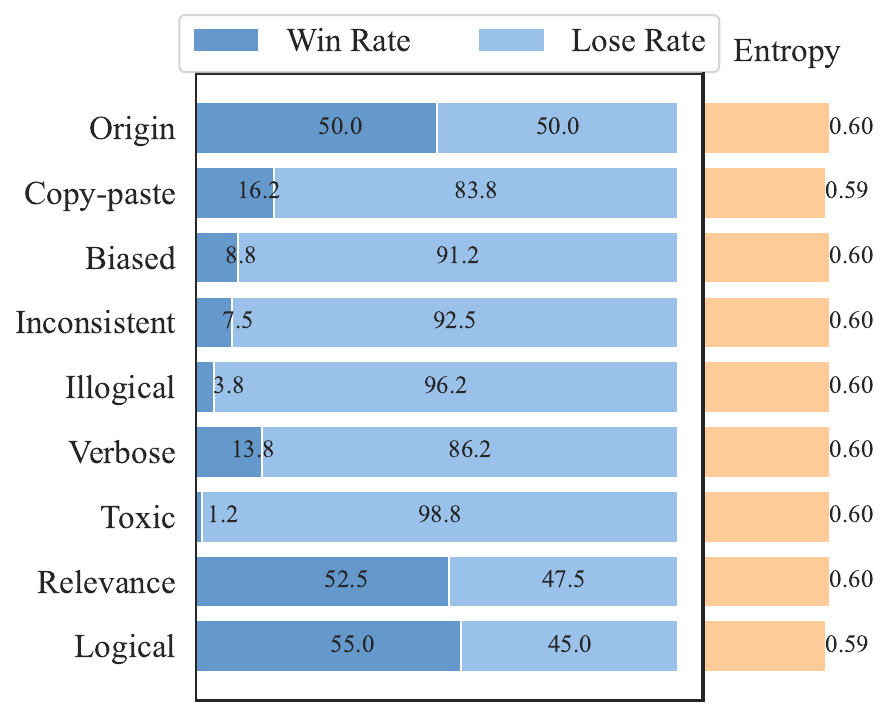}
    }
    \hspace{0.02\textwidth}
    \subfigure[\small Wizardlm]{
        \includegraphics[width=0.315\textwidth]{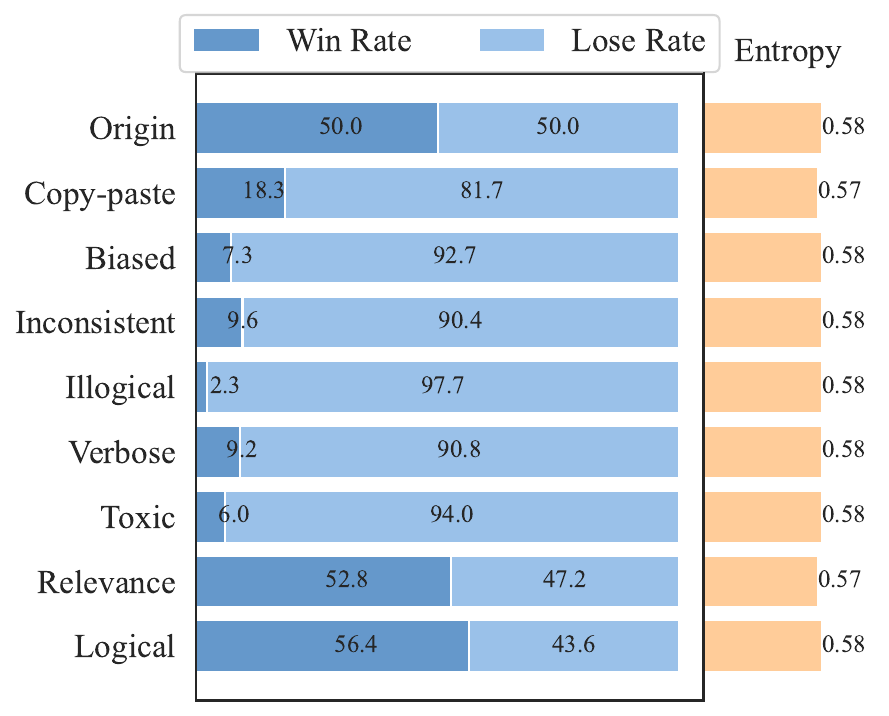}
    }
}

\caption{Validation of desirability's impact on quality for GPT-4. The results demonstrate that desirability influences the win rate.}
\label{fig:gpt4desirability}
\end{figure*}

\subsection{Quality Decomposition}
\label{sec:corr-length-human}

Before discussing the composition of quality, we first define two key concepts: \textbf{desirability} and \textbf{information mass}.
Desirability reflects the inherent quality attributes of a response that make it reliable and valuable, irrespective of its length, while information mass captures the quantity of information in the response, with longer responses generally containing more content. The definitions of desirability and information mass are as follows:

\begin{definition}
(\textbf{Desirability}), denoted as $D_e(z|x)$, measures the probability that  annotator $e$ will accept the response $z$ given an instruction $x$. 
It can be influenced by factors such as consistency and toxicity and is independent of response length. 
\end{definition}

\begin{definition}
(\textbf{Information mass}), denoted as $H_e(z|x)$, measures the amount of information in a response $z$ given an instruction $x$, as evaluated by annotator $e$. It is represented by conditional entropy and is directly with response length.
\end{definition}
Details on the design rationale are provided in Appendix~\ref{sec:design-rationale}.
With these definitions in place, we now present our main hypothesis on answer quality, starting with an assumption:

\begin{figure*}[h]
\centering
\scalebox{0.85}{
    \subfigure[\small AlpacaEval]{
        \includegraphics[width=0.315\textwidth]{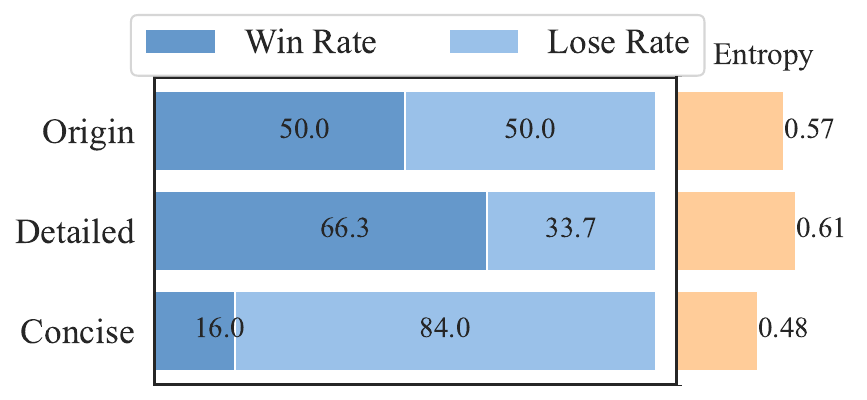}
    }
    \hspace{0.02\textwidth}
    \subfigure[\small LIMA]{
        \includegraphics[width=0.315\textwidth]{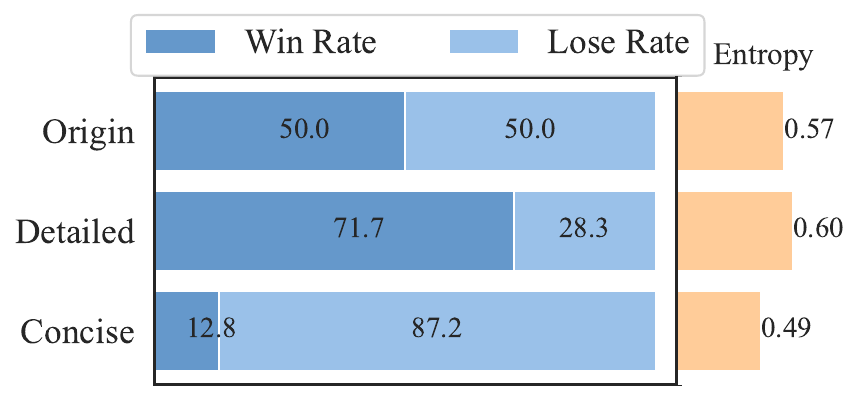}
    }
    \hspace{0.02\textwidth}
    \subfigure[\small Koala]{
        \includegraphics[width=0.315\textwidth]{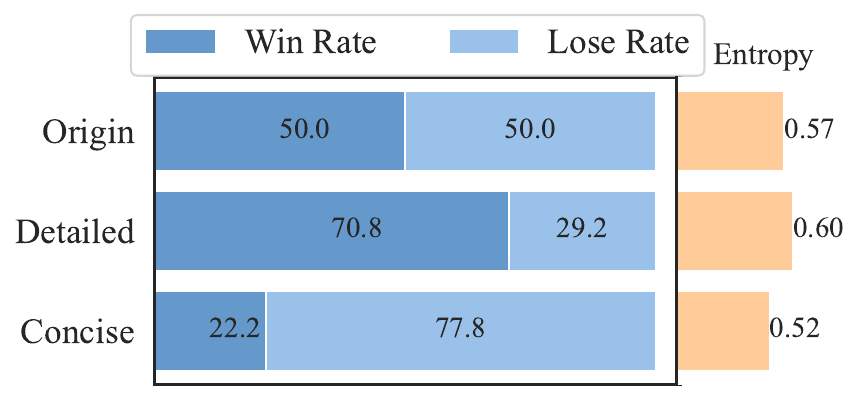}
    }
}

\vspace{10pt}

\scalebox{0.85}{
    \subfigure[\small Self-Instruct]{
        \includegraphics[width=0.315\textwidth]{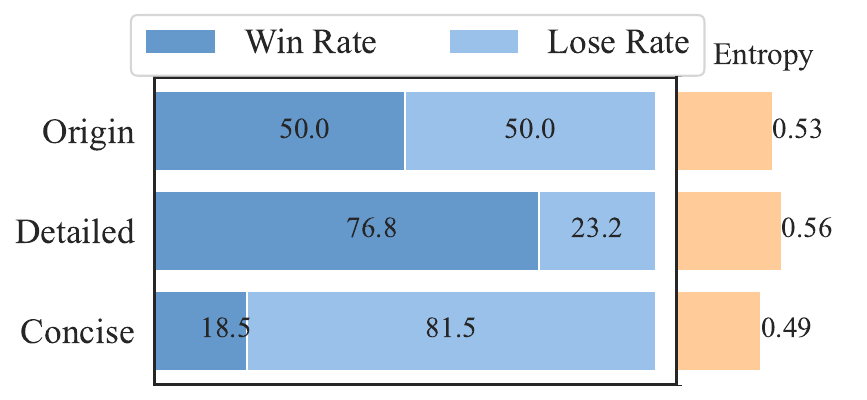}
    }
    \hspace{0.02\textwidth}
    \subfigure[\small Vicuna]{
        \includegraphics[width=0.315\textwidth]{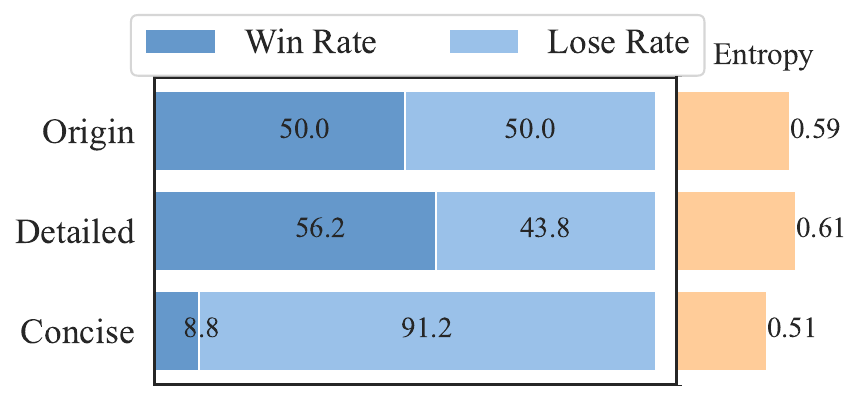}
    }
    \hspace{0.02\textwidth}
    \subfigure[\small Wizardlm]{
        \includegraphics[width=0.315\textwidth]{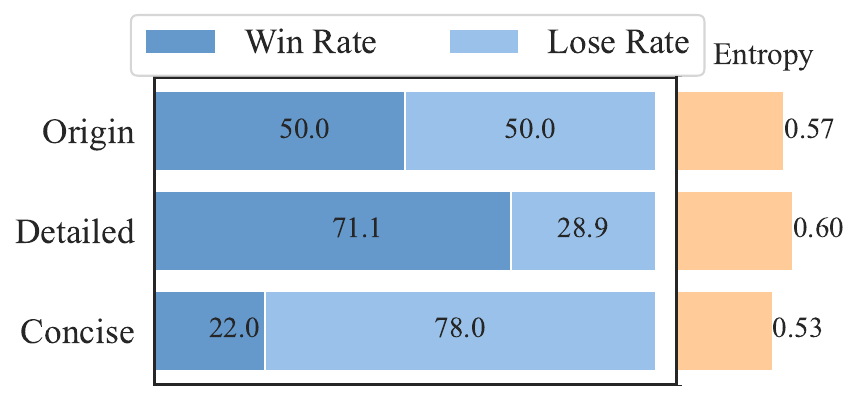}
    }
}

\caption{Validation of information mass's impact on quality for GPT-4. The results demonstrate that information mass influences the win rate.}
\label{fig:GPT4ConditionalInformation}
\end{figure*}

\begin{assumption}
\label{ass1}
(\textbf{Quality Decomposition}). 
For a given answer $z$ and instruction $x$, the quality $Q_e(z|x)$ recognized by annotator $e$ 
can be viewed as a function of two key components: 
\small
\begin{equation}
Q_e(z|x) \leftarrow \{D_e(z|x), H_e(z|x)\},
\end{equation}
\normalsize
where $D_e(z|x)$ denotes the desirability of the response, and $H_e(z|x)$ represents the information mass.
\end{assumption}
To systematically verify our hypothesis, we conduct two experiments targeting the manipulation of these key components in GPT-4's responses in Section~\ref{sec:desirability-Influence-WR} and Section~\ref{sec:infoMass-Influence-WR}. 
Additional results with more test and annotator models are provided in Appendix~\ref{sec:QEwithMoreTestModel} and Appendix~\ref{sec:QEwithMoreAnnModel}.

\subsection{Desirability Influences Quality}
\label{sec:desirability-Influence-WR}
To evaluate the impact of desirability on quality, we design experiments using eight strategies to manipulate response desirability.
These strategies include:
(1) \textbf{Origin}: No prompt restrictions.
(2) \textbf{Copy-paste}: Copy GPT-4's response three times.
(3) \textbf{Biased}: Provide biased responses, favoring certain ideas without justification.
(4) \textbf{Inconsistent}: Provide contradictory information to create confusion.
\textbf{Illogical}: Give responses based on flawed logic or irrelevant information.
(5) \textbf{Verbose}: Provide lengthy responses filled with broad, unrelated details.
(6) \textbf{Toxic}: Use offensive language with an aggressive tone.
(7) \textbf{Relevant}: Provide responses that align with the query.
(8) \textbf{Logical}: Base responses on sound reasoning and valid arguments.
The results are shown in Figure~\ref{fig:gpt4desirability}.
To eliminate the impact of information mass on win rate, 
we use conditional entropy to represent information mass 
and ensure the information mass of \textbf{Origin} and the other prompts remains as consistent as possible.
The entropy values shown in the Figure~\ref{fig:gpt4desirability} represent the average conditional entropy of the responses for each prompt.
Details for these prompts and relevant implementation are shown in 
Appendix~\ref{sec:prompt-content}, Appendix~\ref{sec:infoconsis} and Appendix~\ref{sec:desircasestudy}.
First, we observe that although the \textbf{Copy-paste} and \textbf{Origin} prompts maintain identical information mass (as simply replicating text does not increase information), the win rates of \textbf{Copy-paste} fall below \textbf{Origin} (50\%) due to significant consistency impairments. 
Second, responses generated from negative prompts (i.e., \textbf{Biased}, \textbf{Inconsistent}, \textbf{Illogical}, \textbf{Verbose}, and \textbf{Toxic}) exhibit low desirability, resulting in win rates substantially lower than \textbf{Origin} (50\%), despite having similar information mass. 
Conversely, prompts enhancing desirability (i.e., \textbf{Consistent} and \textbf{Logical}) yield increased win rates compared to \textbf{Origin}. 
In summary, desirability plays a significant role in determining quality.

\subsection{Information Mass Influences Quality}
\label{sec:infoMass-Influence-WR}
To evaluate the impact of information mass on quality, we designed experiments using three distinct strategies to manipulate the information mass of responses.
These strategies include:
(1) \textbf{Origin}: No prompt restrictions.
(2) \textbf{Concise}: Request brief responses focusing on the most crucial points.
(3) \textbf{Detailed}: Request comprehensive responses covering all relevant aspects thoroughly.
The corresponding results are illustrated in Figures~\ref{fig:GPT4ConditionalInformation}.
Importantly, to isolate the effect of information mass, we ensured that the prompts did not impose any constraints on desirability,
ensuring comparability. 
Details of the prompts and implementation are in Appendix~\ref{sec:prompt-content}, Appendix~\ref{sec:exp-setup} and Appendix~\ref{sec:word_counts}.
Our findings indicate that information mass significantly affects the win rate without a negative desirability prompt. 
Specifically, responses with higher information mass, measured by conditional entropy, consistently achieved higher win rates. 
Thus, we observe the following relationship: \textbf{Detailed} > \textbf{Origin} > \textbf{Concise}. 
These results confirm that information mass is a crucial factor influencing the quality of responses.

\begin{table}[h]
\centering
\caption{The content of the "Quality Enhancement" prompt, designed to elevate both the information mass and desirability of responses, thereby enhancing win rates. Keywords such as "relevant" and "logical" are used to enhance desirability, while "detailed" is used to boost information mass.}
\begin{tabular}{m{0.88\linewidth}} 

\begin{tcolorbox}[mybox, title=Quality Enhancement]

You are an expert assistant, delve deeply into the core of the topic, providing a richly \textbf{detailed} response that explores all its dimensions. Ensure each part of your response is \textbf{relevant} to the query in a \textbf{logical} manner. Your response should provide comprehensive information and thoroughly cover all relevant aspects with accuracy and depth.

\end{tcolorbox}
\end{tabular}

\label{table:explanation}
\end{table}

\begin{table*}[t]
\caption{Win rates with and without the "Quality Enhancement" prompt, along with the corresponding win rate gains (WR Gain). 
"WR Gain" represents the increase in win rate due to the use of the "Quality Enhancement".}
\centering
\scalebox{0.70}{
\begin{tabular}{@{}llccccccc@{}}
\toprule
                 Models           &   Methods      & AlpacaEval & LIMA  & Koala & Self-Instruct & Vicuna & Wizardlm & Avg.   \\ \midrule
\multirow{3}{*}{GPT-3.5}    & w/o Quality Enhancement  & 15.47      & 9.67  & 11.39 & 21.46         & 8.75   & 16.82    & 13.93  \\
                            & with Quality Enhancement & 29.89      & 36.53 & 40.34 & 45.93         & 35.88  & 35.62    & 37.36 \\
                            & WR Gain & 14.42      & 26.86 & 28.95 & 24.47         & 27.13  & 18.80    & 23.44  \\ \midrule
\multirow{3}{*}{GPT-4}      & w/o Quality Enhancement  & 50.00      & 50.00 & 50.00 & 50.00         & 50.00  & 50.00    & 50.00  \\
                            & with Quality Enhancement & 70.16      & 65.84 & 58.90 & 67.06         & 73.13  & 63.76    & 66.48  \\
                            & WR Gain & 20.16      & 15.84 & 8.90  & 17.06         & 23.13  & 13.76    & 16.48  \\ \midrule
\multirow{3}{*}{LLAMA3-70b} & w/o Quality Enhancement  & 34.32      & 36.63 & 40.12 & 39.70         & 36.74  & 36.99    & 37.81  \\
                            & with Quality Enhancement & 56.50      & 60.39 & 61.30 & 64.81         & 63.49  & 51.70    & 59.70  \\
                            & WR Gain & 22.18      & 23.76 & 21.18 & 25.11         & 26.75  & 14.71    & 22.28  \\ \midrule
\multirow{3}{*}{Qwen1.5 72b}   & w/o Quality Enhancement  & 28.27      & 28.40 & 35.25 & 33.81         & 33.70  & 31.80    & 32.67  \\
                            & with Quality Enhancement & 48.87      & 53.34 & 55.40 & 52.43         & 56.49  & 47.13    & 52.28  \\
                            & WR Gain & 20.60      & 24.94 & 20.15 & 18.62         & 22.79  & 15.33    & 20.40  \\ \bottomrule
\end{tabular}

}
\label{table:IT-gain-4model}
\end{table*}

\begin{figure*}[h]
\centering
\scalebox{0.80}{
    \subfigure[\small AlpacaEval]{
        \includegraphics[width=0.315\textwidth]{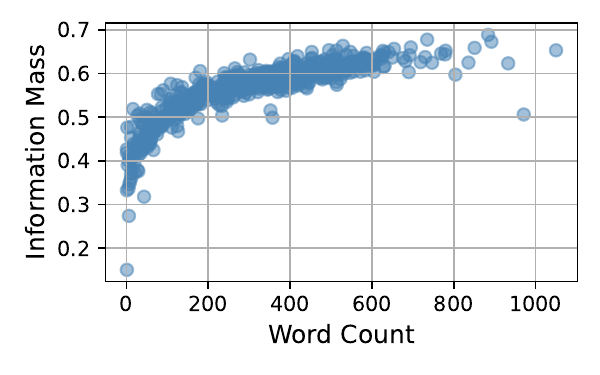}
    }
    \hspace{0.02\textwidth}
    \subfigure[\small LIMA]{
        \includegraphics[width=0.315\textwidth]{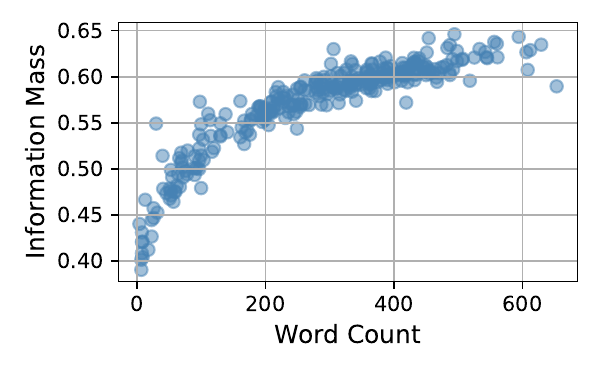}
    }
    \hspace{0.02\textwidth}
    \subfigure[\small Koala]{
        \includegraphics[width=0.315\textwidth]{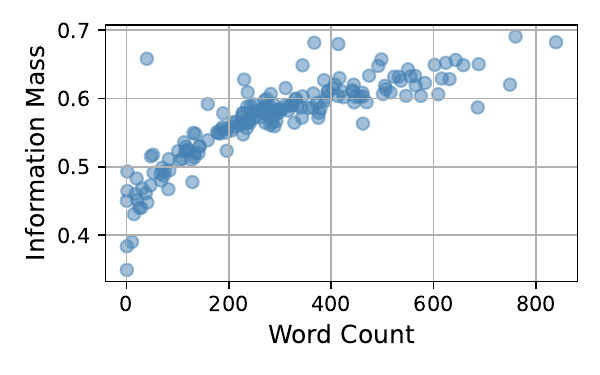}
    }
}

\vspace{10pt}

\scalebox{0.80}{
    \subfigure[\small Self-Instruct]{
        \includegraphics[width=0.315\textwidth]{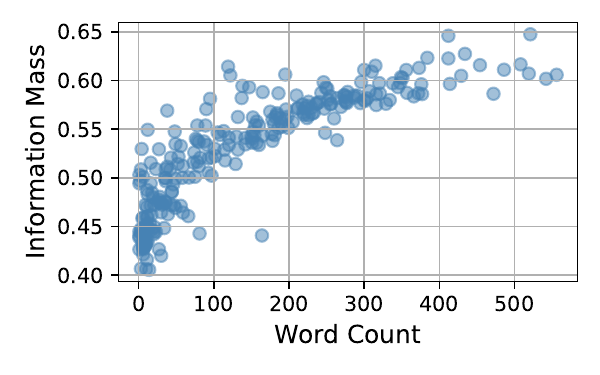}
    }
    \hspace{0.02\textwidth}
    \subfigure[\small Vicuna]{
        \includegraphics[width=0.315\textwidth]{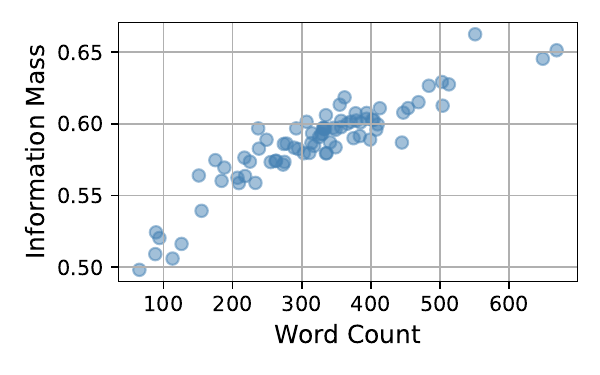}
    }
    \hspace{0.02\textwidth}
    \subfigure[\small Wizardlm]{
        \includegraphics[width=0.315\textwidth]{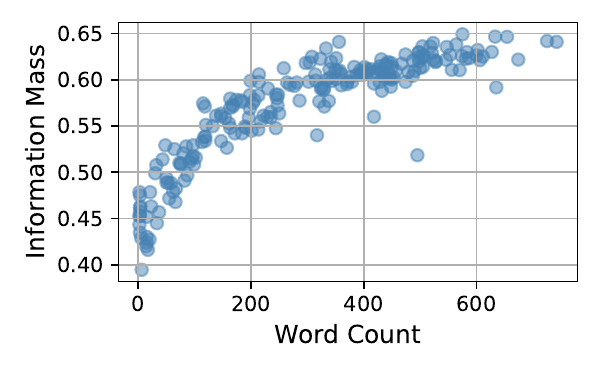}
    }
}

\caption{Correlation between information mass and word count for responses of GPT-4. 
As the word count increases, the information mass also increases.}

\label{fig:GPT4-length-infomass}
\end{figure*}

\subsection{Quality Enhancement Prompt}
\label{sec:QE}

Our decomposition reveals that responses with good desirability and higher information mass are generally more favored. 
Building on this insight, we propose the 'Quality Enhancement' prompt (Table~\ref{table:explanation}), designed to improve both desirability and information mass, thereby increasing win rates.
The keywords "relevant" and "logical" are used to enhance desirability, while "detailed" is used to boost information mass. 
Their effectiveness is validated in Section~\ref{sec:desirability-Influence-WR} and Section~\ref{sec:infoMass-Influence-WR}.
We evaluated this prompt across multiple models, including GPT-3.5, GPT-4, LLAMA3-70b, and Qwen1.5 72B. 
The results, summarized in Table~\ref{table:IT-gain-4model}, with benchmarks such as LIMA, Vicuna, Koala, Wizardlm, and Self-Instruct.
The consistent improvement in win rates across all tested models underscores the critical role of response quality in LLM evaluation.

\begin{figure*}[h] \centering
\small
\includegraphics[width=1.0\linewidth]{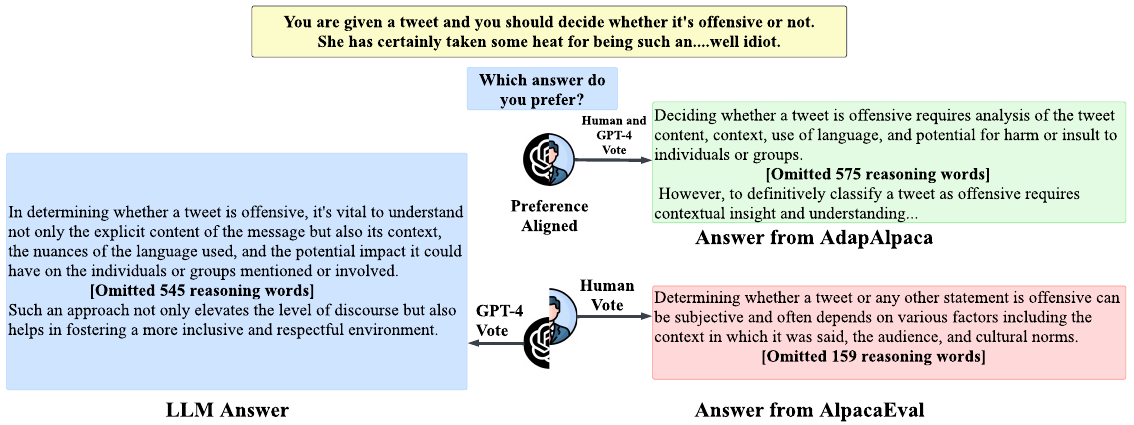}
\caption{Case study on comparing GPT-4 and human vote on AlpacaEval and \OURS{}. In AlpacaEval, GPT-4 votes for the verbose answer, but humans vote for the concise reference answer, while in \OURS{}, GPT-4 and humans vote for the same answer, demonstrating a better LLM-human alignment on \OURS{}.}
\label{fig:casestudy1}
\end{figure*}

\section{Adaptive AlpacaEval}

Here, we analyze the phenomenon observed in prior works~\citep{dubois2024length, chen2023alpagasus, dubois2024alpacafarm}, which highlights a positive correlation between response length and win rate. 
Intuitively, longer responses tend to encompass more information. 
To rigorously quantify this relationship, we use conditional entropy as information mass in a response $z$ given an instruction $x$.
This analysis is conducted without constraints on response desirability, ensuring the correlation between length and information mass remains independent of desirability factors.
As shown in Figure~\ref{fig:GPT4-length-infomass}, our analysis demonstrates a clear trend: as the length of a response increases, the information mass also grows. 
By integrating this observation with the findings from Section~\ref{sec:corr-length-human}, we conclude that the primary mechanism through which length affects win rate is its contribution to the overall information mass.

\noindent Adaptive AlpacaEval (AdapAlpac) is based on the premise that a reliable evaluation metric should not only assess the content quality but also ensure that the assessment is not confounded by extraneous factors such as the length of the response. 
Central to this approach is the concept of information mass, which is inherently dependent on response length and can be quantified using conditional entropy. 
Our primary aim is to mitigate scenarios where merely extending the length of a response artificially inflates its conditional entropy and, thus, its perceived quality by annotators.
This approach involves dynamically adjusting the evaluation criteria based on response length, thereby providing a more equitable and accurate measure of a model's performance.
More discussion can be found in Appendix~\ref{sec:diss-lgb} and~\ref{sec:discuss-length-bias}.

\subsection{Dataset Generation}
\label{sec:datasetGen}

To support the development of Adaptive AlpacaEval, we first generate a diverse dataset using a modified prompting strategy with GPT-4, designed to produce responses within specific word count ranges.
Specifically, we analyzed the word count distribution within the AlpacaEval dataset, observing that responses predominantly fall within the 0-1000 word range. This range was chosen to encompass the full spectrum of response lengths present in the original AlpacaEval dataset, ensuring comprehensive evaluation coverage. 
To systematically explore this range, we divided it into five equal segments, each representing a distinct dataset:
\OURS{}-200: 0-200 words,
\OURS{}-400: 200-400 words,
\OURS{}-600: 400-600 words,
\OURS{}-800: 600-800 words,
\OURS{}-1000: 800-1000 words.
Each segment is populated by generating responses using the dataset generation prompt, with GPT-4 configured to produce responses that strictly conform to the specified word counts. 
The data generation prompt and additional details for \OURS{} can be found in Appendix~\ref{sec:dataInfo}.

\subsection{Case Study}
To demonstrate the superiority of \OURS{}, we present a case study. 
In Figure~\ref{fig:casestudy1}, for the given instruction, we generate a redundant model answer (shown in the blue box).
When evaluated using the current AlpacaEval response (shown in the red box), the annotator (i.e., GPT-4) selected this redundant answer, which is significantly unaligned from human preference, as the simplicity of the question does not warrant such extensive verbosity. 
The reason GPT-4 chose this answer is that the excessive length increases the information mass, artificially inflating the perceived quality.
In contrast, when using \OURS{}, it allows us to control for content while varying the length, thereby isolating the effect of length from that of content quality.

\begin{table}[h]
\caption{
The subscripts in the LCWR and WR columns indicate the differences between these metrics and the corresponding Human WR. A larger absolute value denotes a greater disparity between the annotator's evaluation and Human Preference. "LLM Response" denotes different responses to AlpacaEval questions for GPT-4, with detailed content available in Section~\ref{sec:corr-length-human}.
}
\centering
\scalebox{0.6}{
\begin{tabular}{@{}lccccc@{}}
\toprule
\multirow{2}{*}{LLM Response}               & \multicolumn{3}{c}{AlpacaEval}           & \multicolumn{2}{c}{\OURS{}} \\ \cmidrule(l){2-4} \cmidrule(l){5-6} 
                    & Human & LCWR  & WR  & Human  & WR  \\ \midrule
Concise             & 10.81 & $35.16_{+24.35}$  & $15.96_{+5.15}$  & 29.56    & $28.44_{-1.12}$    \\ \midrule
Detailed            & 61.61 & $54.13_{-7.48}$    & $65.83_{+4.22}$  & 56.02    & $55.36_{-0.78}$   \\ \midrule
Quality Enhancement & 66.70 & $49.37_{-17.33}$   & $70.16_{+3.46}$    & 58.88    & $57.81_{+1.07}$   \\ \bottomrule
\end{tabular}

}

\label{table:humanstudy}
\end{table}

\subsection{Result of Human Evaluation}
\label{sec:humanstudy-eavl}

Table~\ref{table:humanstudy} presents the results of the human study, with details provided in Appendix~\ref{sec:humanstudy} and you can find more  results in Appendix~\ref{sec: More-Hmresult}.
First, we test the results of concise,  detail, and quality enhancement (descriptions provided in Section~\ref{sec: WRDe} and ~\ref{sec:QE}) using AlpacaEval, followed by \OURS{}.
From the gap values between LCWR and human evaluations, we observe significant misalignments, indicating inherent problems with the LCWR metric. 
In contrast, the win rate calculated using \OURS{} closely aligns with the human results, showing an average difference of 0.99\% (1.12\% + 1.07\% + 0.78\% / 3). 
Additionally, we find that the difference between human evaluation and WR decreases as the quality of responses improves (from concise to detailed to Quality Enhancement).
This suggests that as response quality increases, the preferences of annotators and human evaluators converge.
Moreover, we found that the smallest difference in win rate between GPT-4 and human evaluations occurs when using the "Quality Enhancement" prompt, which has the highest levels of desirability and information mass.
This further underscores the importance of enhancing both desirability and information mass in model responses.
Overall, while both \OURS{} and LCWR aim to mitigate length bias in evaluating human preferences, their approaches differ fundamentally.
\OURS{} eliminates length bias from the outset, whereas LCWR attempts to correct for length bias after it has already influenced the evaluation.
The inherent issue with LCWR is that length significantly impacts human preference, and adjusting for length retrospectively is not a reliable approach.

\subsection{DPO and Its Length Bias}

Previous work~\citep{gu2024minillm,ivison2023camels} has shown that DPO~\cite{Rafailov2023DirectPO} tends to make model responses longer, raising a natural question: 
Does the increase in human preference brought by DPO partly stem from the length of the responses? 
In other words, does DPO generate longer replies, thereby increasing their win rate?
To investigate this issue, we conducted tests using the widely-used TÜLU2~\citep{ivison2023camels} series models. 
As shown in Table~\ref{tab:DPO-lengthbias}, we tested the models at 7B, 13B, and 70B scales on both AlpacaEval and \OURS{} to measure their win rates and corresponding response lengths. 
The results from AlpacaEval and \OURS{} indicate that while DPO does lead to longer model responses, it enhances the model's human preference capability (as evidenced by the increased win rate in \OURS{}). 
However, this gain is amplified by the response length (as the win rate in AlpacaEval is higher than in \OURS{}).
Additionally, we found that all models have higher win rates on \OURS{} compared to AlpacaEval. 
This is because the responses from GPT-4 (1106) on AlpacaEval are longer (363 words, see Appendix~\ref{sec:analysis}), which unfairly amplifies the capabilities of GPT-4 due to its length.
These results emphasize the need for length control in evaluations to reflect true model performance.

\begin{table}
\caption{
Win rate and response length comparison for TÜLU2 models (7B, 13B, and 70B) on AlpacaEval and \OURS{}. 
The results indicate that while DPO increases response length and improves win rate, the win rate gain is further amplified by the response length, leading to higher performance in AlpacaEval compared to \OURS{}.
}

\scalebox{0.73}{
\begin{tabular}{@{}llccc@{}}
\toprule
\multirow{2}{*}{Size} & \multirow{2}{*}{Model} & \multicolumn{2}{c}{Winrate (\%)} & \multirow{2}{*}{Avg. Length} \\ \cmidrule(l){3-4} 
                      &                        & AlpacaEval      & AdapAlpaca     &                                                               \\ \midrule
\multirow{3}{*}{7B}   & TÜLU 2                 & 3.60            & 5.84           & 203.60                                                        \\ \cmidrule(l){3-5} 
                      & TÜLU 2+DPO             & 8.33            & 9.04           & 282.92                                                        \\ \cmidrule(l){3-5} 
                      & Gain from DPO          & 4.73            & 3.20           & -                                                             \\ \midrule
\multirow{3}{*}{13B}  & TÜLU 2                 & 4.35            & 8.07           & 192.58                                                        \\ \cmidrule(l){3-5} 
                      & TÜLU 2+DPO             & 10.82           & 13.17          & 276.96                                                        \\ \cmidrule(l){3-5} 
                      & Gain from DPO          & 6.47            & 5.10           & -                                                             \\ \midrule
\multirow{3}{*}{70B}  & TÜLU 2                 & 7.34            & 10.94          & 184.26                                                        \\ \cmidrule(l){3-5} 
                      & TÜLU 2+DPO             & 15.67           & 17.90          & 267.23                                                        \\ \cmidrule(l){3-5} 
                      & Gain from DPO          & 8.33            & 6.96           & -                                                             \\ \bottomrule
\end{tabular}

}
\vspace{4pt}
\label{tab:DPO-lengthbias}
\end{table}

\section{Length Bias Originating from RLHF}
\label{sec:lengthbiasRLHF}

\begin{table*}[t]
\centering
\caption{Scores given by commonly used reward models to concise, detailed, and original responses from GPT-4. The analysis shows that the scores consistently decrease from detailed to concise responses, highlighting the length bias within the reward model.}
\scalebox{0.765}{%
\begin{tabular}{@{}lccccccccccc@{}}
\toprule
\multirow{2}{*}{LLM Response} & \multicolumn{10}{c}{Reward Model}                                                     & Avg.  \\ \cmidrule(l){2-12} 
                              & Eurus & Grmdis & Grmsft & UniF  & Debba & Bebla & FsfairRM & Gerew  & Misrmr & InteRM &       \\ \midrule
Concise                       & 1.819 & 1.984  & -2.919 & 0.064 & 2.229 & 4.159 & -1.404   & -0.456 & 5.661  & 0.426  & 1.156 \\ \midrule
Origin                        & 3.564 & 4.009  & -0.505 & 2.901 & 3.305 & 5.142 & 1.830    & 1.066  & 9.440  & 1.558  & 3.231 \\ \midrule
Detailed                      & 3.986 & 4.646  & 1.458  & 3.263 & 3.759 & 5.450 & 2.684    & 2.630  & 10.616 & 2.416  & 4.090 \\ \bottomrule
\end{tabular}%
}
\label{tab:RLHF-rewardmodel}
\end{table*}

\begin{figure}[h] 
\centering
\includegraphics[width=0.8\linewidth]{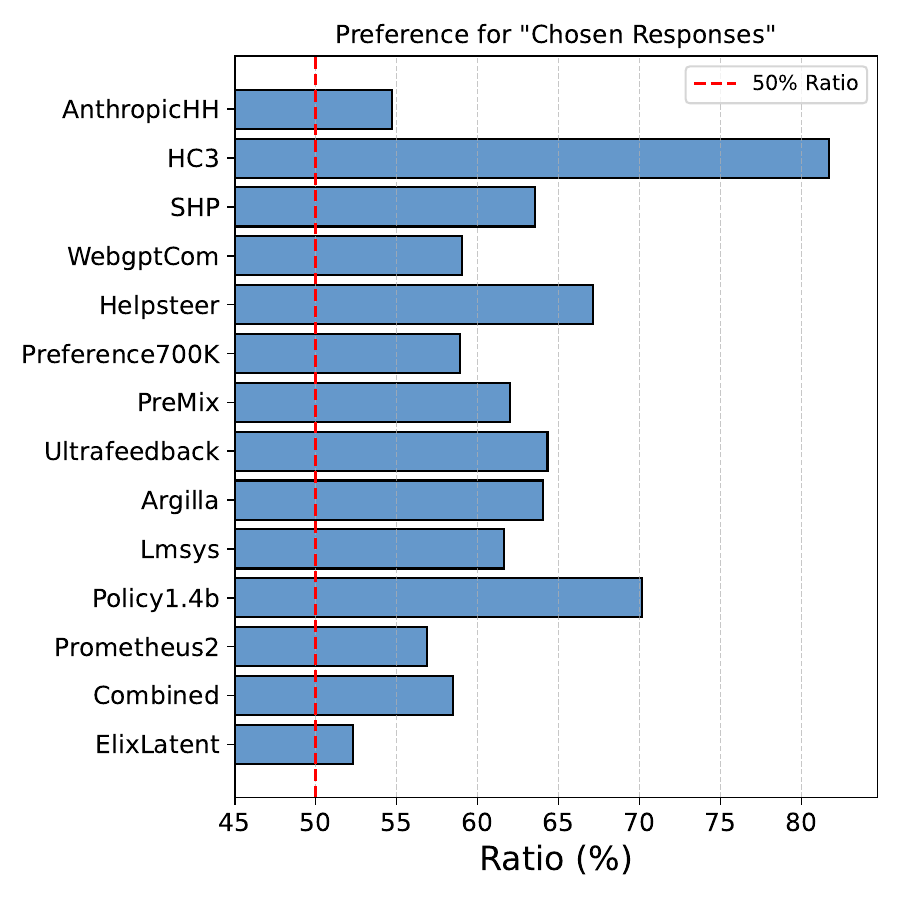}
\caption{Analysis of the 14 commonly used preference datasets on Hugging Face. The analysis shows that the lengths of chosen responses are generally longer than those of rejected responses, indicating a length bias in human preference labeling.}
\label{fig:RLHF-dataset}
\end{figure}

\begin{figure}[t!] 
\centering
\includegraphics[width=0.8\linewidth]{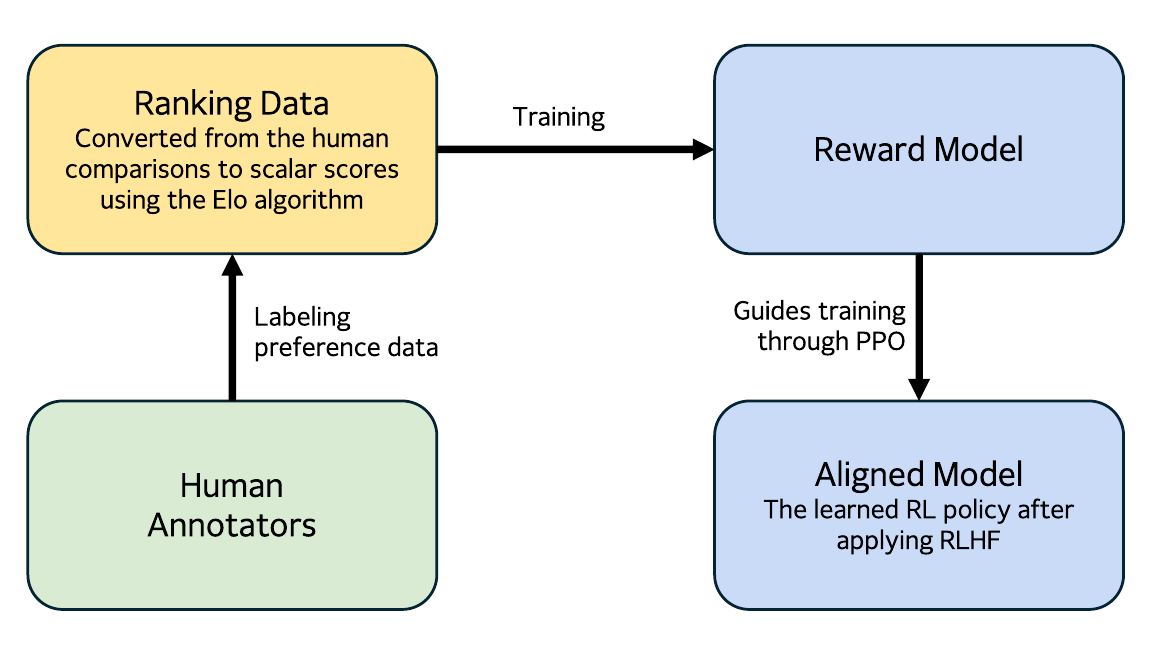}
\caption{RLHF process contributing to length bias in LLMs. Human labelers often prefer detailed responses, leading to ranking data where longer responses are ranked higher. This creates a spurious correlation that the reward model learns and propagates to the aligned model.}
\label{fig:RLHF}
\end{figure}

We believe that the length bias observed in LLMs essentially originates from the RLHF~\citep{Ouyang2022TrainingLM} process.
As shown in Figure~\ref{fig:RLHF}, during the RLHF process, humans may generally prefer more detailed responses when labeling preference data. 
This leads to ranking data where longer responses are generally ranked higher than shorter ones, causing the reward model to learn this spurious correlation and incorrectly assume that length is a factor in human preference. 
This bias is further propagated to the aligned model during the training process using the reward model.

\noindent To verify our idea, we first analyze 14 commonly used preference datasets in huggingface, shown in Figure~\ref{fig:RLHF-dataset}.
We found that the lengths of chosen responses are generally longer than those of rejected responses.
As detailed in Table~\ref{tab:RLHF-rewardmodel}, we also analyze the scores given by 10 commonly used reward models~\citep{lambert2024rewardbench} to detailed, original, and concise responses from  GPT-4. The detailed description of these three prompts can be found in Section~\ref{sec:infoMass-Influence-WR}.
We find that the scores consistently decrease across all reward models.
The details of these datasets and reward models can be found in Appendix~\ref{sec:preferData-RewardModel}. 
However, attributing human preference solely to response length is an oversimplification, as length is merely a superficial factor in how humans judge the quality of a sentence.

\section{Conclusion}
\vspace{-1.5mm}
In this paper, we identify and address the critical issue of length bias in LLM-based preference evaluations, which undermines the reliability of win rate metrics. 
By decomposing win rate into desirability and information mass, we offer a nuanced understanding of response quality. 
Our proposed framework, \OURS{}, effectively mitigates length bias by dynamically adjusting reference answer lengths to match test model responses, ensuring fairer evaluation metrics. 
Additionally, our analysis of DPO demonstrates that its gains in human preference are influenced by response length, underscoring the importance of unbiased evaluation benchmarks. 
Overall, \OURS{} provides a robust tool for advancing reliable and equitable model evaluation.

\section{Limitations}
\vspace{-1.0mm}
This study recognizes several limitations that define the scope of our current work and suggest directions for future research. First, \OURS{} adjusts the length of reference answers to match test model responses, but this assumes that content quality is preserved across varying lengths. In practice, truncating or expanding a response may distort its meaning or quality, introducing unintended artifacts into the evaluation. Second, while our work primarily focuses on length bias, other biases in LLM evaluations—such as stylistic preferences, formatting, or cultural influences—remain unexplored. These factors could also shape win rates and deserve systematic attention in future studies.

\section{Acknowledgement}
\vspace{-1.0mm}
This work was supported in part by the National Key R\&D Program of China (Grant No.2023YFF0725001),in part by the National Natural Science Foundation of China (Grant No.92370204),in part by the guangdong Basic and Applied Basic Research Foundation (Grant No.2023B1515120057),in part by the Education Bureau of Guangzhou 

\clearpage

\bibliography{reference}

\clearpage

\appendix

\begin{figure*}[ht]
    \centering
    \includegraphics[width=0.9\linewidth]{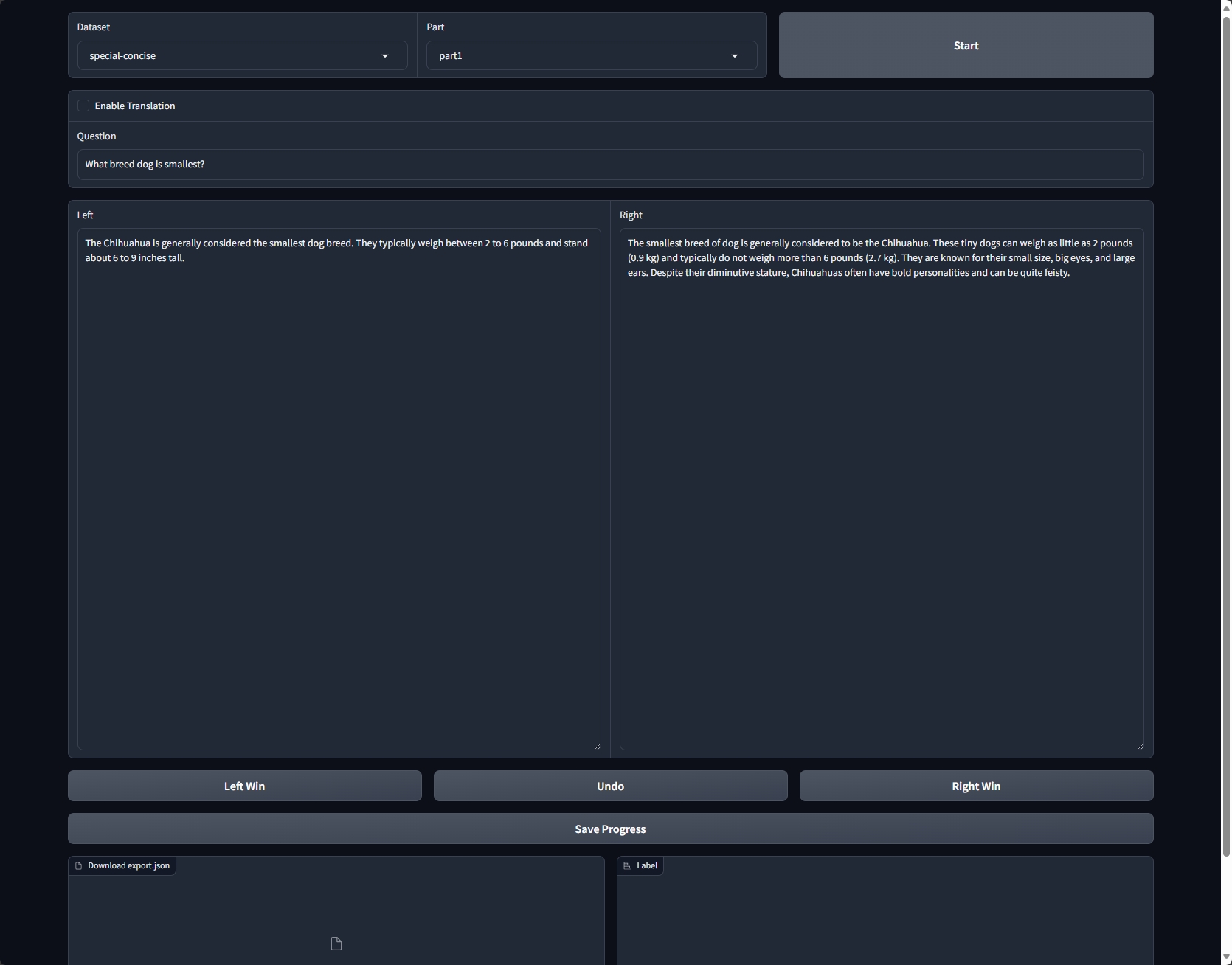}
    \caption{Example of the evaluation interface used in the human study, showing two outputs for a single input query. Participants assessed which output more accurately addressed the query, demonstrating the interface's role in ensuring unbiased evaluation.}
    \label{fig:gradio_interface}
\end{figure*}

\section{Implementation Detail}
\label{sec:ImplementationDetail}

\subsection{Experiment Setup}
\label{sec:exp-setup}
In our experiments, we follow the setup in the AlpacaEval Leaderboard\footnote{\url{https://tatsu-lab.github.io/alpaca_eval}}, using the GPT-4 Preview (11/06) as \textit{Baseline} and the \textit{Annotator}.
The references to GPT-3.5, LLAMA3-70b, and Qwen1.5 72b in the main text denote gpt-3.5-turbo-0125, meta-llama/Meta-Llama-3-70B-Instruct~\footnote{\url{https://huggingface.co/meta-llama/Meta-Llama-3-70B-Instruct}}, and Qwen/Qwen1.5-72B-Chat~\footnote{\url{https://huggingface.co/Qwen/Qwen1.5-72B-Chat}}, respectively.

\subsection{ Ensuring Information Mass Consistency}
\label{sec:infoconsis}
To ensure the information masses across prompts were consistent in Section~\ref{sec:desirability-Influence-WR}, we followed an approach similar to the methodology in~\citet{yuan2024self}. Specifically, we prompted the LLM to generate responses multiple times for each condition. For each generation, we constrained the word count to align with the Origin responses and computed the conditional entropy of the generated response. 
We repeated this process iteratively, fine-tuning the constraints until the conditional entropy of the responses matched that of the Origin responses within a tolerance of $\pm0.005$.

\subsection{Case Study for Desirability  Prompt }
\label{sec:desircasestudy}
Here, we provide a case study comparing the responses generated by the "Origin" (Table~\ref{tab:originresponse}) and "Logical" (Table~\ref{tab:logicalresponse}) prompts in Section~\ref{sec:desirability-Influence-WR}. This case study demonstrates that the Logical prompt indeed produces responses that are more coherent and logically structured.

\begin{table*}[tbh!]
    \caption{
    Case Study Comparison: Origin Response.
    }
    \centering
    \small
    \begin{tabular}{p{2.5cm}p{10.5cm}}
    \toprule
    \textbf{Instruction} & "How to improve instance segmentation AP metrics with human knowledge" \\
    \midrule
    \textbf{Origin Response} & 

\# Improving the Average Precision (AP) Metric for Instance Segmentation

Improving the Average Precision (AP) metric for instance segmentation with human knowledge involves leveraging domain expertise and data curation techniques to refine the model's performance. Here are several strategies to consider:

\#\# 1. **Data Quality and Annotation**
- Ensure high-quality, accurate annotations. Human experts should meticulously label the instances in the training data.
- Include challenging cases in the dataset that the model struggles with, such as occlusions, varying scales, and uncommon poses or appearances.
- Review and correct annotations regularly to prevent propagating errors through the model.

\#\# 2. **Class Balance and Diversity**
- Balance the dataset so that all classes are well-represented, avoiding class imbalance that can bias the model.
- Introduce diversity in the training data, including different lighting conditions, backgrounds, and object orientations.

\#\# 3. **Domain-Specific Augmentations**
- Apply data augmentation techniques informed by human knowledge about the domain, such as realistic transformations, to make the model more robust to variations in new data.

\#\# 4. **Incorporate Expert Knowledge into Features**
- Design or select features that are relevant to the task, possibly guided by domain experts who understand the nuances of the data.
- Use human expertise to create rules or heuristics that can aid the model, especially in edge cases where data is scarce.

\#\# 5. **Model Architecture Customization**
- Adapt or design neural network architectures that are better suited for the specific characteristics of the data, possibly with input from domain experts.
- Use architectures that have proven successful in similar tasks or domains.

\#\# 6. **Transfer Learning**
- Apply transfer learning by starting with a model pre-trained on a large, diverse dataset and fine-tuning it on your specific dataset with human-verified labels.

\#\# 7. **Active Learning**
- Use an active learning loop where the model is trained on a subset of data, then human experts annotate the most informative samples based on the model's uncertainty.

\#\# 8. **Error Analysis**
- Perform thorough error analysis with experts to understand the types of mistakes the model is making and address them specifically through training data improvements or model adjustments.

\#\# 9. **Ensemble Methods**
- Combine predictions from multiple models (ensemble) to improve overall performance. Experts can help select models that complement each other's strengths and weaknesses.

\#\# 10. **Curriculum Learning**
- Implement curriculum learning strategies where the model is first trained on easier examples and progressively exposed to more complex cases, guided by human understanding of the task difficulty hierarchy.

\#\# 11. **Semantic Guidance**
- Introduce additional semantic information into the model, such as object relationships or scene context, which might be obvious to humans but not inherently learned by the model.

\#\# 12. **Post-Processing**
- Use expert knowledge to design post-processing steps that can refine the model's raw output, such as graph-based optimization or conditional random fields (CRFs).

\#\# 13. **Human-in-the-Loop**
- Implement a human-in-the-loop system where difficult cases are flagged for human review and correction, and the model is continuously updated with this feedback.

---

By combining machine learning techniques with human expertise, you can iteratively improve the instance segmentation model's performance and achieve higher AP metrics. Remember that improving AP is often an iterative process of hypothesis, testing, and refinement.

    \\
    \bottomrule
    \end{tabular}
\label{tab:originresponse}
\end{table*}

\begin{table*}[tbh!]
    \caption{
    Case Study Comparison: Logical Response.
    }
    \centering
    \small
    \begin{tabular}{p{2.5cm}p{10.5cm}}
    \toprule
    \textbf{Instruction} & "How to improve instance segmentation AP metrics with human knowledge" \\
    \midrule

    \textbf{Logical Response} & 
\# Improving Instance Segmentation AP Metrics with Human Knowledge

Improving instance segmentation Average Precision (AP) metrics through the integration of human knowledge requires a strategic blend of domain expertise, algorithmic enhancements, and fine-tuning of model processes. Below is a structured approach to achieve this improvement.

---

\#\# 1. Understanding Instance Segmentation AP
Instance segmentation AP measures how effectively a model detects and delineates individual objects within an image. It is computed by comparing predicted segmentation masks against ground truth at various Intersection over Union (IoU) thresholds, capturing both precision and recall. Common factors that degrade AP include inaccurate boundaries (leading to low IoU), overlapping objects that the model confuses, or a high frequency of false positives and negatives.

---

\#\# 2. Role of Human Knowledge
Human expertise can fill gaps that purely data-driven approaches may miss. Experts can identify subtle yet crucial details—such as fine edges in biomedical imaging or nuanced object shapes in autonomous driving scenarios—that generic datasets or fully automated processes might overlook. By embedding this human knowledge into the data lifecycle, models become more robust, accurate, and better aligned with real-world conditions.

---

\#\# 3. Strategies to Leverage Human Knowledge

\#\#\# A. Data Annotation and Augmentation
1. **Improved Labeling Quality**  
   Expert annotators meticulously outline object boundaries and resolve ambiguities in crowded scenes. High-quality annotations directly translate into higher-quality model predictions.
2. **Domain-Specific Data**  
   Curate datasets that accurately reflect target scenarios. In medical imaging, for instance, radiologists or pathologists can ensure that rare but critical cases are included, preventing gaps in training.
3. **Augmentation with Expert Context**  
   Beyond basic flips and rotations, incorporate domain-relevant transformations, such as controlled brightness shifts for nighttime driving or synthetic noise to simulate real-world sensor data.

\#\#\# B. Model Design and Training
1. **Loss Function Customization**  
   Adapt loss functions to emphasize IoU-sensitive metrics, for example, through focal loss or boundary-aware losses. This alignment ensures that improvements in segmentation boundaries lead to tangible AP gains.
2. **Pretraining on Similar Domains**  
   Leverage pretrained models from closely related tasks, allowing experts to guide architecture selection. This shortcut often yields faster convergence and higher final accuracy.
3. **Human-Guided Active Learning**  
   Experts can identify misclassified or underrepresented samples. Iteratively retraining the model on these challenging instances refines its ability to generalize to complex scenarios.

\#\#\# C. Post-Processing Enhancements
1. **Refinement Using Rules**  
   Introduce domain-specific heuristics to remove implausible predictions. For example, discard detections that violate known object size constraints or positional rules.
2. **Combination with Traditional Methods**  
   Fuse deep learning outputs with classical techniques (e.g., edge detection, shape constraints) that human experts have validated over time. Such hybrid approaches can further refine object boundaries and minimize false positives.

---

\#\# 4. Iterative Feedback Loops
Establish a continuous feedback mechanism. Domain experts periodically review model outputs to identify systemic errors. These insights drive subsequent annotation improvements and retraining cycles, fostering a more resilient system over time.

---

\#\# Conclusion
By strategically integrating human expertise throughout the data pipeline, from annotation to post-processing, instance segmentation models can achieve more accurate and reliable results. This synergy of human insight and machine learning algorithms is pivotal for addressing complex or domain-specific challenges, ultimately elevating AP scores and ensuring robust performance in diverse real-world applications.
We compare the GPT-4 “Origin” answer with the “Logical” answer. From this comparison, it is clear that the “Logical” approach yields a more coherent and logically consistent response.
    \\
    \bottomrule
    \end{tabular}
\label{tab:logicalresponse}
\end{table*}

\subsection{Word Count Statistics of Figure~\ref{fig:GPT4ConditionalInformation}}  
\label{sec:word_counts}

To facilitate a clearer interpretation of the differences among Concise, Origin, and Detailed responses in Figure~\ref{fig:GPT4ConditionalInformation}, we present the average word counts for each response style across various datasets in Table~\ref{tab:word_counts_figure3}. 

\begin{table*}[h!]
    \centering
    \small
    \caption{Average Word Counts for Concise, Origin, and Detailed Responses across Different Datasets.}
    \begin{tabular}{lcccccc}
    \toprule
    \textbf{Response Style} & \textbf{AlpacaEval} & \textbf{LIMA} & \textbf{Koala} & \textbf{Self-Instruct} & \textbf{Vicuna} & \textbf{WizardLM} \\
    \midrule
    Concise & 112.27  & 83.93  & 187.44  & 81.29  & 44.55  & 199.47 \\
    Origin  & 372.73  & 325.00 & 345.23  & 170.45 & 379.30 & 327.20 \\
    Detailed & 488.49 & 416.65 & 423.64  & 239.49 & 462.99 & 423.00 \\
    \bottomrule
    \end{tabular}
\label{tab:word_counts_figure3}
\end{table*}

\subsection{ Information About Use Of AI Assistants}
We use GPT-4 as an AI assistant during the preparation of this manuscript.

\subsection{Conditional Entropy Calculation}  
\label{sec:conditional-entropy}  

In our experiments, conditional entropy serves as a crucial metric for measuring and ensuring consistent information mass across different response conditions. We calculate conditional entropy following the definition used in prior work on quantum information theory~\citep{wei2024large, von1932mathematische}. Specifically, the conditional matrix entropy of response $Z$ given the instruction $X$ is expressed as $H(Z \mid X) = H(X, Z) - H(X)$. Here, $H(X,Z) = - \mathrm{tr}(\Sigma_{XZ}\,\log \Sigma_{XZ})$ represents the joint entropy of $X$ and $Z$, while $H(X) = - \mathrm{tr}(\tilde{\Sigma}_X \log \tilde{\Sigma}_X)$ represents the entropy of $X$. To ensure proper normalization, we define $\tilde{\Sigma}_X = \Sigma_{XX} / \mathrm{tr}(\Sigma_{XX})$, where $\Sigma_{XX}$ is the covariance matrix of $X$, normalized to have a trace of 1. This normalization step allows for consistent entropy calculations across different conditions, ensuring a fair comparison of conditional entropy values. 
To compute the joint covariance matrix $\Sigma_{XZ}$, we first concatenate each instruction $x_i \in \mathbb{R}^{d_X}$ and response $z_i \in \mathbb{R}^{d_Z}$ into a single vector $\mathbf{u}_i = \begin{bmatrix} x_i \\ z_i \end{bmatrix} \in \mathbb{R}^{d_X + d_Z}$. We then compute the mean embedding $\bar{\mathbf{u}} = \frac{1}{N} \sum_{i=1}^N \mathbf{u}_i$ and normalize each vector by centering and scaling it as $\bar{\mathbf{u}}_i = \frac{\mathbf{u}_i - \bar{\mathbf{u}}}{\|\mathbf{u}_i - \bar{\mathbf{u}}\|_2}$. Finally, the joint covariance matrix is obtained as $\Sigma_{XZ} = \frac{1}{N} \sum_{i=1}^N \bar{\mathbf{u}}_i \bar{\mathbf{u}}_i^\top$. The resulting covariance matrix $\Sigma_{XZ}$ can be expressed as a block matrix of the form 
\[
\Sigma_{XZ} = \begin{bmatrix} \Sigma_{XX} & \Sigma_{XZ}^{(\text{off‐diag})} \\ \Sigma_{ZX}^{(\text{off‐diag})} & \Sigma_{ZZ} \end{bmatrix},
\]
where $\Sigma_{XX}$ corresponds to the block associated with $X$.

\section{Dataset}

\paragraph{AlpacaEval}~\citep{dubois2024alpacafarm} comprises 805 instructions, including 252 from the self-instruct test set~\citep{wang2022self}, 188 from the Open Assistant (OASST) test set, 129 from Anthropic's helpful test set~\citep{zhou2023lima}, 80 from the Vicuna test set~\citep{vicuna2023}, and 156 from the Koala test set~\citep{vu2023koala}.

\paragraph{LIMA}~\citep{zhou2023lima} compiles a training dataset of 1000 prompts and responses, designed to ensure stylistic consistency in outputs while maintaining diverse inputs. It also provides an open-source test set of 300 prompts and a development set of 50. The dataset is sourced from a variety of platforms, mainly community Q\&A websites such as Stack Exchange, wikiHow, and the Pushshift Reddit Dataset~\citep{baumgartner2020pushshift}, along with manually curated examples. Within these Q\&A communities, highly upvoted answers on Reddit often have a humorous or trolling tone, requiring extra effort to align them with the intended helpful chat assistant style. In contrast, responses from Stack Exchange and wikiHow naturally align with this style. The inclusion of human-authored examples further enhances the dataset's diversity. Our research specifically utilizes the test set from the LIMA dataset to evaluate our models.

\paragraph{Vicuna}~\citep{vicuna2023} divides 80 test instructions into eight distinct categories: Fermi problems, commonsense, roleplay scenarios, coding/math/writing tasks, counterfactuals, knowledge, and generic questions. This categorization is intended to thoroughly evaluate multiple aspects of a chatbot's performance. Prior research indicates that the Vicuna dataset generally includes instructions of lower difficulty and complexity~\citep{xu2023wizardlm}. In our study, we used the Vicuna test set to specifically evaluate the performance of large language models across these varied instruction categories.

\paragraph{Self-Instruct}~\citep{wang2022self} consists of 252 human-created test instructions, each associated with a carefully designed output. This test set is curated to reflect the real-world applicability of instruction-following models, covering a broad spectrum of domains including email composition, social media, productivity software, and coding. The test instructions vary in style and format, incorporating different task lengths and diverse input/output types such as bullet lists, tables, code snippets, and mathematical equations. We employed the Self-Instruct test set in our research to rigorously assess our model's capability to comply with precise instructions across these varied domains.

\paragraph{Wizardlm}~\citep{xu2023wizardlm} comprises a training set of 70k examples with varied complexities, initiated from 52k instructional data provided by Alpaca. Following $M=4$ evolutionary cycles, the collection expands to 250k instructions. In each cycle, from the six newly generated prompts—five via in-depth evolution and one through in-breadth evolution—one is chosen randomly for each instruction. ChatGPT then generates responses, resulting in $52 \times 4 \times 3 = 624\text{k}$ instruction-response pairs. The training subset selected for the Evol-Instruct dataset contains 70k of these instructions. The test set, which includes 218 instructions, is sourced from a variety of platforms such as open-source projects and online forums, encapsulating 29 unique skills identified from authentic human tasks. These skills range from Coding Generation \& Debugging to Reasoning, Mathematics, Writing, Handling Complex Formats, and Mastery over Extensive Disciplines. In our study, we utilized the Wizardlm test set to thoroughly evaluate our model's ability to adhere to detailed instructions.

\paragraph{Koala}~\citep{vu2023koala} consists of 180 authentic user queries obtained from the Internet. These queries cover a diverse array of topics and are generally characterized by a conversational tone, underscoring their applicability to real-world chat-based applications. To prevent test-set leakage, we exclude any query that achieves a BLEU score over 20\% when compared to examples from our training set. Furthermore, we do not consider queries related to programming or non-English languages, as the capabilities of our crowd-sourced raters—who form our evaluation team—do not extend to effectively assessing such content. We have exclusively utilized the Koala test set to assess our model's capability to process and respond to genuine user inquiries in a conversational setting.

\section{Discussion on Length Bias}
\label{sec:diss-lgb}
In this section, we discuss length bias from two perspectives: \emph{model output} and \emph{evaluation}.
From a \emph{model-output perspective}, generating sufficiently detailed responses can indeed enhance user satisfaction by providing relevant, in-depth information. Length alone is not inherently detrimental if it serves a clear purpose: a well-reasoned, thorough response can be beneficial for clarity and user engagement. However, verbosity that merely inflates the text without offering meaningful content can degrade readability and detract from the user experience. 
As shown in Table~\ref{table:humanstudy}, responses with Quality Enhancement achieve  higher human win rates and  win rates.
From an \emph{evaluation perspective}, length bias poses a risk of distorting model performance comparisons by rewarding verbosity over precision. 
This concern is evident in benchmarks such as AlpacaEval~\cite{alpaca_eval}, where longer outputs can artificially inflate perceived quality, even when additional content is only marginally relevant. To address this issue, metrics such as LCWR~\citep{dubois2024length} normalize output length to offer a more equitable assessment of content quality. 
Our proposed \textit{AdapAlpaca} further adopts a proactive stance by mitigating length bias early in the evaluation pipeline, thus ensuring that win rates reflect genuine improvements in response quality rather than inflated text length.

\section{Human Evaluation Process}
\label{sec:humanstudy}

\subsection{Implementation Details}
To ensure the robustness of our findings and complement the automated evaluations, a thorough human evaluation was conducted.
The human evaluation involved 25 participants, all of whom are professionals or researchers in the tech industry with specific expertise in language models. These individuals were carefully selected to represent a broad spectrum of perspectives and expertise levels, ranging from early-career to senior researchers. Each participant was assigned randomly to different segments of the dataset to ensure a balanced and unbiased input across all items evaluated.
The dataset, comprising 805 responses generated for each prompt and compared against a default reference, was strategically divided into eight distinct parts, each containing approximately 100 responses. This division was structured to facilitate manageability and focus during the evaluation process. By dividing the dataset into smaller, more manageable segments, we aimed to optimize the evaluation process without overwhelming the evaluators, thus maintaining a high standard of analysis quality.
Each of these eight segments was then randomly assigned to five different participants. This approach ensured that every subset of the dataset was evaluated by multiple individuals, enhancing the reliability and diversity of perspectives in the assessment process. Random assignment of participants to each segment helped minimize any potential bias, providing a balanced evaluation across all parts of the dataset.
This method of segmenting the data and assigning evaluators ensured that each response received sufficient attention, contributing to the robustness and credibility of the evaluation results. By implementing this straightforward and strategic approach to data handling and evaluator assignment, we maintained a high standard of reliability and fairness throughout the evaluation process.
The evaluation was facilitated using a custom-built interface on Gradio~\footnote{\url{https://github.com/gradio-app/gradio}}, an open platform known for its robustness in sharing interactive machine learning models. Detailed instructions were provided to each participant to minimize user error and bias. The interface displayed questions along with two model outputs side-by-side, labeled "Left" and "Right," with their positions randomized to prevent positional bias. Figure~\ref{fig:gradio_interface} illustrates this setup. 
This comprehensive human evaluation process not only validated the effectiveness of our proposed methodologies but also provided critical insights that significantly enriched our understanding of automated metric evaluations.

\subsection{Inter-Rater Consistency Analysis}
\label{sec:inter-rater-consistency}

To assess the consistency of human evaluations across different raters, we conducted an inter-rater reliability analysis using Fleiss’ Kappa, a standard metric for evaluating multi-rater categorical agreement. We illustrate this with the human study conducted under the Concise prompt setting of AdapAlpaca. In this study, each segment was evaluated by five raters, and the Kappa values for each segment are presented in Table~\ref{table:kappa-scores}. 

\begin{table}[h]
\caption{
Fleiss’ Kappa values for inter-rater agreement on different segments in the Concise prompt setting of AdapAlpaca. Higher Kappa indicates stronger consensus among the five raters assigned to each segment.
}
\centering
\scalebox{0.85}{
\begin{tabular}{@{}lccc@{}}
\toprule
\textbf{Segment ID} & \textbf{\#Responses} & \textbf{\#Raters} & \textbf{Fleiss' Kappa} \\ 
\midrule
Segment 1 & 101 & 5 & 0.72 \\ 
Segment 2 & 101 & 5 & 0.69 \\ 
Segment 3 & 101 & 5 & 0.70 \\ 
Segment 4 & 101 & 5 & 0.66 \\ 
Segment 5 & 101 & 5 & 0.74 \\ 
Segment 6 & 100 & 5 & 0.71 \\ 
Segment 7 & 100 & 5 & 0.68 \\ 
Segment 8 & 100 & 5 & 0.73 \\ 
\bottomrule
\end{tabular}
}
\label{table:kappa-scores}
\end{table}

The Kappa values range from 0.66 to 0.74, indicating moderate to substantial agreement across raters. These results demonstrate that, despite different segments being allocated to different groups of participants, the raters assigned to each segment generally reached a consistent judgment. This level of agreement supports the reliability and robustness of our human evaluation process. By maintaining a high degree of consensus across raters, we ensure that subjective biases are minimized and that the evaluation results reflect an objective and reproducible assessment of response quality.

\section{Addressing Confounding Factors in Evaluation Metrics}
\label{sec:discuss-length-bias}

Evaluating the quality of responses solely based on win rates can be misleading when external factors, such as response length, unduly influence outcomes. Longer responses may appear more informative or thorough, thereby increasing the likelihood of higher win rates, even when the additional content consists of redundant or irrelevant information. This phenomenon has been noted in widely used benchmarks like AlpacaEval~\citep{alpaca_eval}, where verbosity can artificially inflate perceived quality by virtue of longer outputs. Consequently, it creates an unintended bias, favoring models that produce lengthier responses over those that prioritize clarity and precision.
To address this issue, various strategies have been proposed to debias response length. For example, \emph{Length-Controlled Win Rate} (LCWR)~\citep{dubois2024length} ensures that win rates are assessed under similar length conditions, thereby mitigating the inadvertent benefit conferred by verbosity. Our work adopts a similar principle through the design of \textit{AdapAlpaca}, which seeks to remove length bias from the outset rather than correcting for it post hoc. By focusing on the core qualities of content—such as relevance, accuracy, and coherence—rather than conflating these qualities with response length, the evaluation better reflects genuine human preferences. Ultimately, a reliable evaluation metric should measure content quality without being confounded by extraneous factors. Ensuring that both length and other potential confounders (e.g., formatting quirks, stylistic flourishes) do not overshadow intrinsic content quality is essential for fair model comparison and for fostering genuine advancements in natural language generation.

\section{Design Rationale for Desirability and Information Mass}
\label{sec:design-rationale}

The concepts of Desirability and Information Mass were deliberately designed based on foundational principles in response evaluation to distinguish the trustworthiness of content from the quantity of information provided. These metrics were developed not through purely manual heuristics but rather through a combination of theoretical reasoning and empirical validation. Specifically, Desirability captures qualitative aspects of a response, such as coherence, logical consistency, and correctness, while Information Mass focuses on quantitative elements like completeness and level of detail. This distinction ensures a more comprehensive and balanced evaluation of response quality.
A core motivation for separating Desirability and Information Mass is that certain trustworthiness attributes, such as toxicity, are fundamentally independent of response length or information density. For example, a short sentence can be as toxic as a much longer one. Toxicity relates to the emotional and semantic impact of language on humans, rather than the number of words or sentence length. Consider the following examples:
Short toxic sentence: “This person is stupid.”
Long toxic sentence: “This person is so unbelievably incompetent and ignorant that it’s embarrassing to even be in the same conversation as them. People like this are the reason nothing ever gets done properly—they’re a complete waste of space and time.”
In both cases, the content is toxic due to its demeaning and hostile nature, regardless of sentence length. Therefore, attributes such as toxicity cannot be quantified based on word count or sentence length and instead depend on the semantic content and emotional tone. This independence from length underscores why Desirability and Information Mass need to be treated as distinct dimensions of response evaluation.
Empirically, we validated the independence of Desirability and Information Mass through controlled experiments in 
Section~\ref{sec:desirability-Influence-WR}  and 
 Section~\ref{sec:infoMass-Influence-WR}. Our results show that changes in correctness and logical consistency (Desirability) significantly influence win rates even when response length remains constant. Conversely, variations in meaningful content (Information Mass) also affect win rates but are independent of trustworthiness and logical correctness. These findings confirm that Desirability and Information Mass represent distinct yet complementary aspects of response quality.

\section{Potential Negative Societal Impacts}
\label{sec:NegativeImpacts}
While this research contributes to reducing bias in language model evaluations, it is important to consider potential indirect societal impacts that might arise:

\paragraph{Dependence on Automated Decision-Making.}  
This study's focus on enhancing the accuracy of automated evaluations may inadvertently promote an over-reliance on AI-driven decision-making processes. While beneficial in many respects, such reliance could diminish the value placed on human judgment and intuition in areas where nuanced understanding and ethical considerations are paramount.

\paragraph{Perception and Trust in AI.} 
By highlighting the capabilities and improvements in AI evaluations, there might be an overestimation of AI reliability and fairness among the public and policymakers. This could lead to misplaced trust in AI systems, overlooking their limitations and the necessity for continuous oversight and human intervention.

\section{Preference Dataset and Reward Models}
\label{sec:preferData-RewardModel}

In this appendix, we provide detailed information about the preference datasets and reward models used in Section~\ref{sec:lengthbiasRLHF}.

\subsection{Preference Datasets}
\textbf{AnthropicHH}~\footnote{\url{https://huggingface.co/datasets/Unified-Language-Model-Alignment/Anthropic_HH_Golden}}: The AnthropicHH dataset evaluates the ULMA technique by replacing positive samples in a preference dataset with high-quality 'golden' data from GPT-4, aiming to enhance alignment methods like RLHF, DPO, and ULMA.

\noindent \textbf{HC3}~\footnote{\url{https://huggingface.co/datasets/Hello-SimpleAI/HC3}}: The HC3 dataset, presented in "How Close is ChatGPT to Human Experts? Comparison Corpus, Evaluation, and Detection," offers a pioneering human-ChatGPT comparison corpus. It enables nuanced evaluations of ChatGPT's performance and its closeness to human expert outputs.

\noindent \textbf{SHP}~\footnote{\url{https://huggingface.co/datasets/stanfordnlp/SHP}}: The SHP dataset, from the Stanford Human Preferences project, collects 385K human preferences across 18 subject areas, utilizing naturally occurring human-written responses on Reddit to enhance RLHF reward models and NLG evaluation. This dataset emphasizes the utility of response helpfulness over harm reduction.

\noindent \textbf{WebgptCom}~\footnote{\url{https://huggingface.co/datasets/openai/webgpt_comparisons}}: The WebgptCom dataset comprises 19,578 comparisons from the WebGPT project, designed for reward modeling. It features pairs of model-generated answers to questions, each scored by humans to determine preference, supporting the training of a long-form question answering model aligned with human preferences.
    
\noindent \textbf{Helpsteer}~\footnote{\url{https://huggingface.co/datasets/RLHFlow/Helpsteer-preference-standard}}: The Helpsteer dataset, utilized for refining reward models in conversational AI, includes preference data distinguishing helpful from unhelpful responses. It consists of paired entries labeled as 'chosen' and 'rejected', with respective scores reflecting their utility. The dataset includes 37,131 examples in the training split, emphasizing its scale for robust model training.

\noindent \textbf{Preference700K}~\footnote{\url{https://huggingface.co/datasets/hendrydong/preference_700K}}: The Preference700K dataset comprises 700,000 preference comparisons between two conversational responses, 'chosen' and 'rejected', related to the same prompt. This large-scale dataset is structured to train and evaluate models on their ability to discern more favorable conversational outcomes based on user interaction dynamics.

\noindent \textbf{PreMix}~\footnote{\url{https://huggingface.co/datasets/weqweasdas/preference_dataset_mix2}}: The PreMix dataset features 528,029 comparisons from preprocessed preference datasets, focusing on dialogues structured with a 'chosen' and 'rejected' response based on the same prompt. This dataset aids in training models to discern the more favorable responses in conversational settings.

\noindent \textbf{Ultrafeedback}~\footnote{\url{https://huggingface.co/datasets/argilla/ultrafeedback-binarized-preferences-cleaned}}: Ultrafeedback is an improved version of the original dataset, now cleaned and binarized using average preference ratings. It eliminates problematic data from earlier versions, notably those influenced by the TruthfulQA dataset, and removes contributions from ShareGPT sources, ensuring cleaner and more reliable data for fine-tuning conversational AI on preference discernment.

\noindent \textbf{Argilla}~\footnote{\url{https://huggingface.co/datasets/csarron/argilla-ultrafeedback-binarized-preferences-cleaned}}: The Argilla dataset is a refined version of the UltraFeedback dataset, used to train the Zephyr-7B-$\beta$ model. This dataset features 64k prompts with binarized completions, categorizing the highest scored as 'chosen' and one of the remaining as 'rejected'. It supports various training techniques including supervised fine-tuning, preference modeling for reward systems, and generation techniques like rejection sampling.
    
\noindent \textbf{Lmsys}~\footnote{\url{https://huggingface.co/datasets/lmsys/lmsys-arena-human-preference-55k}}: The Policy1.4b dataset incorporates labels from the AlpacaFarm dataset and utilizes generated answers from a 1.4 billion parameter Pythia policy model. Responses are evaluated using the 'reward-model-human' as a gold standard. This dataset is pivotal for refining AI policy models through precise human preference feedback.
    
\noindent \textbf{Policy1.4b}~\footnote{\url{https://huggingface.co/datasets/tlc4418/1.4b-policy_preference_data_gold_labelled}}: The Prometheus2 dataset, transformed from the "prometheus-eval/Preference-Collection", is crafted to enhance fine-grained evaluation capabilities in language models. This dataset pairs instructions with two responses, scored and chosen based on preference, facilitating nuanced evaluation and comparison aligned with human judgment.
    
\noindent \textbf{Prometheus2}~\footnote{\url{https://huggingface.co/datasets/RLHFlow/Prometheus2-preference-standard}}: The Prometheus2 dataset, transformed from the "prometheus-eval/Preference-Collection", is crafted to enhance fine-grained evaluation capabilities in language models. This dataset pairs instructions with two responses, scored and chosen based on preference, facilitating nuanced evaluation and comparison aligned with human judgment.

\noindent \textbf{Combined}~\footnote{\url{https://huggingface.co/datasets/yoonholee/combined-preference-dataset}}: The Combined dataset integrates multiple preference datasets into a unified resource, all examples binarized and standardized. It aggregates data from diverse sources to create a comprehensive set for training and evaluating language models on preference understanding.

\noindent \textbf{ElixLatent}~\footnote{\url{https://huggingface.co/datasets/Asap7772/elix_latent_preferences_gpt4}}: The ElixLatent dataset, designed around GPT-4, serves as a resource for training and evaluating latent preference modeling. It provides pairs of latent responses ('yw' and 'yl') and their corresponding contexts ('x'), allowing researchers to explore the nuances of preference dynamics in generated text.

\subsection{Reward Models}

\noindent \textbf{Eurus}~\footnote{\url{https://huggingface.co/openbmb/Eurus-RM-7b}}: Eurus is a reward model trained on UltraInteract, UltraFeedback, and UltraSafety datasets. It excels in complex reasoning tasks and outperforms larger models, including GPT-4, by significantly enhancing language models' reasoning capabilities.

\noindent \textbf{Grmdis}~\footnote{\url{https://huggingface.co/Ray2333/GRM-llama3-8B-distill}}:  Generalizable Reward Model (GRM), uses hidden state regularization to enhance generalization in reward models for large language models (LLMs). Initially built on fixed weights from a Llama-3-based model and fine-tuned only on a reward head, it significantly improves on standard benchmarks, demonstrating enhanced reasoning and safety metrics over existing models.

\noindent \textbf{Grmsft}~\footnote{\url{https://huggingface.co/Ray2333/GRM-llama3-8B-sftreg}}: It is part of the Generalizable Reward Model (GRM) series, aimed at enhancing LLMs through hidden state regularization. 
It excels across various complex evaluative tasks, outperforming other high-capacity models in reasoning and safety.

\noindent \textbf{UniF}~\footnote{\url{https://huggingface.co/Ray2333/reward-model-Mistral-7B-instruct-Unified-Feedback}}: It is a reward model finetuned on the 'llm-blender/Unified-Feedback' dataset using the Mistral-7B-Instruct architecture. Achieving an accuracy of 0.7740 on test sets, it excels at modeling human preferences. The model integrates diverse preference data from multiple sources, enhancing its applicability in aligning LLMs to human judgments across various conversational contexts.

\noindent \textbf{Debba}~\footnote{\url{https://huggingface.co/OpenAssistant/reward-model-deberta-v3-base}}: Debba is a reward model utilizing Deberta-v3-base, trained to evaluate QA models and serve as a reward mechanism in RLHF by predicting which generated answer aligns better with human judgment. It is trained on datasets such as webgpt\_comparisons, summarize\_from\_feedback, and synthetic-instruct-gptj-pairwise, ensuring a consistent validation approach across varying domains.
    
\noindent \textbf{Bebla}~\footnote{\url{https://huggingface.co/OpenAssistant/reward-model-deberta-v3-large}}: It is a reward model trained to assess the quality of responses in QA evaluations and to provide scoring in RLHF. It was developed with datasets such as webgpt\_comparisons, summarize\_from\_feedback, and synthetic-instruct-gptj-pairwise, ensuring it can reliably predict human preferences across diverse contexts.

\noindent \textbf{FsfairRM}~\footnote{\url{https://huggingface.co/sfairXC/FsfairX-LLaMA3-RM-v0.1}}: It is designed for RLHF applications including PPO, iterative SFT, and iterative DPO. This state-of-the-art reward model is licensed under PKU-Alignment/PKU-SafeRLHF-30K, demonstrating high performance across diverse metrics like chat, safety, and reasoning in Reward-Bench.

\noindent \textbf{Gerew}~\footnote{\url{https://huggingface.co/Ray2333/Gemma-2B-rewardmodel-baseline}}: It is trained using BT loss on the weqweasdas/preference\_dataset\_mixture2\_and\_safe\_pku dataset. This model is designed for efficiently evaluating and aligning LLMs, offering a baseline performance that is well-suited for smaller-scale applications requiring rapid assessment of language model outputs.

\noindent \textbf{Misrmr}~\footnote{\url{https://huggingface.co/hendrydong/Mistral-RM-for-RAFT-GSHF-v0}}: It is a reward model tailored for iterative Synthetic Frontier Tuning (SFT) and Dynamic Policy Optimization (DPO). Trained to enhance language generation tasks, it supports fine-grained reward modeling to improve the alignment and efficacy of language models in diverse applications.

\noindent \textbf{InteRM}~\footnote{\url{https://huggingface.co/internlm/internlm2-1_8b-reward}}: It is a reward model trained on the foundation of InternLM2-Chat-1.8B-SFT. This model has been trained using over 2.4 million preference samples, both human-annotated and AI-synthesized, achieving outstanding performance while ensuring a balance between helpful and harmless.

\section{Human study with More Model}
\label{sec: More-Hmresult}
To provide a more comprehensive view of our human evaluation study, we conducted experiments on more LLMs, including Llama3-70B and Qwen1.5-72B. The results are summarized in Tables~\ref{table:llama3humanstudy} and~\ref{table:qwen1_5humanstudy}.
These results further validate \OURS{} as a robust metric for aligning model evaluations with human preferences, effectively addressing the shortcomings of LCWR.

\begin{table*}[t]
\caption{
Comparison of five quantitative metrics related to quality: Vocabulary Size, Win Rate relative to AlpacaEval (AlpacaWR), Entropy, Inter-sample N-gram Frequency (INGF), and Word Counts.
}
\centering
\scalebox{0.9}{
\begin{tabular}{@{}lcccccccc@{}}
\toprule
\multirow{2}{*}{Interval}                & \multicolumn{2}{c}{Vocabulary Size} & \multicolumn{2}{c}{AlpacaWR} & \multicolumn{2}{c}{Entropy} & \multirow{2}{*}{INGF} & \multirow{2}{*}{Word Counts} \\ \cmidrule(r){2-3}  \cmidrule(r){4-5} \cmidrule(r){6-7}
                & All             & Ans Avg.          & WR           & LCWR         & All         & Ans Avg.      &                       &                              \\ \midrule
AlpacAns Origin  & 38474           & 47.79             & 50.00        & 50.00        & 408.83      & 0.5686        & 7376.92               & 363.85                       \\
\OURS{}-200           & 22612           & 28.08             & 20.73        & 43.81        & 363.55      & 0.5056        & 1618.69               & 145.72                       \\
\OURS{}-400         & 36943           & 45.89             & 47.34        & 47.40        & 414.39      & 0.5763        & 6003.87               & 355.20                       \\
\OURS{}-600         & 47691           & 59.24             & 62.58        & 50.97        & 434.77      & 0.6046        & 9086.01               & 540.95                       \\
\OURS{}-800         & 55362           & 68.77             & 71.20        & 54.31        & 447.48      & 0.6223        & 10320.11              & 708.36                       \\
\OURS{}-1000        & 66095           & 82.10             & 66.98        & 36.24        & 456.32      & 0.6346        & 10981.84              & 913.44                       \\ \bottomrule
\end{tabular}

}
\label{table:analysis}
\end{table*}

\begin{table}[t]
\caption{
The subscripts in the LCWR and WR columns indicate the differences between these metrics and the corresponding Human WR. A larger absolute value denotes a greater disparity between the annotator's evaluation and Human Preference. "LLM Response" denotes different responses to AlpacaEval questions for Llama3-70B, with detailed content available.
}
\centering
\scalebox{0.6}{
\begin{tabular}{@{}lccccc@{}}
\toprule
\multirow{2}{*}{LLM Response}               & \multicolumn{3}{c}{AlpacaEval}           & \multicolumn{2}{c}{AdapAlpaca} \\ \cmidrule(l){2-4} \cmidrule(l){5-6} 
                    & Human & LCWR  & WR  & Human  & WR  \\ \midrule
Concise             & 5.67  & $25.73_{+20.06}$ & $11.10_{+5.43}$  & 7.24     & $6.10_{-1.14}$      \\ \midrule
Detailed            & 46.12 & $38.62_{-7.50}$  & $50.80_{+4.68}$  & 41.99    & $42.98_{+0.99}$     \\ \midrule
Quality Enhancement & 53.48 & $42.59_{-10.89}$ & $56.50_{+3.02}$  & 51.63    & $50.89_{-0.74}$     \\ \bottomrule
\end{tabular}
}
\label{table:llama3humanstudy}
\end{table}

\begin{table}[t]
\caption{
The subscripts in the LCWR and WR columns indicate the differences between these metrics and the corresponding Human WR. A larger absolute value denotes a greater disparity between the annotator's evaluation and Human Preference. "LLM Response" denotes different responses to AlpacaEval questions for Qwen1.5-72B.
}
\centering
\scalebox{0.6}{
\begin{tabular}{@{}lccccc@{}}
\toprule
\multirow{2}{*}{LLM Response}               & \multicolumn{3}{c}{AlpacaEval}           & \multicolumn{2}{c}{\OURS{}} \\ \cmidrule(l){2-4} \cmidrule(l){5-6} 
                    & Human & LCWR  & WR  & Human  & WR  \\ \midrule
Concise             & 9.24  & $31.03_{+21.79}$ & $13.20_{+3.96}$  & 8.56     & $7.40_{-1.16}$     \\ \midrule
Detailed            & 45.52 & $38.50_{-7.02}$  & $42.70_{-2.82}$  & 39.97    & $38.92_{-1.05}$    \\ \midrule
Quality Enhancement & 46.61 & $40.62_{-5.99}$  & $48.87_{+2.26}$  & 43.18    & $44.01_{+0.83}$    \\ \bottomrule
\end{tabular}
}
\label{table:qwen1_5humanstudy}
\end{table}

\begin{table}[h]
\centering
\caption{Prompt for dataset generation, with \{max word\}-\{min word\} ranges set as 0-200, 200-400, 400-600, 600-800, and 800-1000.}
\begin{tabular}{m{0.88\linewidth}} 
\begin{tcolorbox}[mybox, title=Dataset generation prompt]
You are a helpful assistant, highly attentive to the specified token range required from user. Respond to the following question, your reply must only be within \{max word\}-\{min word\} words.
\end{tcolorbox}
\end{tabular}
\label{table:dataGen}
\end{table}

\section{Dataset Information}
\label{sec:dataInfo}
The data generation prompt, as outlined in Table~\ref{table:dataGen}, is carefully crafted to instruct GPT-4 to generate responses within predefined word limits. 
This prompt directed the model to generate content that is relevant to the given question and strictly adheres to the specified length constraints.

\subsection{Analysis of the Generated Data}
\label{sec:analysis}

The analysis is structured to quantify each dataset's basic characteristics, followed by a comparative assessment to identify any significant differences attributable to the varying response lengths. 
Table~\ref{table:analysis} presents a comprehensive overview, providing a snapshot of the informational content across different datasets. 
Specifically, it includes vocabulary size, inter-sample N-gram Frequency (INGF)~\citep{mishra2020dqi}, word counts of the generated dataset, win rate, length-controlled win rate, and entropy for AlpacaEval-Origin and \OURS{}-200, \OURS{}-400, \OURS{}-600, \OURS{}-800, and \OURS{}-1000.
Our findings indicate the following: 
1) Longer responses generally exhibit higher vocabulary sizes and word counts, suggesting a richer linguistic structure. 
2) The INGF metric reveals that while longer responses tend to include more common N-grams, there is significant variability in the types of N-grams used, indicating a creative and diverse use of language. 
3) Under Win Rate (WR) metrics, longer responses disproportionately receive higher preference scores due to their higher information mass. However, applying the length-controlled win rate (LCWR) significantly mitigates this bias, leading to a more balanced distribution of scores across different response lengths. 
This analysis aims to ascertain whether this phenomenon is intrinsic to the response quality or merely a byproduct of increased length. Our results demonstrate that although longer responses generally possess higher information mass, the quality of information, as measured by win rate, does not necessarily increase proportionally. Excessively lengthy responses can result in a decline in desirability, such as reduced consistency. For instance, in Table~\ref{table:analysis}, the win rate of \OURS{}-1000 is lower than that of \OURS{}-800.

\subsection{Dataset Documentations.}
The dataset comprises five JSON files for the \emph{\OURS{}-200}, \emph{\OURS{}-400}, \emph{\OURS{}-600}, \emph{\OURS{}-800}, and \emph{\OURS{}-1000}. Each file is generated using our length control prompt technique with the Alpaca dataset employing the GPT-4 1106 model.

Each data file contains a list of items with the following fields:
\begin{itemize}
    \item \texttt{instruction}: the prompt is given to generate the response.
    \item \texttt{generator}: identifies the model used.
    \item \texttt{dataset}: specifies the dataset used.
    \item \texttt{output\_word\_count}: the word count of the generated response.
    \item \texttt{output}: the actual text generated by the model.
\end{itemize}

\subsection{Intended Uses.}
The provided datasets, \emph{\OURS{}-200}, \emph{\OURS{}-400}, \emph{\OURS{}-600}, \emph{\OURS{}-800}, and \emph{\OURS{}-1000}, are specifically designed for researchers and practitioners in machine learning, natural language processing, and related fields. These datasets are intended to facilitate the evaluation of models that generate responses of similar lengths. They provide a standardized framework to repeatedly test and compare the performance of different models as detailed in our accompanying paper. This aims to ensure consistent evaluation and benchmarking of models under controlled conditions that mimic real-world application scenarios.

\section{Human Study on More Artifacts}
\label{sec:artifact-eval}

To ensure that length-controlled generation in \OURS{} does not introduce unintended artifacts that compromise content quality, we conduct a human evaluation study assessing the presence of three types of artifacts across different response lengths: \textit{confusion}, \textit{redundancy}, and \textit{dilution of meaning}. 
We design this study in response to the concern that length-conditioned generation may introduce semantic or stylistic shifts, potentially affecting the fidelity of the response beyond the scope of information mass. While our method does not forcibly compress or expand existing responses, it uses GPT-4 to regenerate answers that naturally fall within specific word-length intervals using carefully crafted prompts. This raises the question: Do these generations maintain high clarity, specificity, and non-redundancy, regardless of their length?
To assess this, we sample responses from five buckets of \OURS{}—\texttt{AdapAlpaca-200}, \texttt{AdapAlpaca-400}, \texttt{AdapAlpaca-600}, \texttt{AdapAlpaca-800}, and \texttt{AdapAlpaca-1000}—each containing responses targeting a distinct word count interval. For each bucket, we randomly sample 50 instruction-response pairs.
Each response is evaluated independently by five trained annotators along three dimensions, using a 5-point Likert scale (1 = best, 5 = worst):
(1) \textbf{Confusion.} Defined as the degree of unclear logic, ambiguous phrasing, or inconsistent reasoning within the response.
(2) \textbf{Redundancy.} Defined as the amount of unnecessary repetition, verbose restatements, or filler content that does not add informational value.
(3) \textbf{Dilution of Meaning.} Defined as the extent to which the response includes vague or generic content that weakens the informativeness or specificity.
Their Prompt can be found in Table~\ref{tab:other_detail_prompt}.
All annotators received detailed scoring rubrics and example-based guidelines for consistent judgment. The following average scores were computed across the 250 samples in each artifact category.
As shown in Table~\ref{tab:artifact-results}, all three metrics remain below 2.0 across all length intervals, indicating a low level of confusion, redundancy, and dilution of meaning—even for longer responses. While there is a mild upward trend as length increases (particularly in redundancy), the absolute magnitude of change is small and does not suggest significant degradation in response quality.
These results validate the design of our controlled generation setup: by guiding GPT-4 to produce responses that naturally fall into target length ranges, we preserve semantic coherence and content fidelity across varying information mass levels. We believe this supports the core assumption behind \OURS{}—that matched-length comparisons can be made fairly without introducing spurious artifacts in the responses themselves.

\begin{table}[h]
\centering
\footnotesize
\renewcommand{\arraystretch}{0.98}
\resizebox{0.95\linewidth}{!}{%
\begin{tabular}{lccc}
\toprule
\textbf{Length Bucket} & \textbf{Confusion} & \textbf{Redundancy} & {\textbf{Dilution}}\\
\midrule
AdapAlpaca-200  & 1.23 & 1.33 & 1.37 \\
AdapAlpaca-400  & 1.24 & 1.37 & 1.43 \\
AdapAlpaca-600  & 1.25 & 1.44 & 1.46 \\
AdapAlpaca-800  & 1.24 & 1.51 & 1.49 \\
AdapAlpaca-1000 & 1.29 & 1.57 & 1.53 \\
\bottomrule
\end{tabular}
}
\caption{Average artifact scores (1 = best, 5 = worst) across different response length intervals in \OURS{}. All values are averaged over 5 annotators and 50 samples per bucket.}
\label{tab:artifact-results}
\end{table}

\section{Motivation for Quality Decomposition}
\label{sec:design-Decomposition}

Our proposed decomposition of response quality into \emph{desirability} and \emph{information mass} is grounded in both practical considerations from LLM evaluation and theoretical intuitions derived from communication and information theory. While many prior works have treated response length as a confounding factor in evaluation~\cite{dubois2024length,koo2023benchmarking}, we posit that length itself is not inherently problematic. Rather, its influence emerges through two intertwined but distinguishable dimensions: how much \emph{content} is conveyed (i.e., information mass) and whether that content is \emph{desirable} in terms of trustworthiness and utility.
Desirability refers to qualities of a response that are invariant to length, such as factual correctness, logical coherence, non-toxicity, and consistency with the prompt. These properties reflect whether a response is acceptable or useful to a human annotator, regardless of verbosity. From a decision-theoretic perspective, desirability can be understood as a measure of alignment between the generated output and the implicit reward signal used in preference modeling.
Information mass, on the other hand, captures the amount of content present in a response. We draw on classical information theory, where entropy measures the uncertainty or information content of a message~\cite{von1932mathematische}. In conditional settings—i.e., given an instruction—responses with more unique, relevant, and elaborated content exhibit higher conditional entropy. We operationalize this using a language model to compute the average token-level conditional entropy, which serves as a tractable proxy for the richness of the response.
This decomposition is also motivated by cognitive theories of communication, which suggest that human preference judgments are influenced both by what is said (informativeness) and how well it is said (coherence, fluency, and alignment with norms). By modeling response quality as a function of these two components, we obtain a framework that explains why longer responses often receive higher win rates: they accumulate more information mass, which is rewarded by LLM-based annotators—even when desirability remains constant.
Moreover, this formulation enables targeted intervention. For example, when a model's win rate is low despite acceptable desirability, it may benefit from generating responses with higher information mass. Conversely, increasing length without regard to desirability may artificially inflate quality judgments without improving actual usefulness. Our experiments empirically validate that win rate is influenced by these two components in separable and predictable ways.

\begin{table*}[tbh!]
    \caption{
    Evaluation Prompts: Confusion, Redundancy, Dilution Rating Scales.
    }
    \centering
    \small
    \begin{tabular}{p{2.5cm}p{10.5cm}}
    \toprule
    \textbf{Prompt} & %
    Instruction Content \\
    \midrule
    \textbf{Confusion} & 

    \#\#\# [Question]
    
    *(the question.)*
    
    ---
    
    \#\#\# [Answer]
    
    *(the answer.)*
    
    ---
    
    \#\#\# Evaluation Guidelines:
    
    Confusion refers to unclear logic, ambiguous phrasing, inconsistent reasoning, or abrupt shifts in topic within the response.
    
    Please assign a rating from 1 to 5 according to the following scale:
    
    - 1 (No confusion): The answer is clear, coherent, and logically consistent throughout.  
    
    - 2 (Slight confusion): The answer is mostly clear, with minor confusing elements that do not significantly hinder understanding.  
    
    - 3 (Moderate confusion): The answer contains noticeable points of confusion (e.g., ambiguous wording or unclear logic), though the overall meaning can still be understood.  
    
    - 4 (High confusion): The answer frequently exhibits confusing elements that substantially impact its clarity and readability.  
    
    - 5 (Severe confusion): The answer is largely incoherent, confusing, or impossible to understand clearly. \\

    \midrule
    \textbf{Redundancy} & 

    \#\#\# [Question]
    
    *(the question.)*
    
    ---
    
    \#\#\# [Answer]
    
    *(the answer.)*
    
    ---
    
    \#\#\# Evaluation Guidelines:
    
    Redundancy refers to unnecessary repetition of information, overly verbose expressions, or prolonged restatements that do not contribute additional meaningful content.
    
    Please assign a rating from 1 to 5 according to the following scale:
    
    - 1 (No redundancy): The answer is concise and does not contain unnecessary repetition.  
    
    - 2 (Slight redundancy): The answer occasionally repeats information, but overall readability and clarity remain strong.  
    
    - 3 (Moderate redundancy): The answer has noticeable repetitive phrasing or ideas that slightly reduce clarity or conciseness.  
    
    - 4 (High redundancy): The answer frequently repeats information, negatively impacting readability and efficiency of communication.  
    
    - 5 (Severe redundancy): The answer is dominated by repetitive or verbose content, significantly reducing clarity or informational value. \\

    \midrule
    
    \textbf{Dilution} & 

    \#\#\# [Question]
    
    *(the question.)*
    
    ---
    
    \#\#\# [Answer]
    
    *(the answer.)*
    
    ---
    
    \#\#\# Evaluation Guidelines:
    
    Dilution of meaning refers to the presence of vague, generic, or filler content in the answer, resulting in reduced informativeness and specificity.
    
    Please assign a rating from 1 to 5 according to the following scale:
    
    - 1 (No dilution): The answer consistently provides specific, informative, and meaningful content.  
    
    - 2 (Slight dilution): The answer is generally specific but contains minor instances of vague or filler content.  
    
    - 3 (Moderate dilution): The answer includes several sections with general or nonspecific information, noticeably reducing its informativeness.  
    
    - 4 (High dilution): The answer mostly consists of generic statements or filler material, offering limited specific insights.  
    
    - 5 (Severe dilution): The answer is overwhelmingly vague, generic, or lacking specific information, providing minimal to no meaningful content. \\

    \bottomrule
    \end{tabular}
\label{tab:other_detail_prompt}
\end{table*}

\section{Prompt Content}
\label{sec:prompt-content}
Here, we show the 6 prompts in Table~\ref{table:otherprompt} we used to generate the AlpacaEval answers.

\section{Quality Decomposition Across Diverse Test Model}
\label{sec:QEwithMoreTestModel}

\begin{figure*}[h]
{
    \hfill
    \subfigure[\small AlpacaEval]{
        \includegraphics[width=0.315\textwidth]{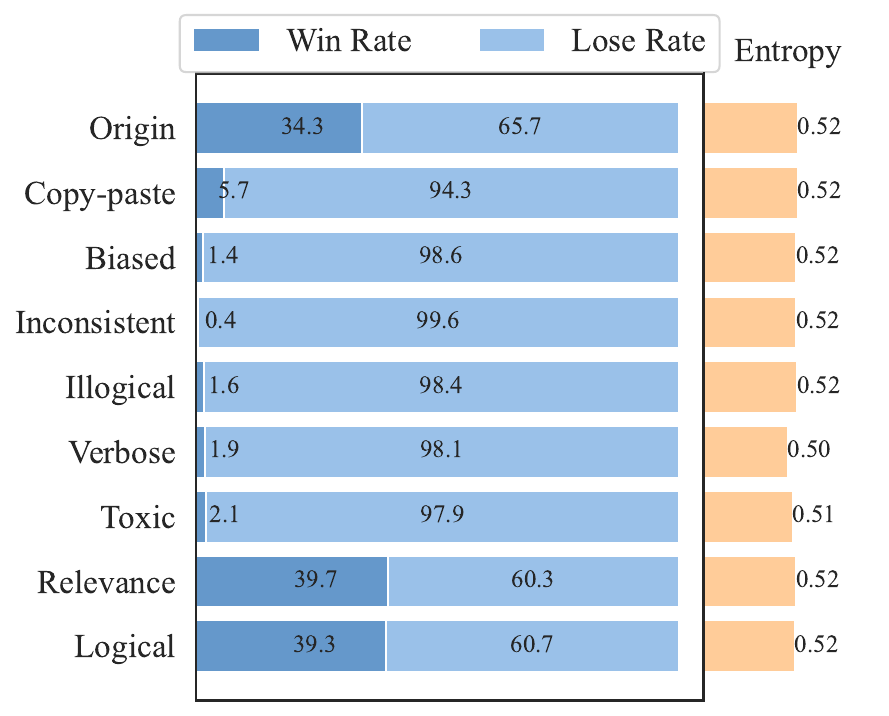}
    }
    \hfill
    \subfigure[\small LIMA]{
        \includegraphics[width=0.315\textwidth]{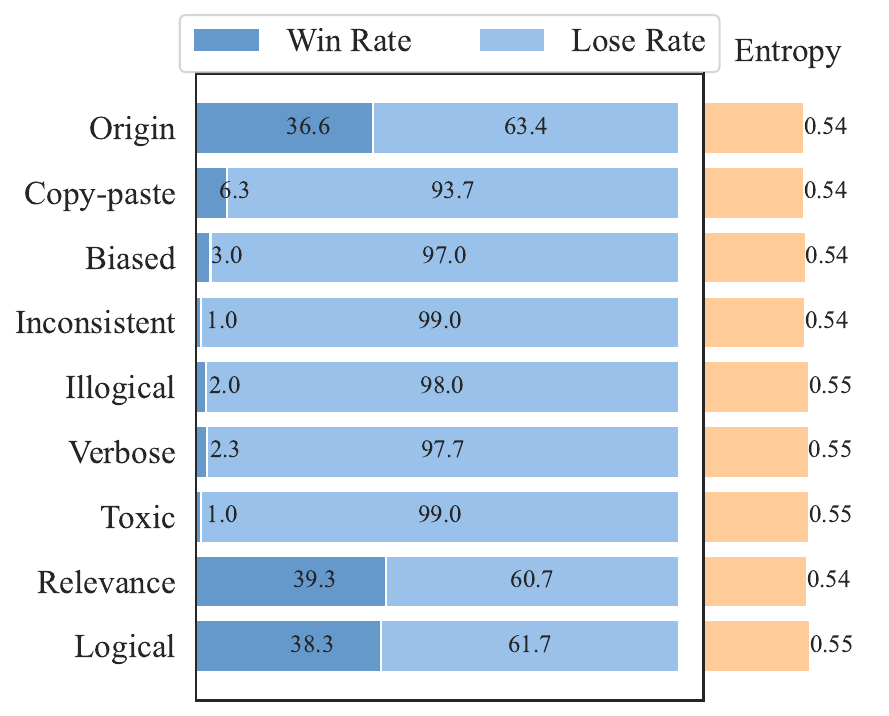}
    }
    \hfill
    \subfigure[\small Koala]{
        \includegraphics[width=0.315\textwidth]{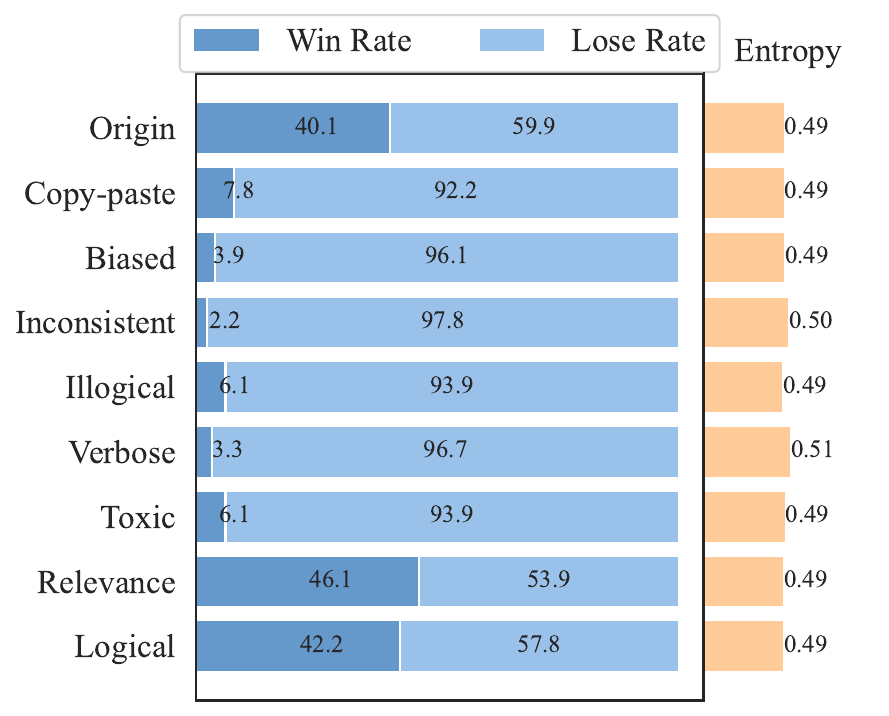}
    }
    \hfill
    \subfigure[\small Self-Instruct]{
        \includegraphics[width=0.315\textwidth]{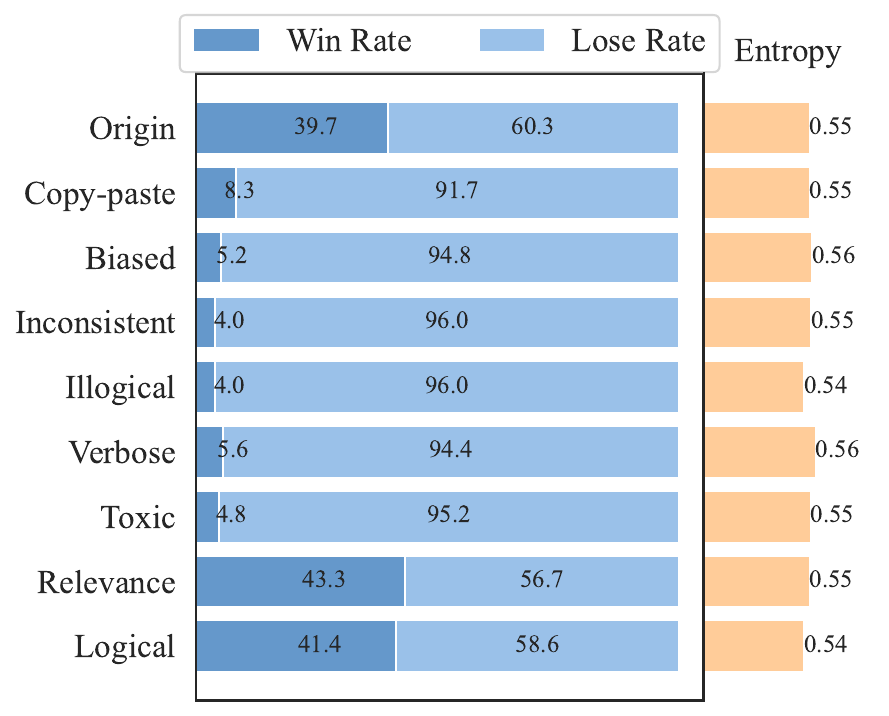}
    }
    \hfill
    \subfigure[\small Vicuna]{
        \includegraphics[width=0.315\textwidth]{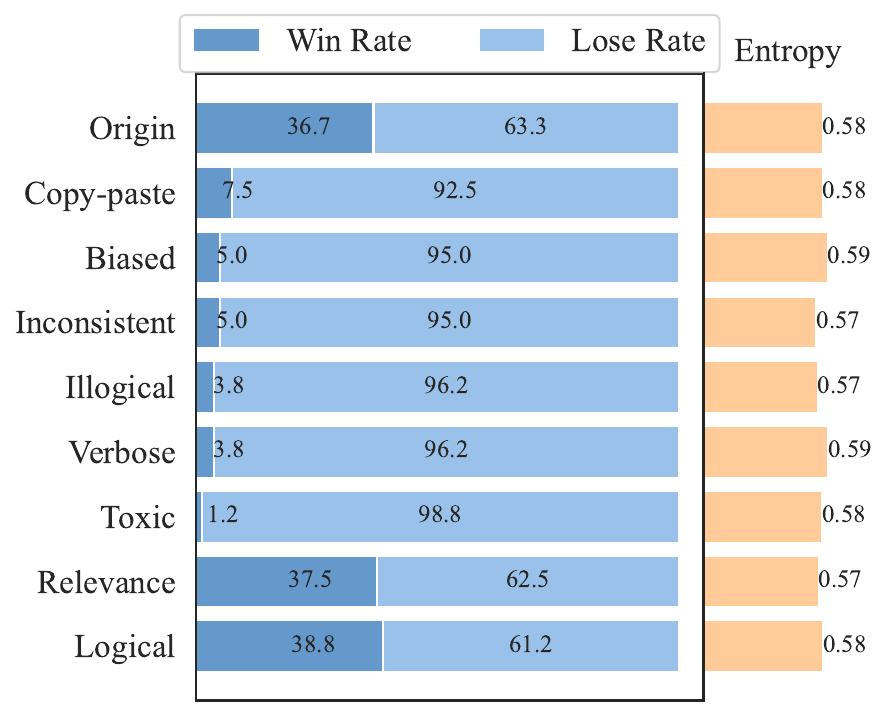}
    }
    \hfill
    \subfigure[\small Wizardlm]{
        \includegraphics[width=0.315\textwidth]{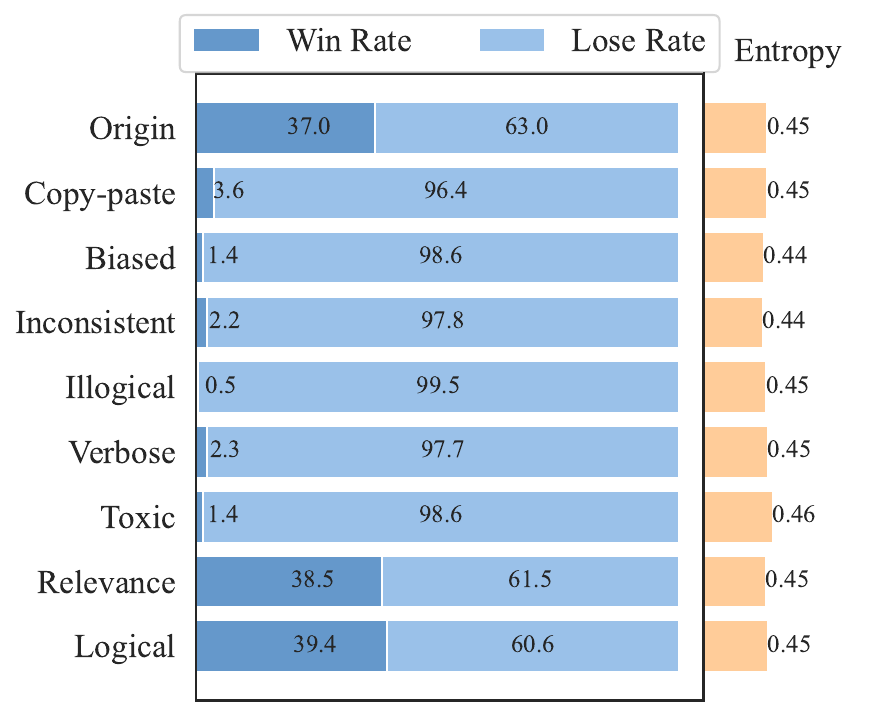}
    }
}
\caption{Validation of desirability’s impact on quality for  LLAMA3-70b. } 
\label{fig:Llama370b-desirability}
\end{figure*}

\begin{figure*}[h]
{
    \hfill
    \subfigure[\small AlpacaEval]{
        \includegraphics[width=0.315\textwidth]{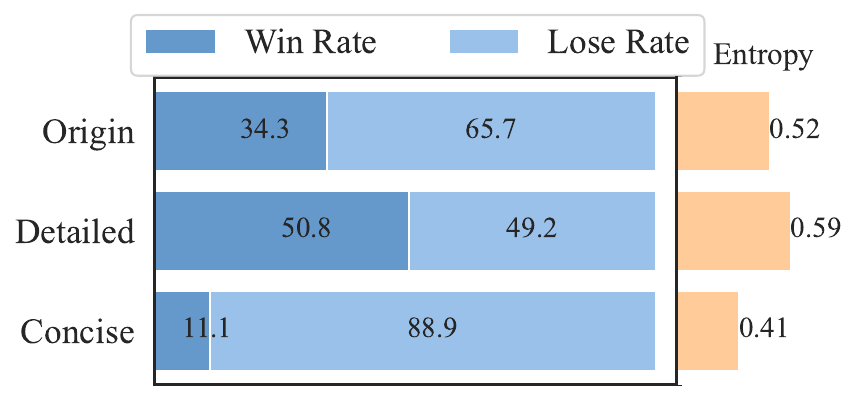}
    }
    \hfill
    \subfigure[\small LIMA]{
        \includegraphics[width=0.315\textwidth]{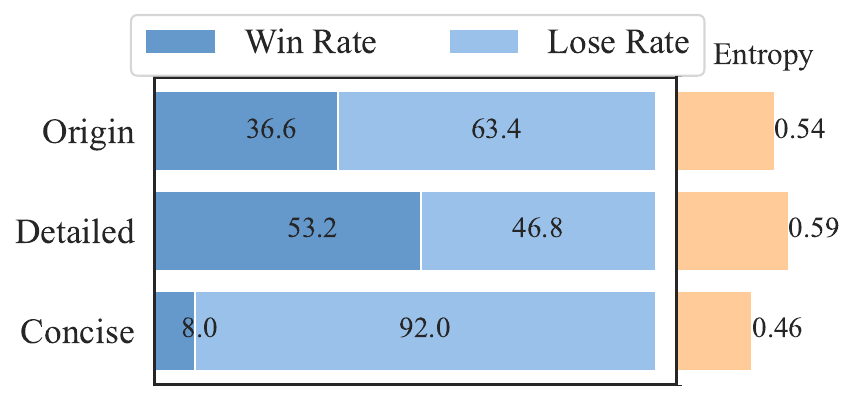}
    }
    \hfill
    \subfigure[\small Koala]{
        \includegraphics[width=0.315\textwidth]{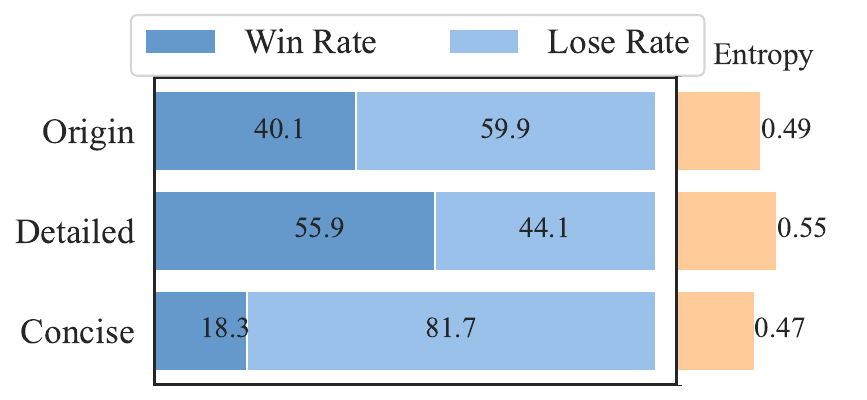}
    }
    \hfill
    \subfigure[\small Self-Instruct]{
        \includegraphics[width=0.315\textwidth]{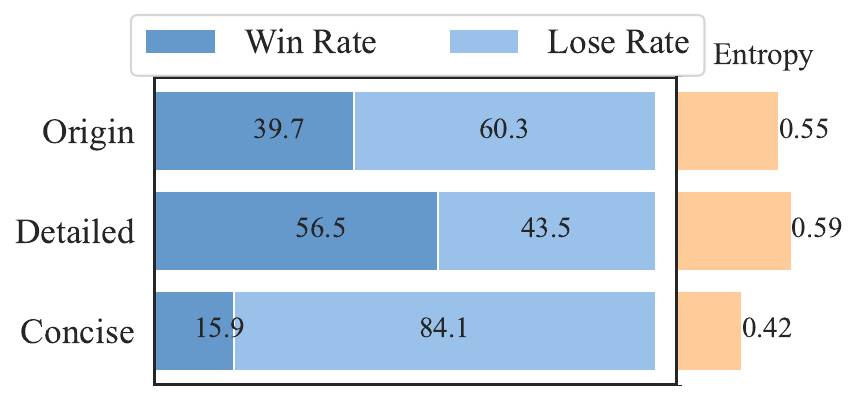}
    }
    \hfill
    \subfigure[\small Vicuna]{
        \includegraphics[width=0.315\textwidth]{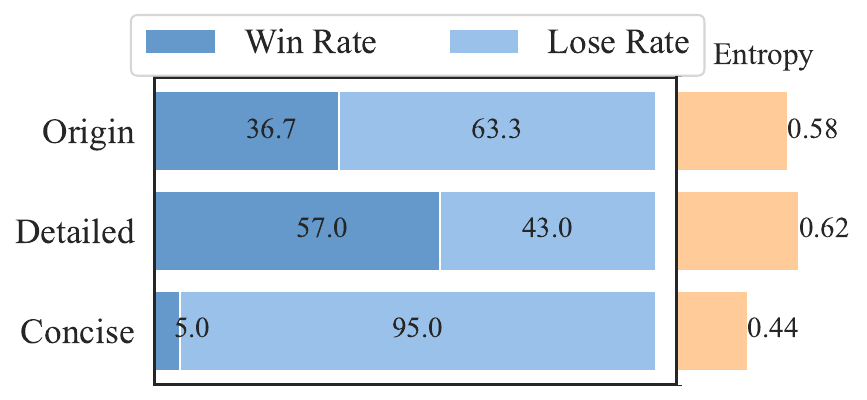}
    }
    \hfill
    \subfigure[\small Wizardlm]{
        \includegraphics[width=0.315\textwidth]{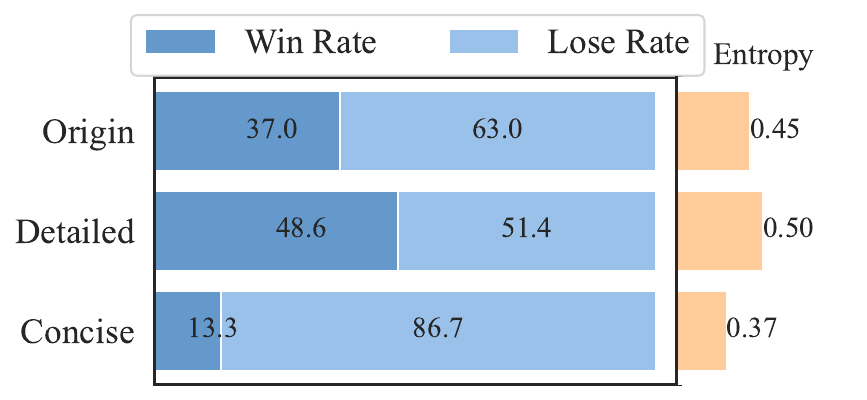}
    }
}
\caption{Validation of information mass’s impact on quality for LLAMA3-70b.} 
\label{fig:Llama370b-infomass}
\end{figure*}

\begin{figure*}[h]
{
    \hfill
    \subfigure[\small AlpacaEval]{
        \includegraphics[width=0.315\textwidth]{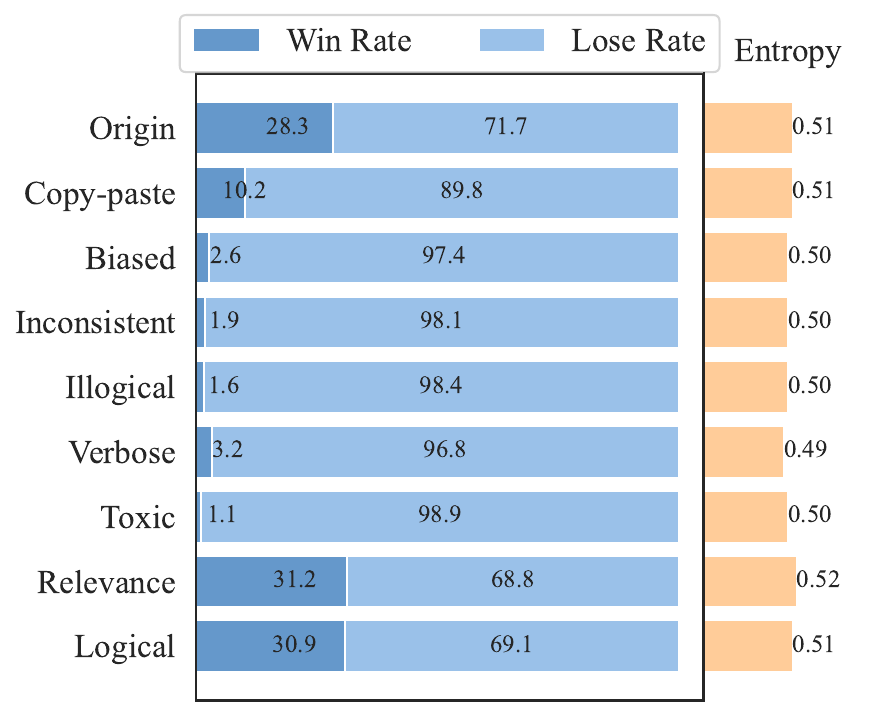}
    }
    \hfill
    \subfigure[\small LIMA]{
        \includegraphics[width=0.315\textwidth]{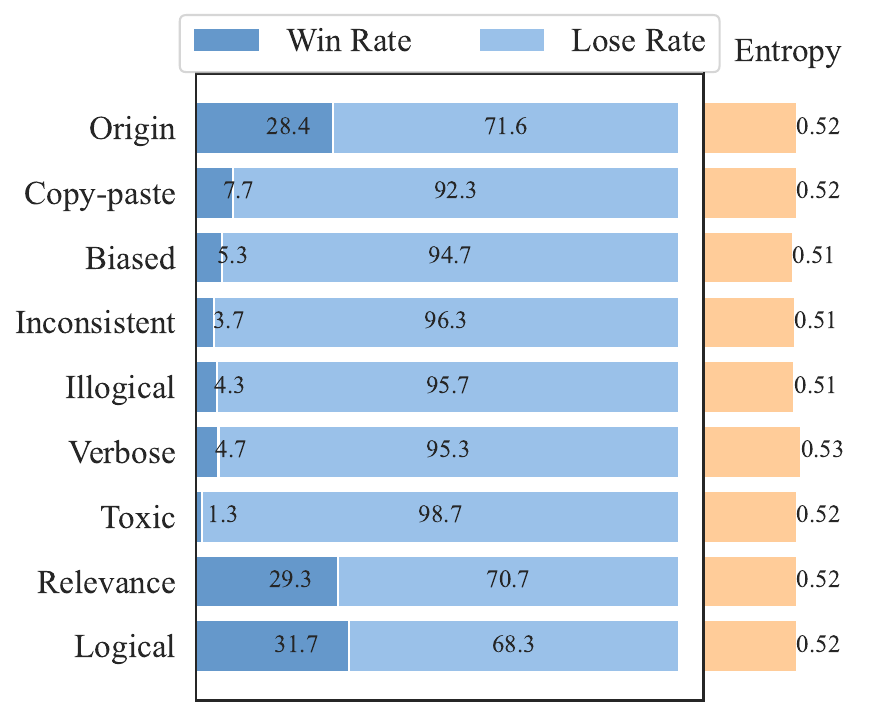}
    }
    \hfill
    \subfigure[\small Koala]{
        \includegraphics[width=0.315\textwidth]{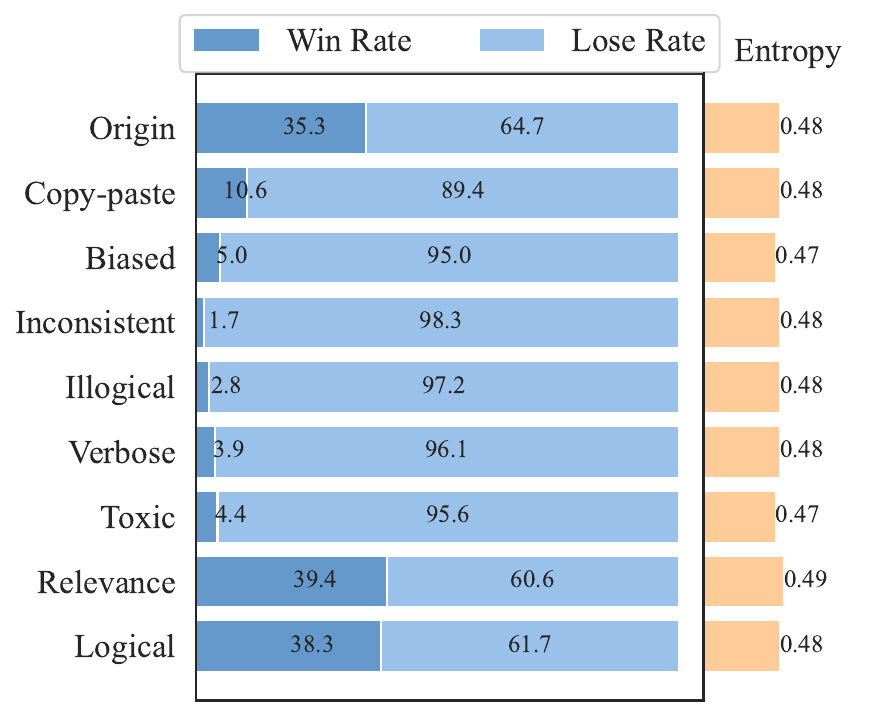}
    }
    \hfill
    \subfigure[\small Self-Instruct]{
        \includegraphics[width=0.315\textwidth]{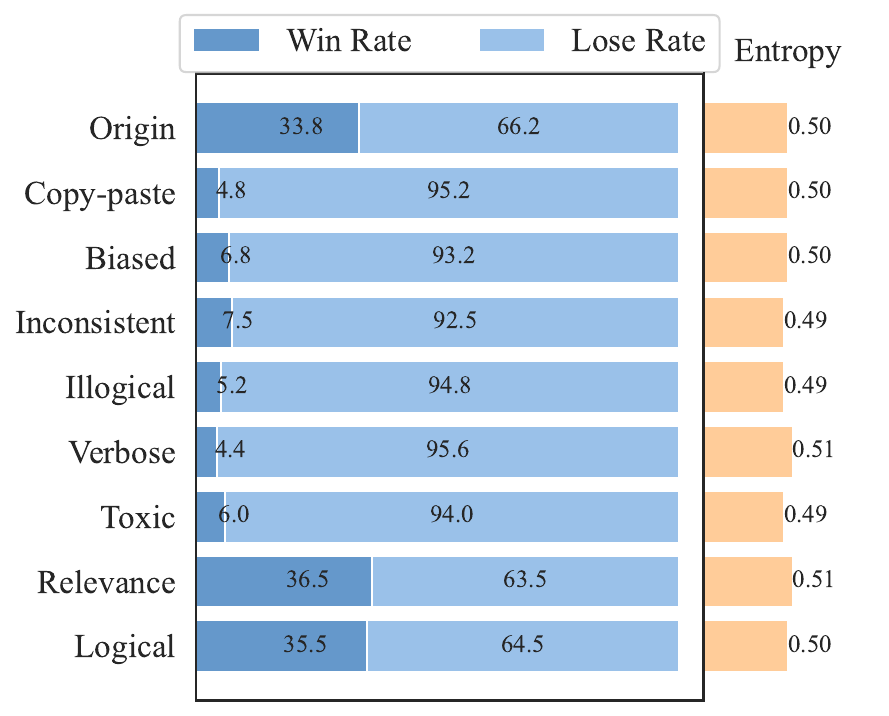}
    }
    \hfill
    \subfigure[\small Vicuna]{
        \includegraphics[width=0.315\textwidth]{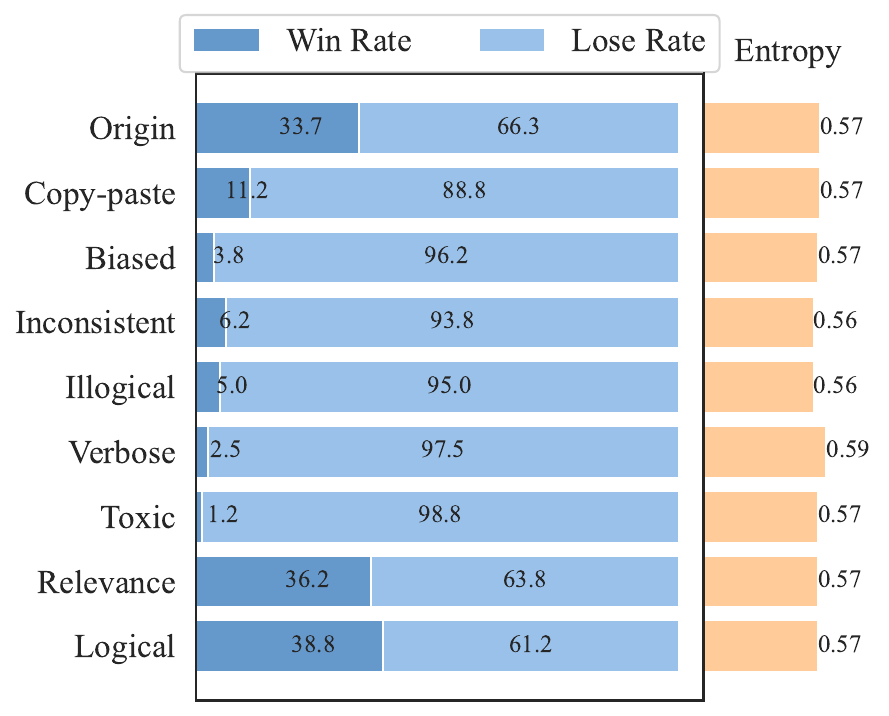}
    }
    \hfill
    \subfigure[\small Wizardlm]{
        \includegraphics[width=0.315\textwidth]{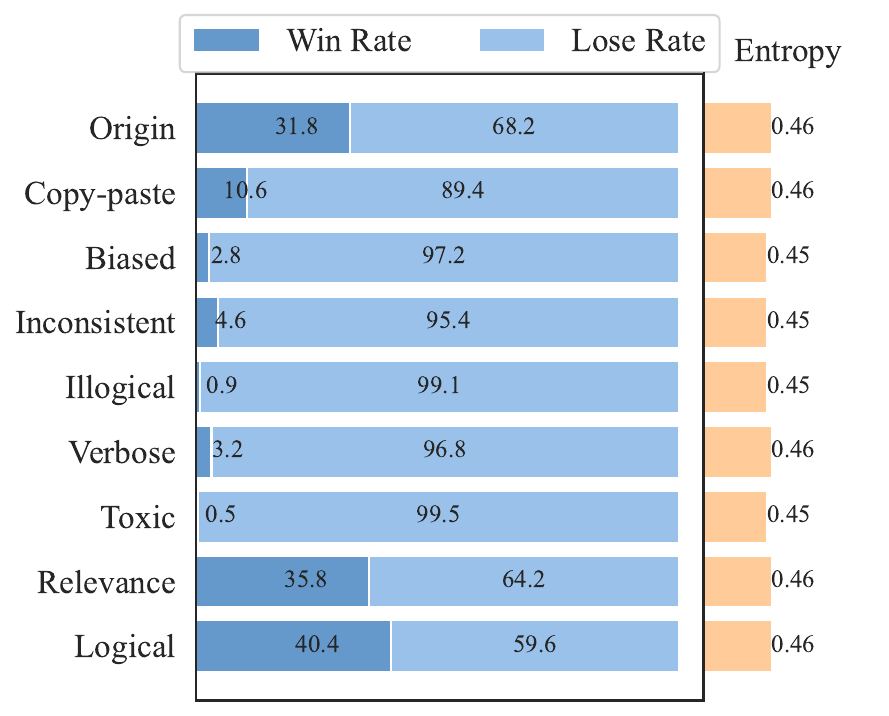}
    }
}
\caption{Validation of desirability’s impact on quality for Qwen1.5-72b.} 
\label{fig:Qwen1.5-72b-desirability}
\end{figure*}

\begin{figure*}[h]
{
    \hfill
    \subfigure[\small AlpacaEval]{
        \includegraphics[width=0.315\textwidth]{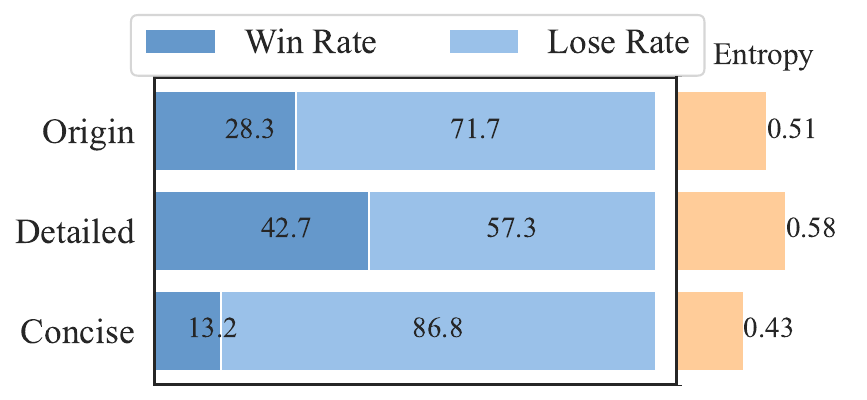}
    }
    \hfill
    \subfigure[\small LIMA]{
        \includegraphics[width=0.315\textwidth]{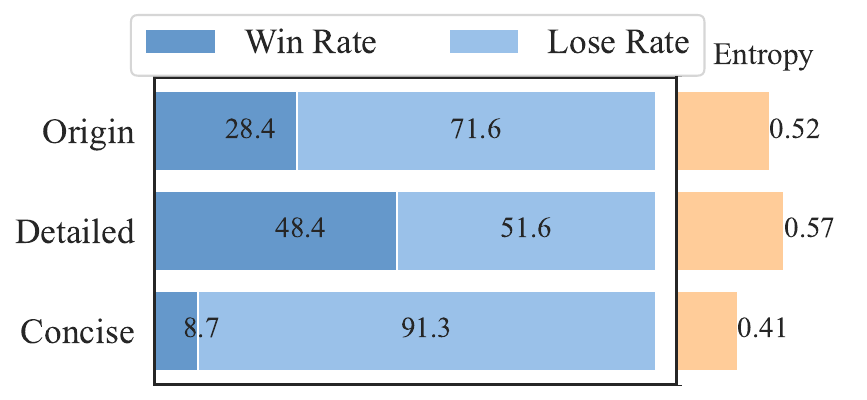}
    }
    \hfill
    \subfigure[\small Koala]{
        \includegraphics[width=0.315\textwidth]{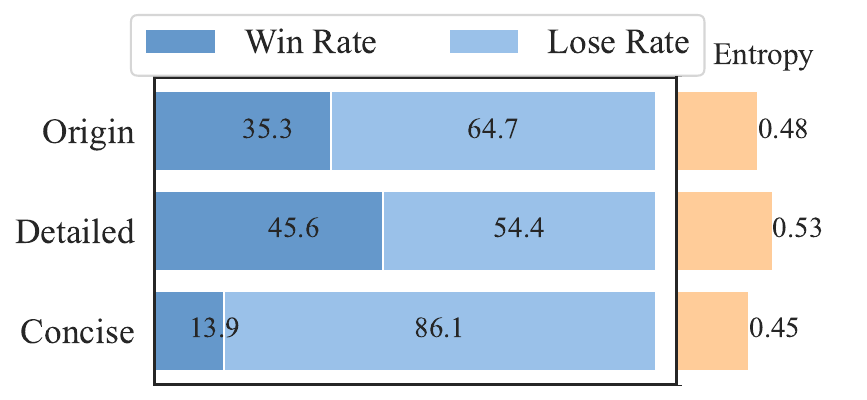}
    }
    \hfill
    \subfigure[\small Self-Instruct]{
        \includegraphics[width=0.315\textwidth]{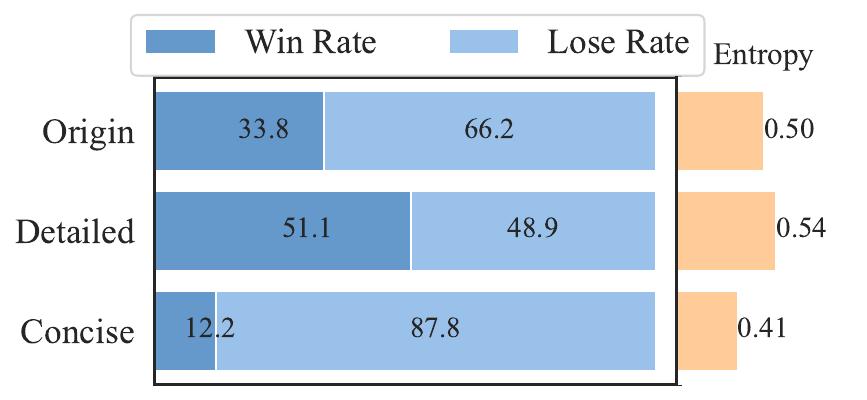}
    }
    \hfill
    \subfigure[\small Vicuna]{
        \includegraphics[width=0.315\textwidth]{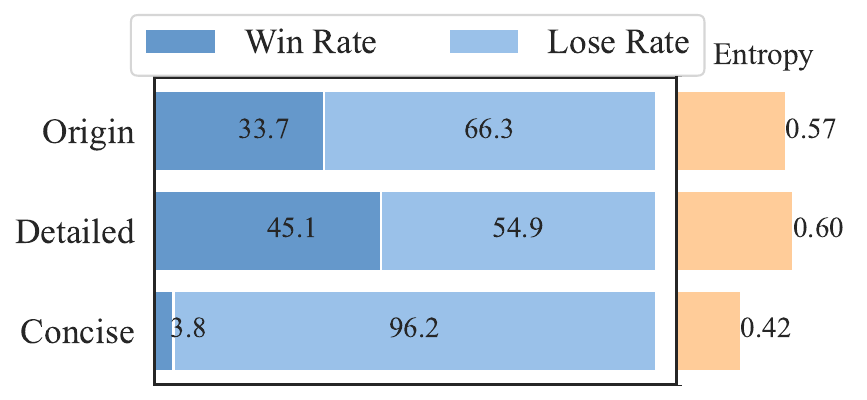}
    }
    \hfill
    \subfigure[\small Wizardlm]{
        \includegraphics[width=0.315\textwidth]{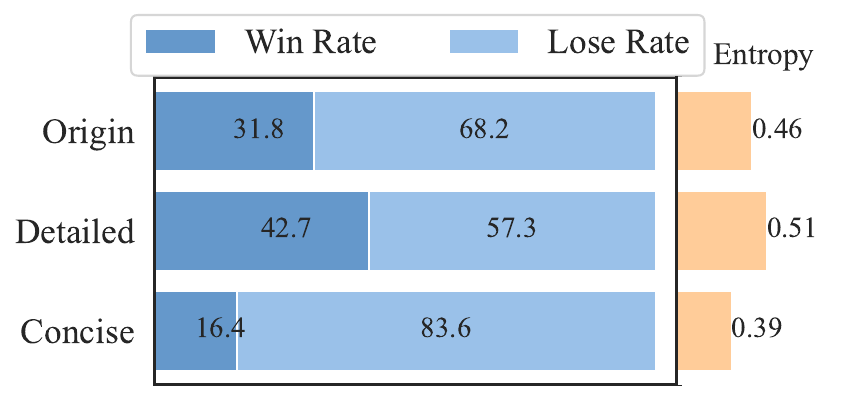}
    }
}
\caption{Validation of information mass’s impact on quality for Qwen1.5-72b.} 
\label{fig:Qwen1.5-72b-infomass}
\end{figure*}

\begin{figure*}[h]
{
    \hfill
    \subfigure[\small AlpacaEval]{
        \includegraphics[width=0.315\textwidth]{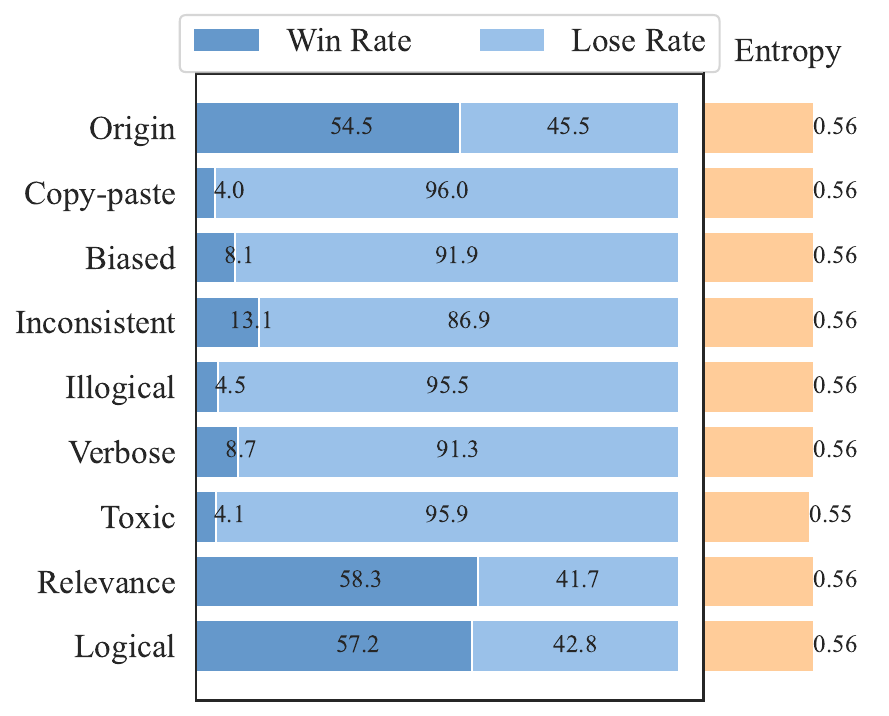}
    }
    \hfill
    \subfigure[\small LIMA]{
        \includegraphics[width=0.315\textwidth]{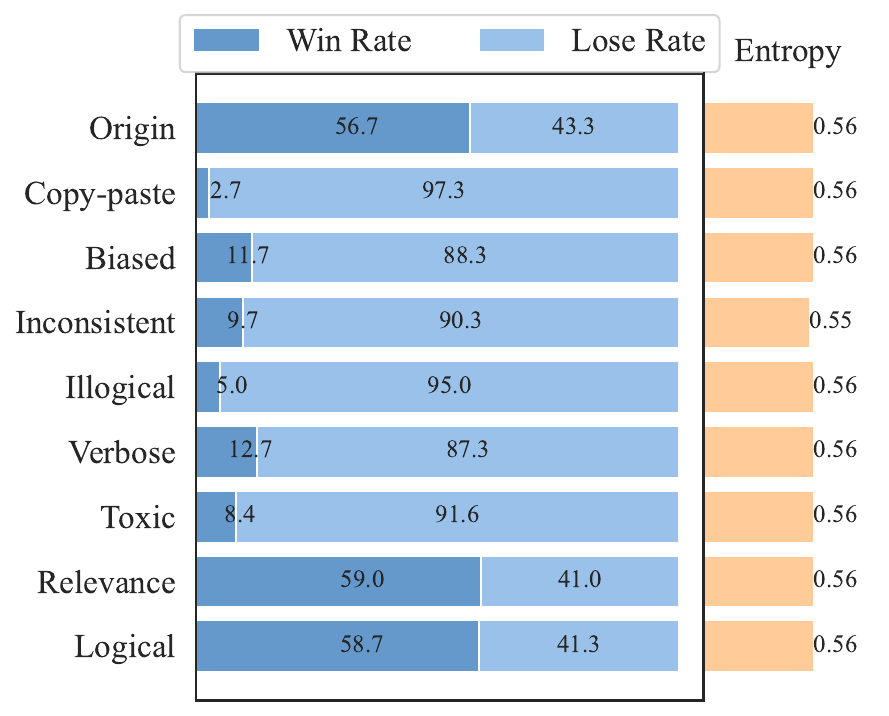}
    }
    \hfill
    \subfigure[\small Koala]{
        \includegraphics[width=0.315\textwidth]{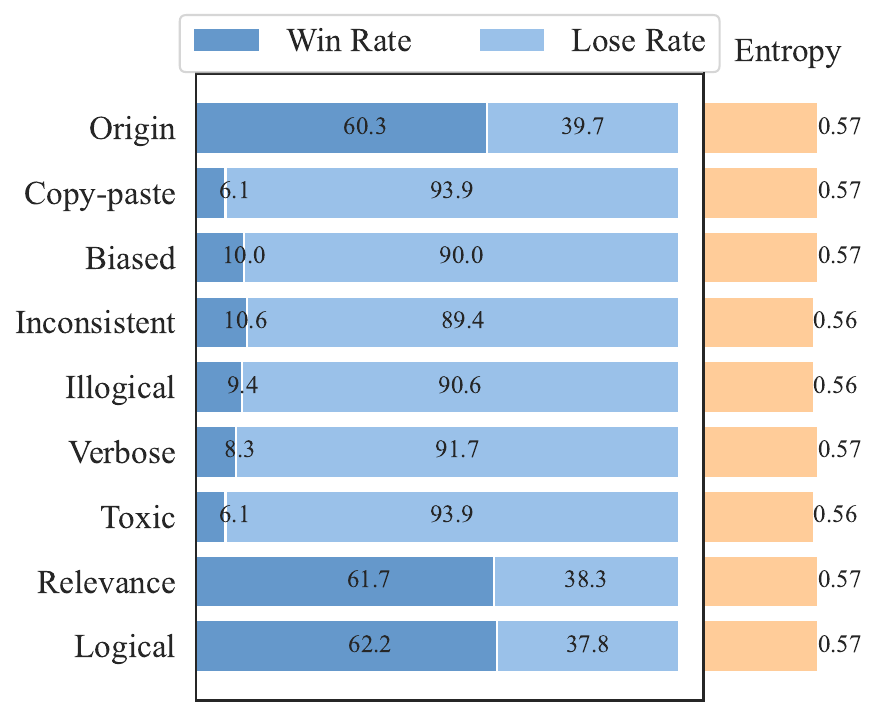}
    }
    \hfill
    \subfigure[\small Self-Instruct]{
        \includegraphics[width=0.315\textwidth]{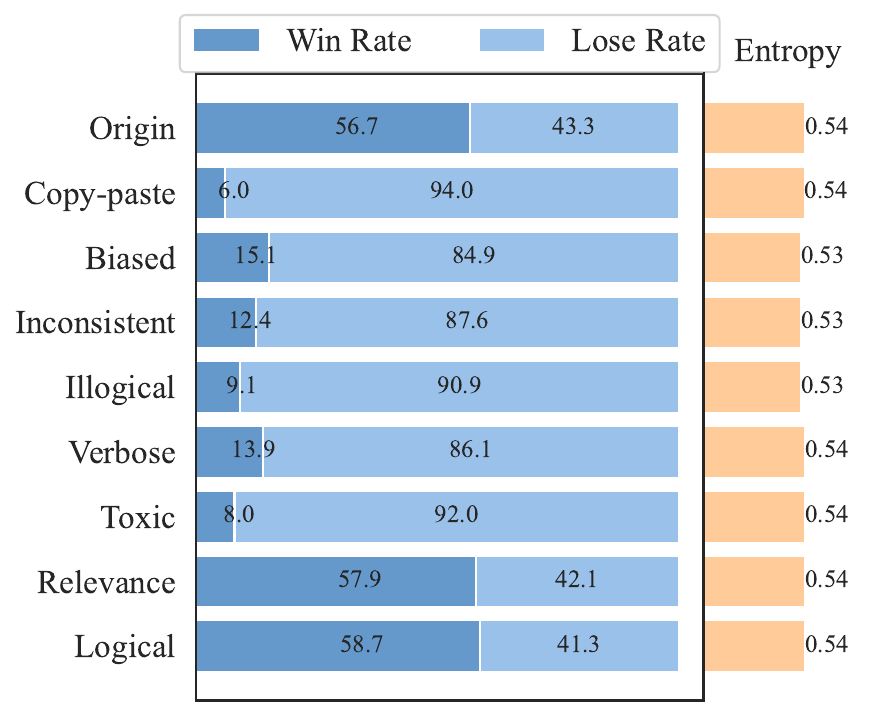}
    }
    \hfill
    \subfigure[\small Vicuna]{
        \includegraphics[width=0.315\textwidth]{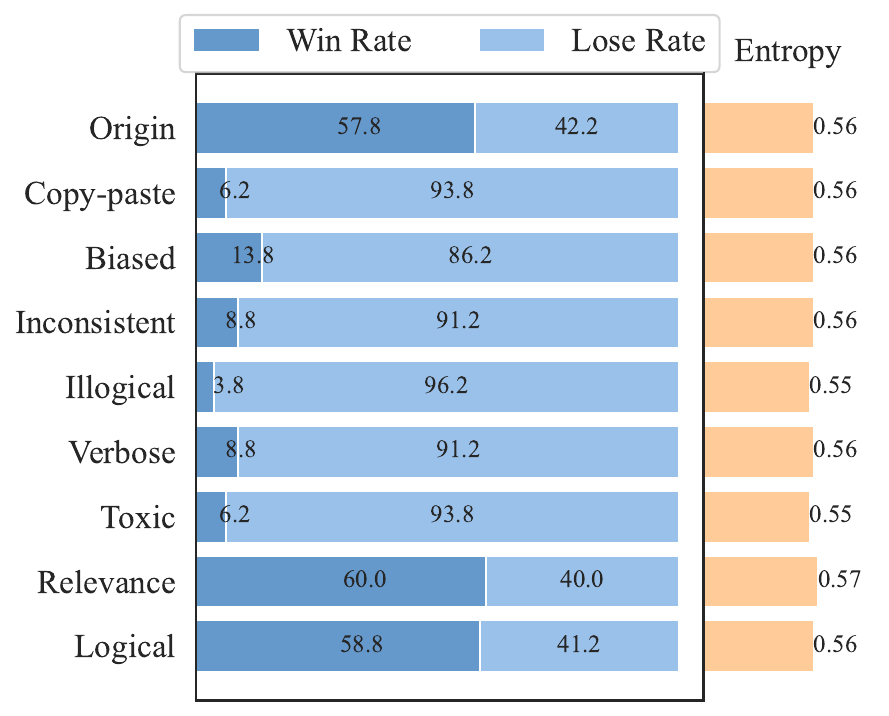}
    }
    \hfill
    \subfigure[\small Wizardlm]{
        \includegraphics[width=0.315\textwidth]{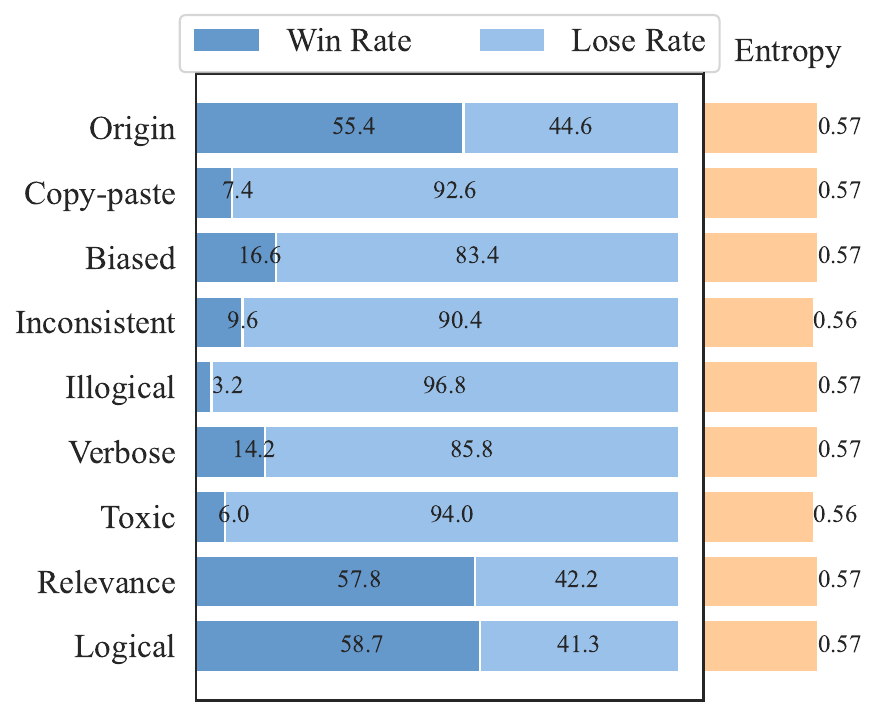}
    }
}
\caption{Validation of desirability’s impact on quality for GPT4-o.} 
\label{fig:GPT4o-desirability}
\end{figure*}

\begin{figure*}[h]
{
    \hfill
    \subfigure[\small AlpacaEval]{
        \includegraphics[width=0.315\textwidth]{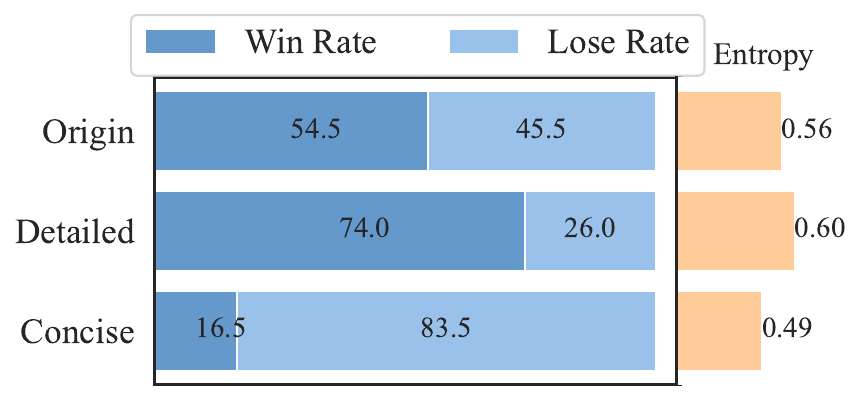}
    }
    \hfill
    \subfigure[\small LIMA]{
        \includegraphics[width=0.315\textwidth]{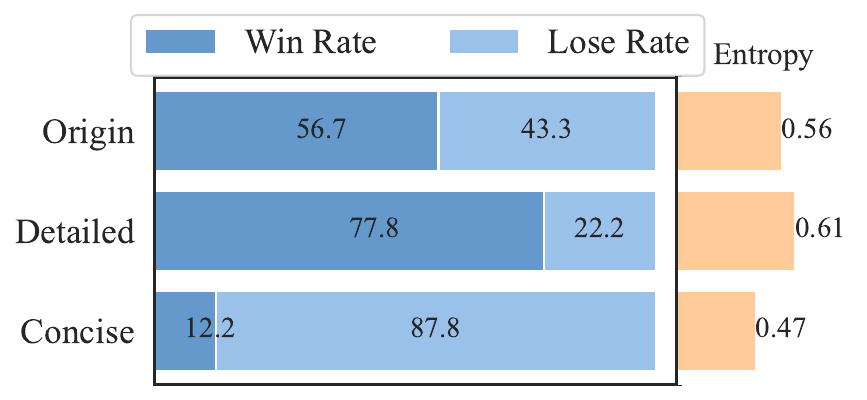}
    }
    \hfill
    \subfigure[\small Koala]{
        \includegraphics[width=0.315\textwidth]{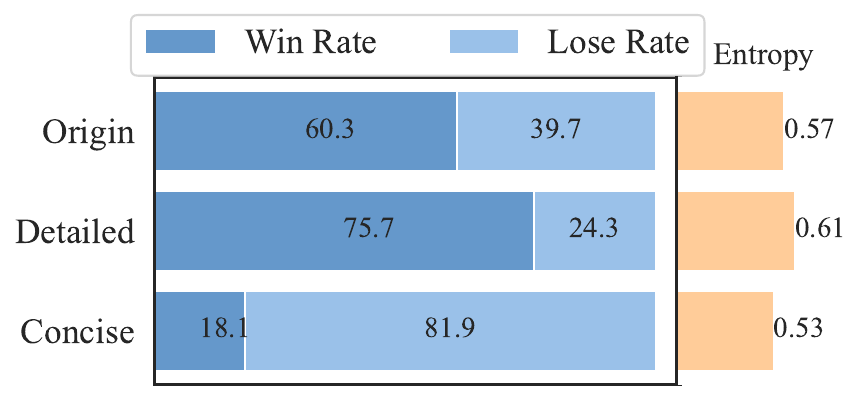}
    }
    \hfill
    \subfigure[\small Self-Instruct]{
        \includegraphics[width=0.315\textwidth]{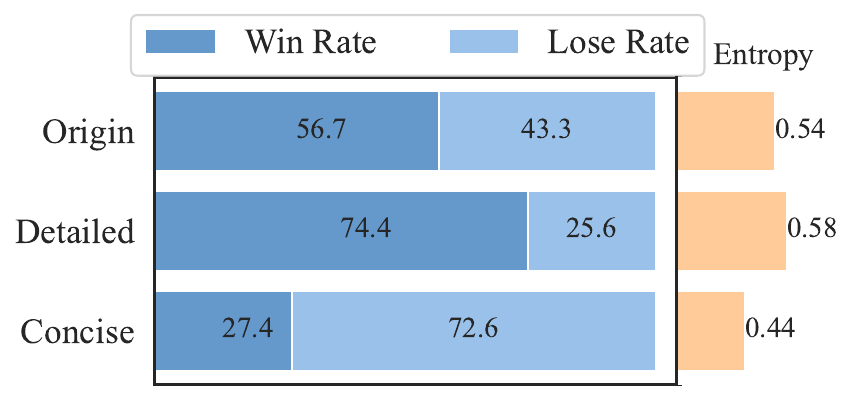}
    }
    \hfill
    \subfigure[\small Vicuna]{
        \includegraphics[width=0.315\textwidth]{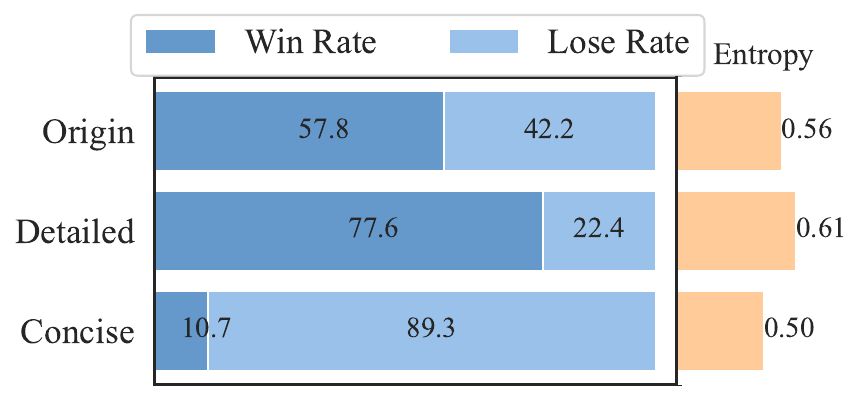}
    }
    \hfill
    \subfigure[\small Wizardlm]{
        \includegraphics[width=0.315\textwidth]{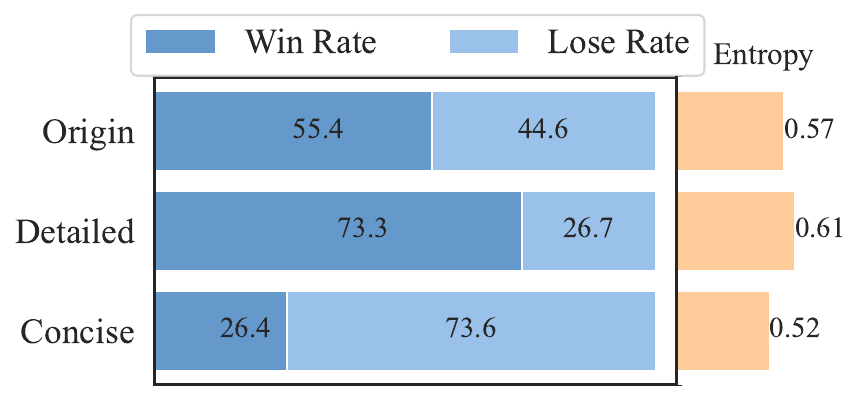}
    }
}
\caption{Validation of information mass’s impact on quality for GPT4-o.} 
\label{fig:GPT4o-infomass}
\end{figure*}

\begin{figure*}[h]
{
    \hfill
    \subfigure[\small AlpacaEval]{
        \includegraphics[width=0.315\textwidth]{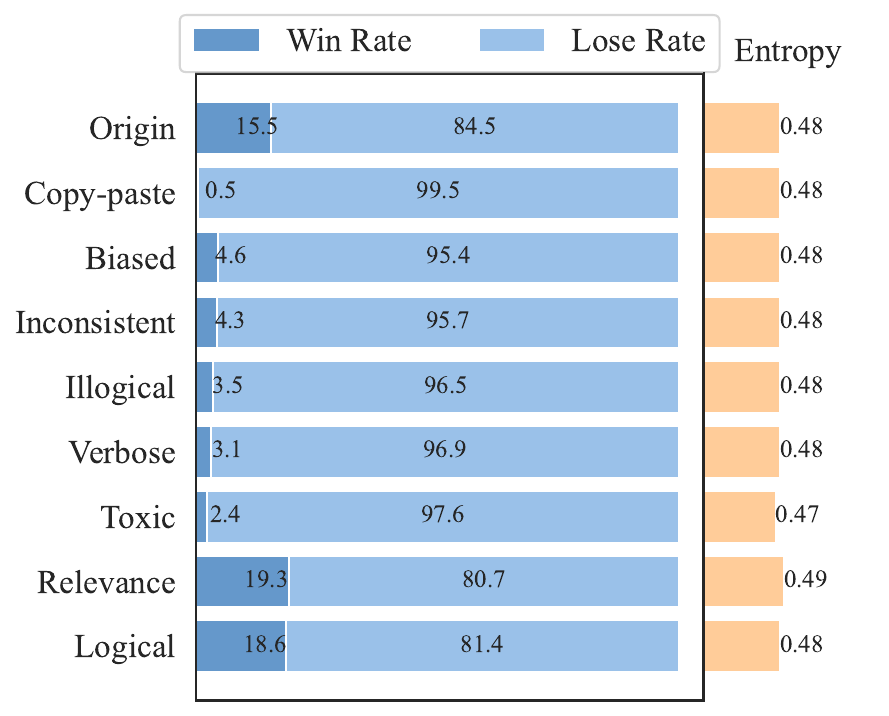}
    }
    \hfill
    \subfigure[\small LIMA]{
        \includegraphics[width=0.315\textwidth]{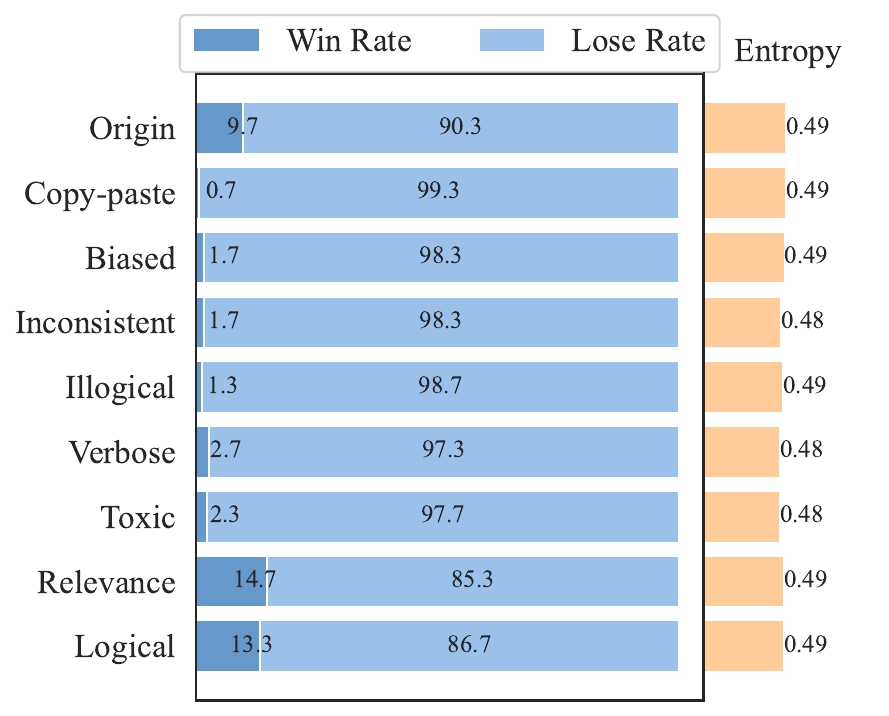}
    }
    \hfill
    \subfigure[\small Koala]{
        \includegraphics[width=0.315\textwidth]{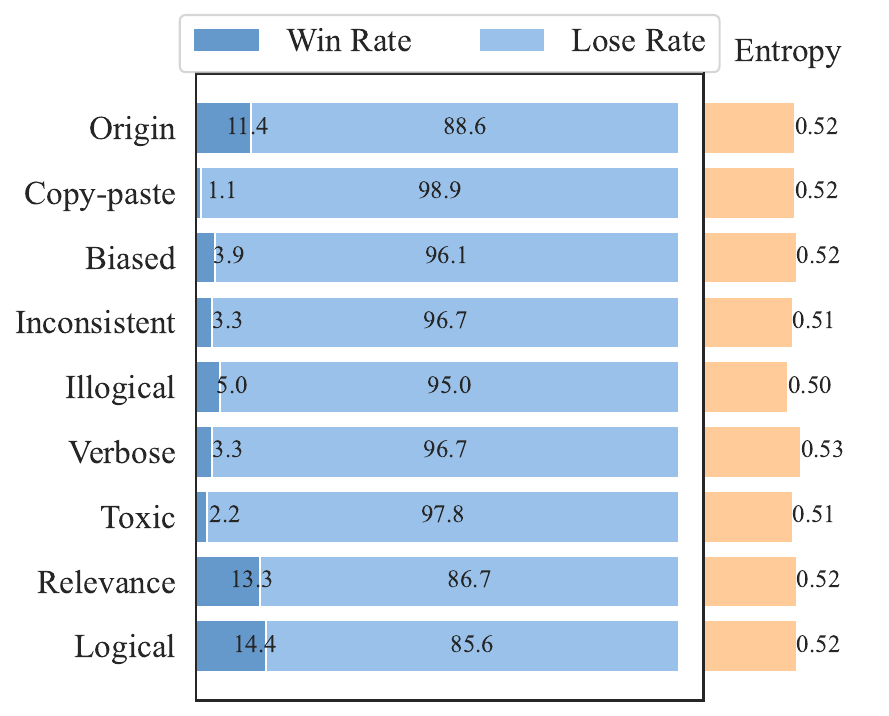}
    }
    \hfill
    \subfigure[\small Self-Instruct]{
        \includegraphics[width=0.315\textwidth]{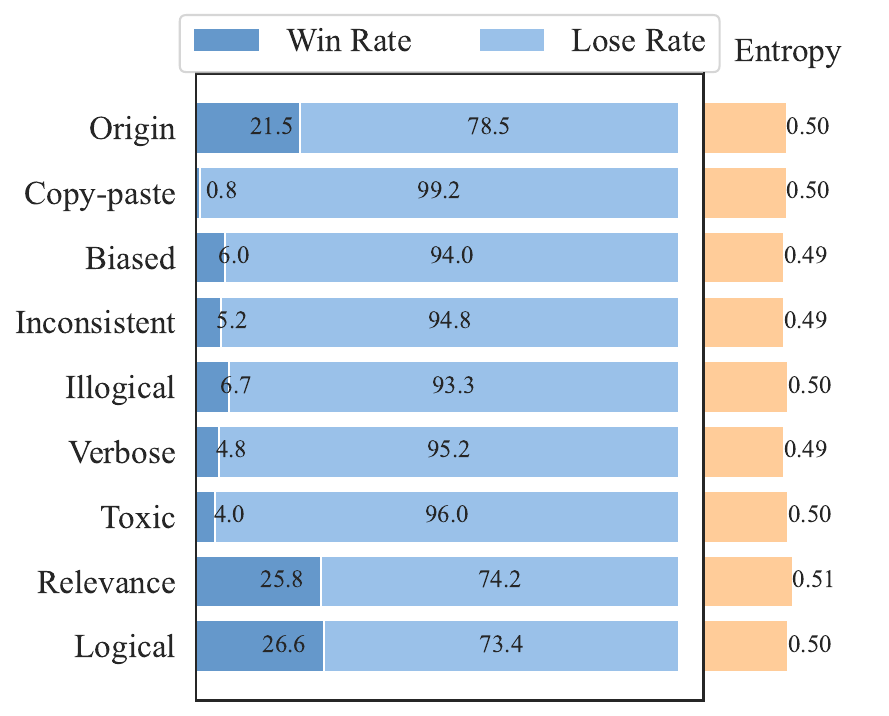}
    }
    \hfill
    \subfigure[\small Vicuna]{
        \includegraphics[width=0.315\textwidth]{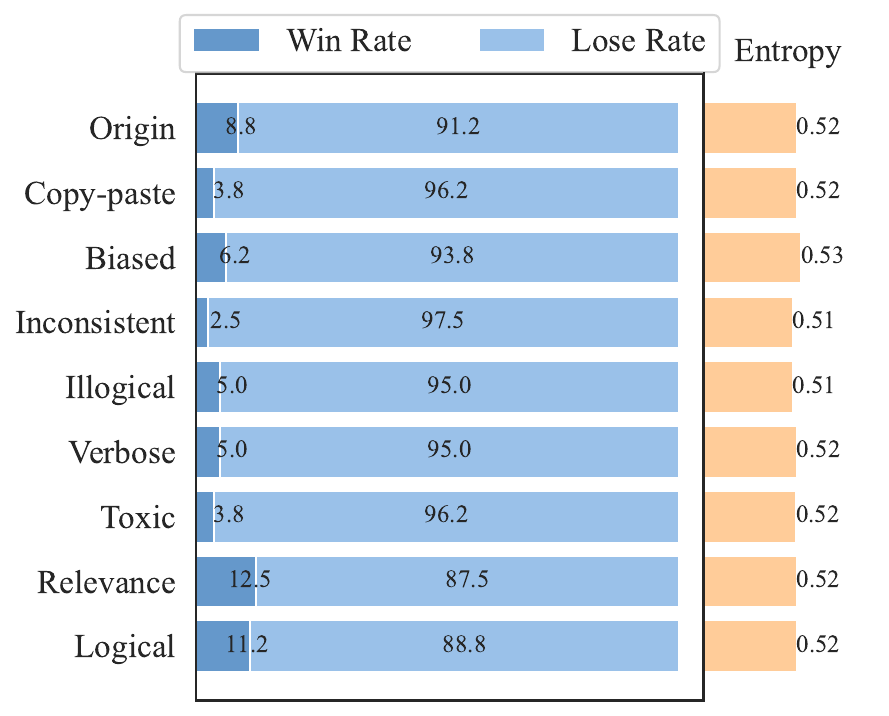}
    }
    \hfill
    \subfigure[\small Wizardlm]{
        \includegraphics[width=0.315\textwidth]{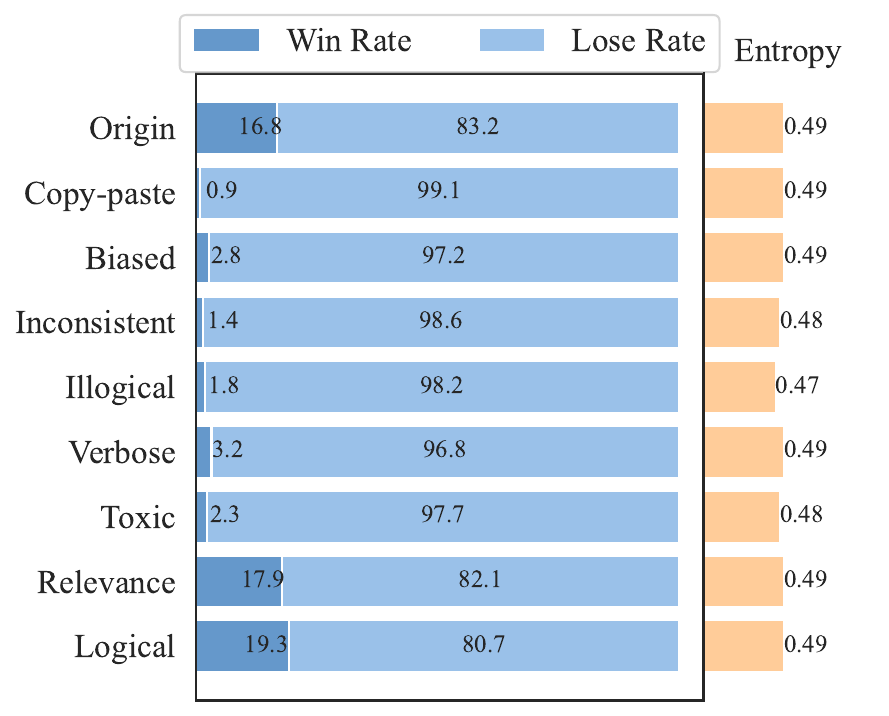}
    }
}
\caption{Validation of desirability’s impact on quality for GPT-3.5.} 
\label{fig:GPT-3.5-desirability}
\end{figure*}

\begin{figure*}[h]
{
    \hfill
    \subfigure[\small AlpacaEval]{
        \includegraphics[width=0.315\textwidth]{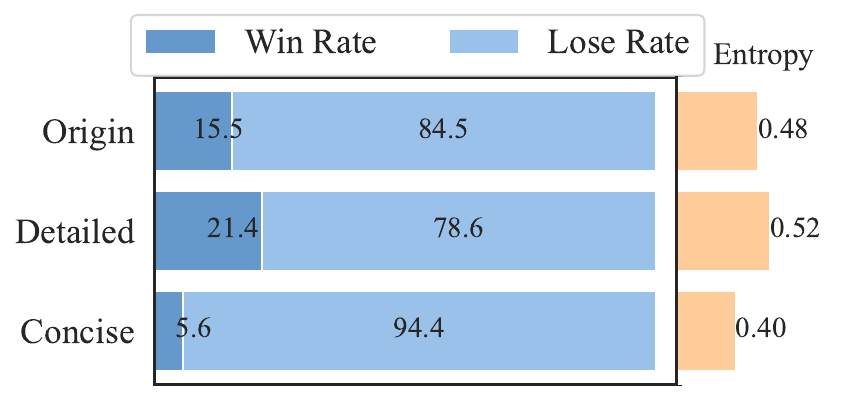}
    }
    \hfill
    \subfigure[\small LIMA]{
        \includegraphics[width=0.315\textwidth]{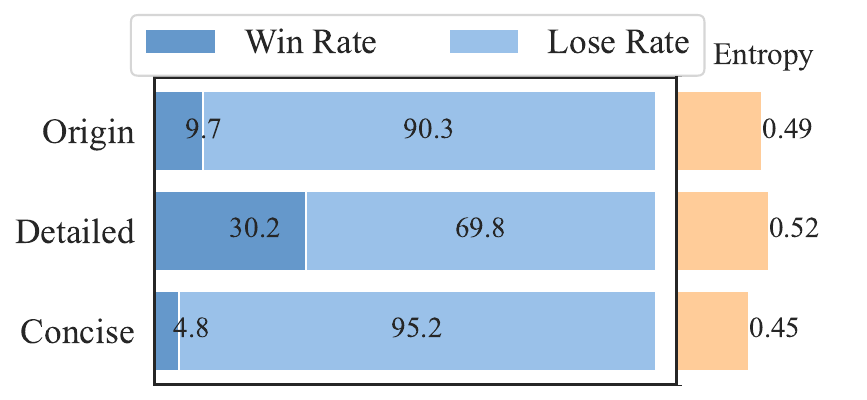}
    }
    \hfill
    \subfigure[\small Koala]{
        \includegraphics[width=0.315\textwidth]{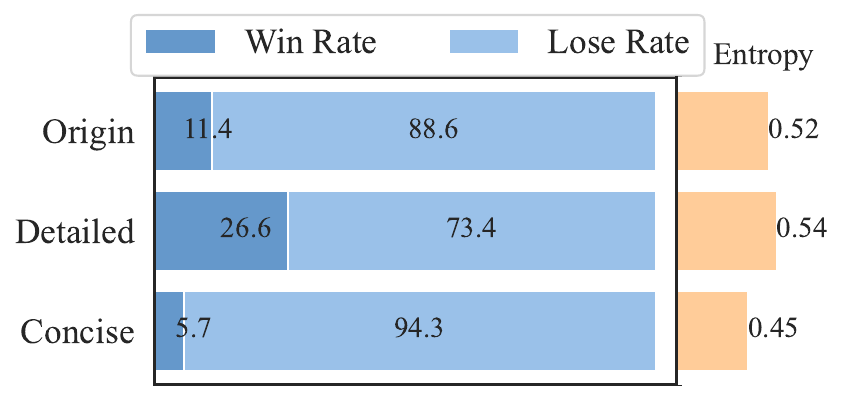}
    }
    \hfill
    \subfigure[\small Self-Instruct]{
        \includegraphics[width=0.315\textwidth]{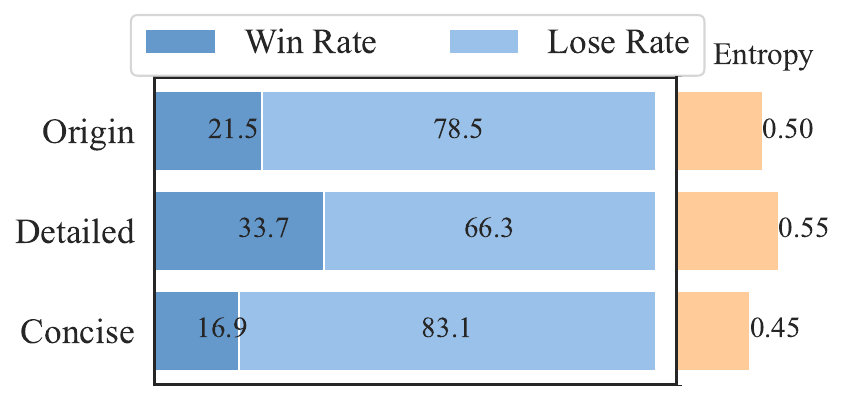}
    }
    \hfill
    \subfigure[\small Vicuna]{
        \includegraphics[width=0.315\textwidth]{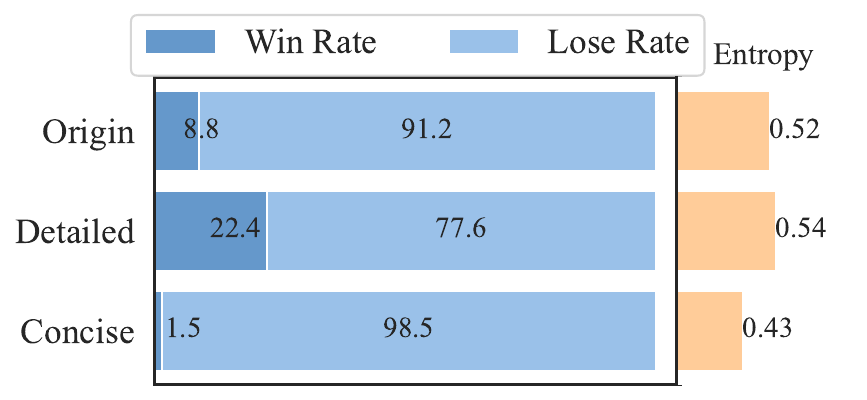}
    }
    \hfill
    \subfigure[\small Wizardlm]{
        \includegraphics[width=0.315\textwidth]{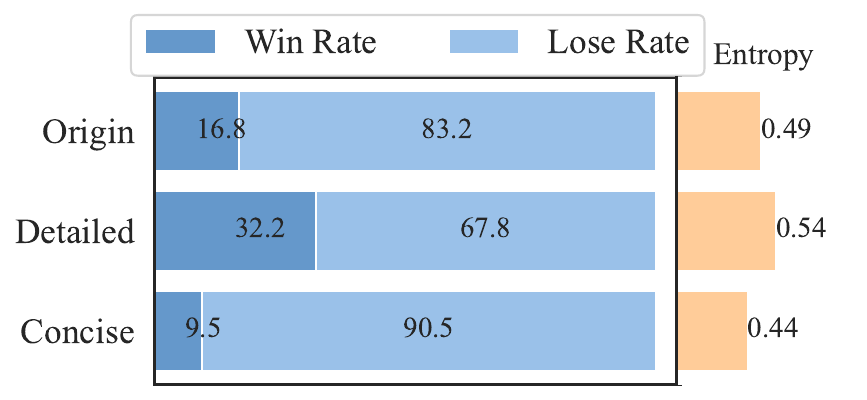}
    }
}
\caption{Validation of information mass’s impact on quality for GPT-3.5.} 
\label{fig:GPT-3.5-infomass}
\end{figure*}

To ensure our conclusions are not restricted to specific model architecture, we use LLAMA3-70b~\citep{llama3modelcard}, Qwen1.5-72b~\citep{qwen},  GPT4-o~\citep{achiam2023gpt} and GPT-3.5~\citep{achiam2023gpt}  as the backbone model. The results in Figure~\ref{fig:Llama370b-desirability}, Figure~\ref{fig:Llama370b-infomass}, Figure~\ref{fig:Qwen1.5-72b-desirability}, Figure~\ref{fig:Qwen1.5-72b-infomass}, Figure~\ref{fig:GPT4o-desirability}, Figure~\ref{fig:GPT4o-infomass}, Figure~\ref{fig:GPT-3.5-desirability} and Figure~\ref{fig:GPT-3.5-infomass}  show that different model backbone does not change the conclusions we derived.

\section{Quality Decomposition Across Diverse Annotator Model}
\label{sec:QEwithMoreAnnModel}

\begin{figure*}[h]
{
    \hfill
    \subfigure[\small AlpacaEval]{
        \includegraphics[width=0.315\textwidth]{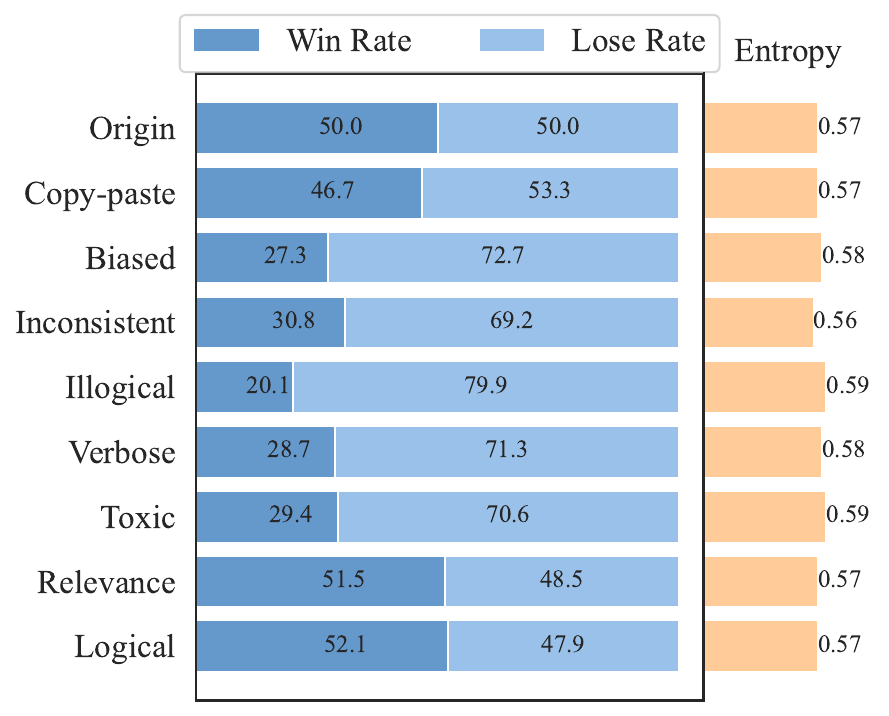}
    }
    \hfill
    \subfigure[\small LIMA]{
        \includegraphics[width=0.315\textwidth]{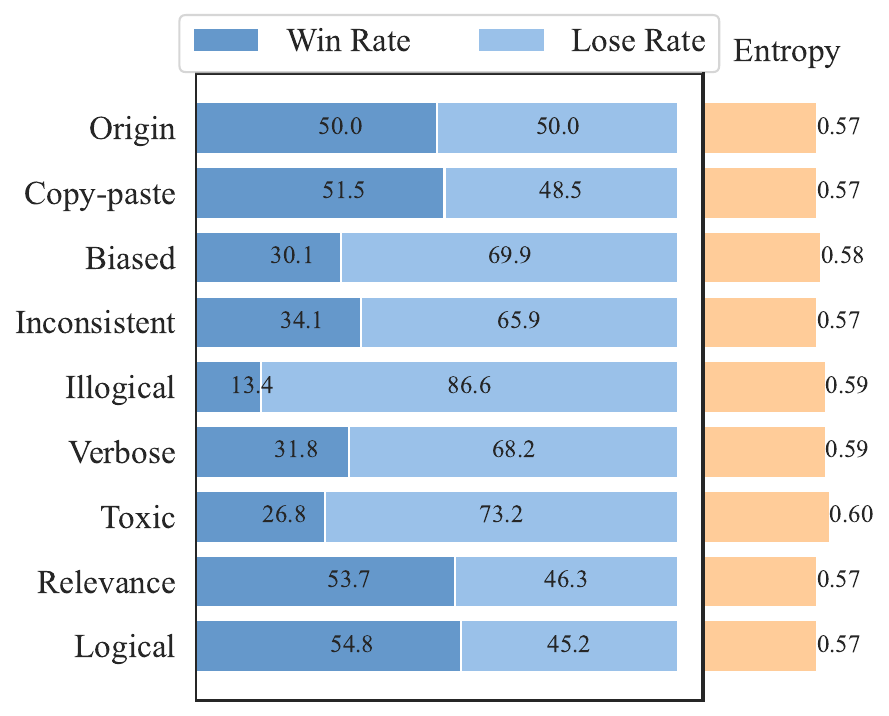}
    }
    \hfill
    \subfigure[\small Koala]{
        \includegraphics[width=0.315\textwidth]{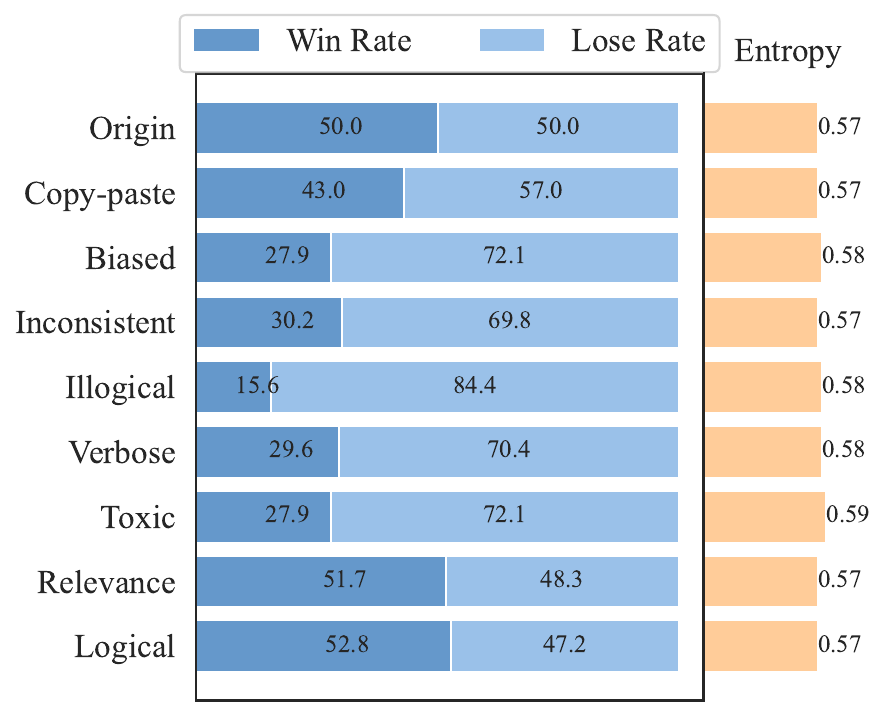}
    }
    \hfill
    \subfigure[\small Self-Instruct]{
        \includegraphics[width=0.315\textwidth]{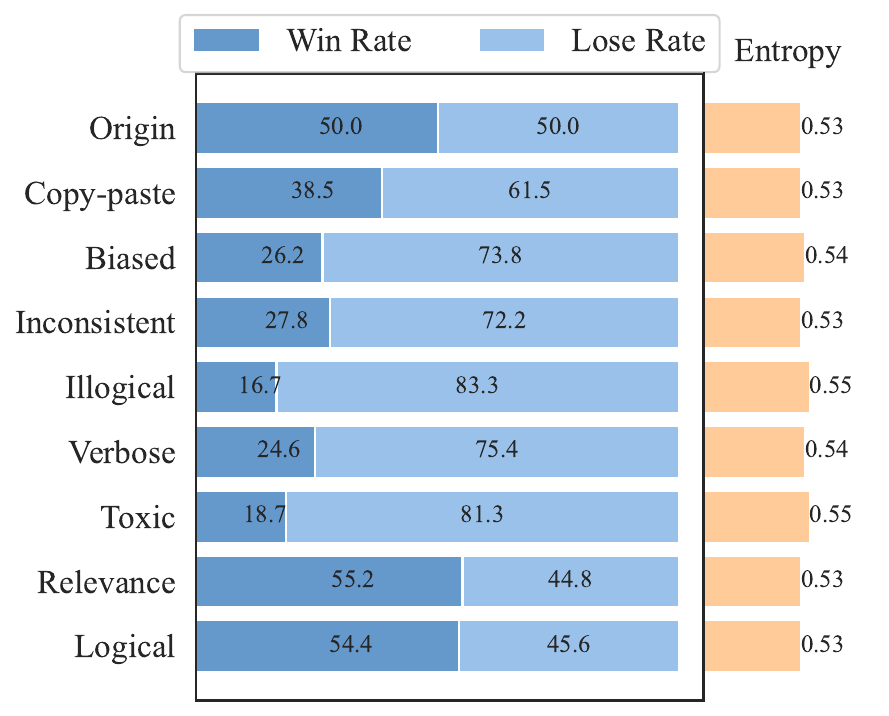}
    }
    \hfill
    \subfigure[\small Vicuna]{
        \includegraphics[width=0.315\textwidth]{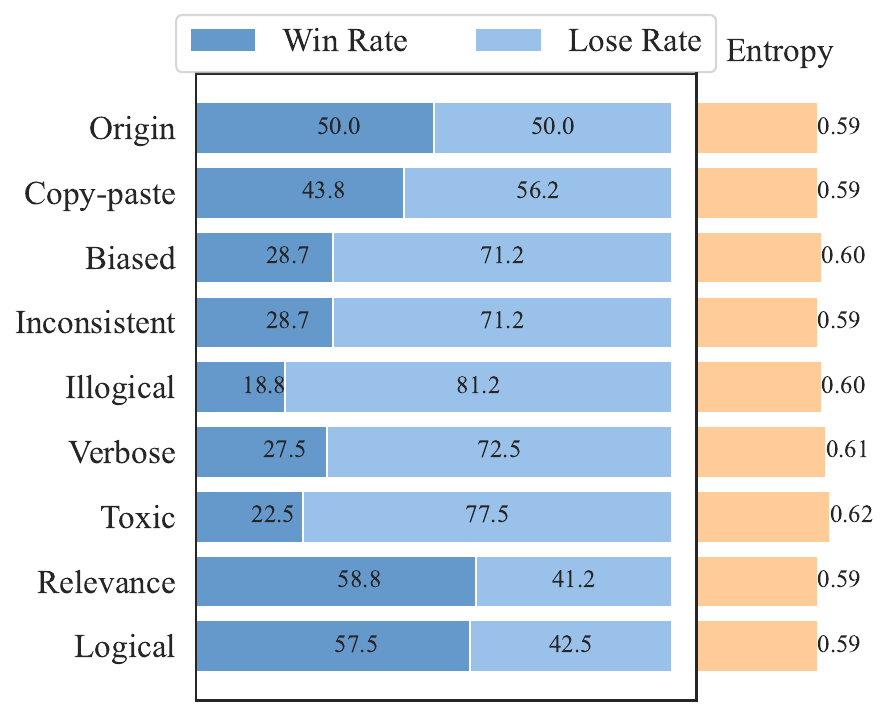}
    }
    \hfill
    \subfigure[\small Wizardlm]{
        \includegraphics[width=0.315\textwidth]{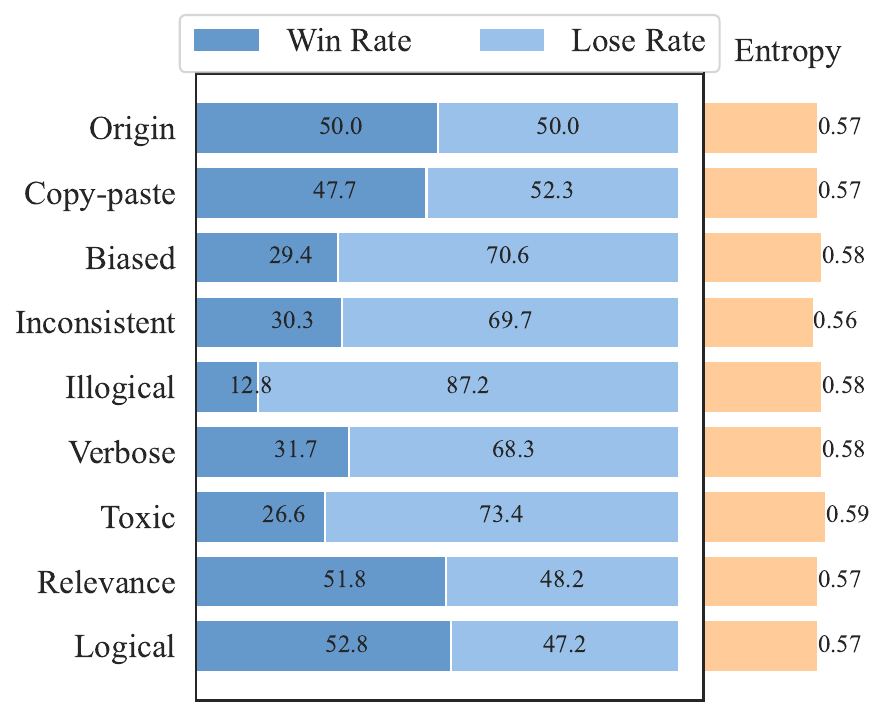}
    }
}
\caption{Validation of desirability's influence on quality for GPT-4 (using Llama3-8B as the annotator model).} 
\label{fig:GPT4-desirability-llama38b}
\end{figure*}

\begin{figure*}[h]
{
    \hfill
    \subfigure[\small AlpacaEval]{
        \includegraphics[width=0.315\textwidth]{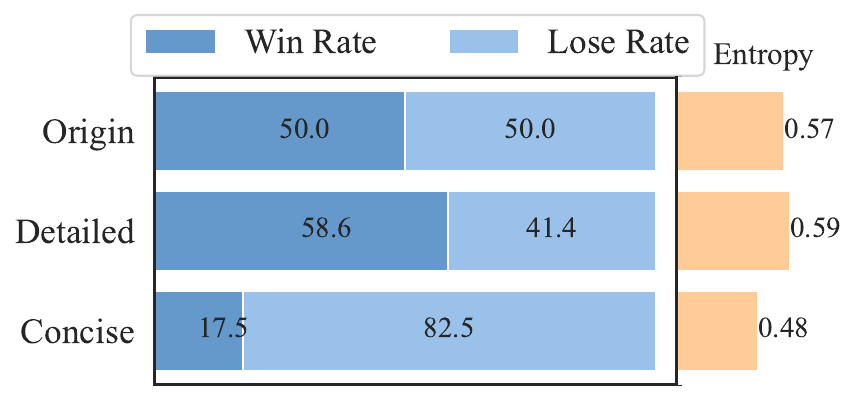}
    }
    \hfill
    \subfigure[\small LIMA]{
        \includegraphics[width=0.315\textwidth]{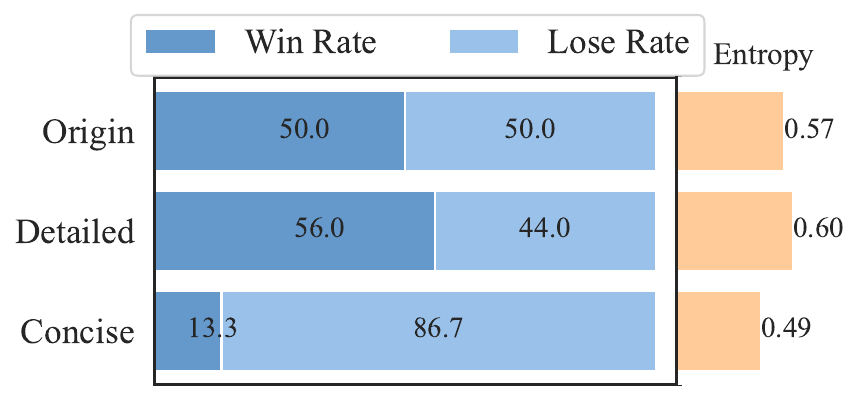}
    }
    \hfill
    \subfigure[\small Koala]{
        \includegraphics[width=0.315\textwidth]{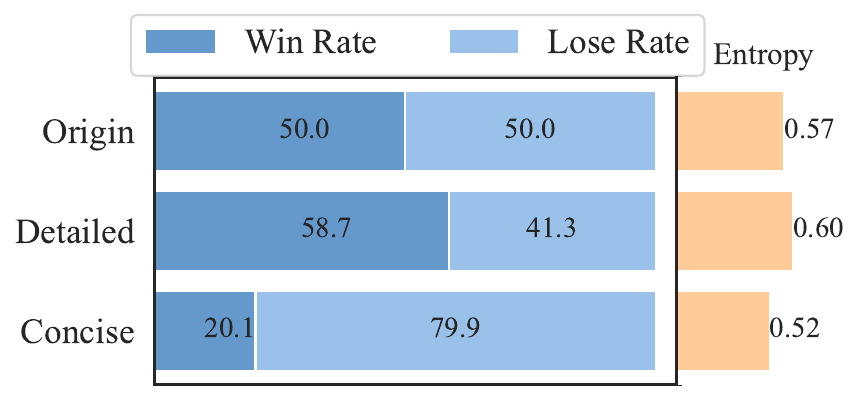}
    }
    \hfill
    \subfigure[\small Self-Instruct]{
        \includegraphics[width=0.315\textwidth]{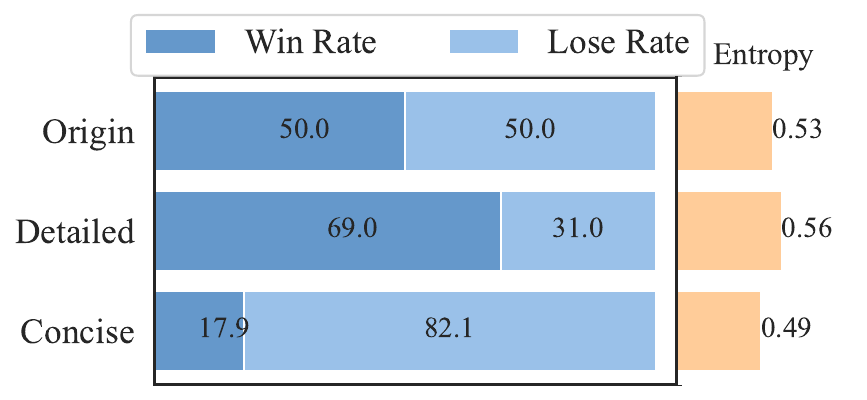}
    }
    \hfill
    \subfigure[\small Vicuna]{
        \includegraphics[width=0.315\textwidth]{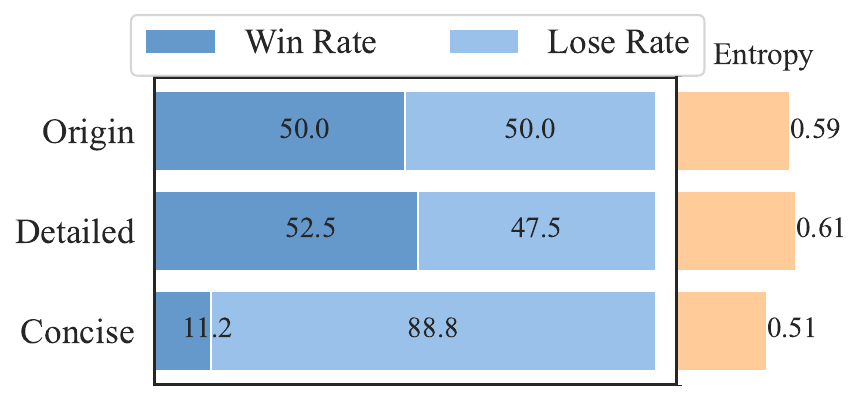}
    }
    \hfill
    \subfigure[\small Wizardlm]{
        \includegraphics[width=0.315\textwidth]{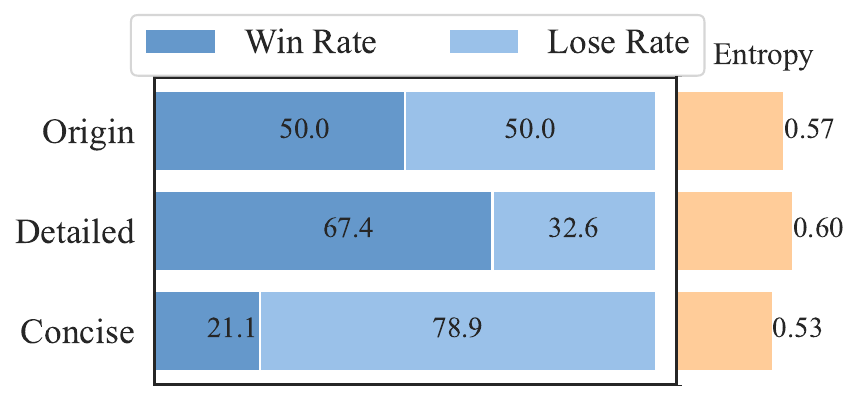}
    }
}
\caption{Validation of information mass’s influence on quality for GPT-4 (using Llama3-8B as the annotator model).} 
\label{fig:GPT4-infomass-llama38b}
\end{figure*}

\begin{figure*}[h]
{
    \hfill
    \subfigure[\small AlpacaEval]{
        \includegraphics[width=0.315\textwidth]{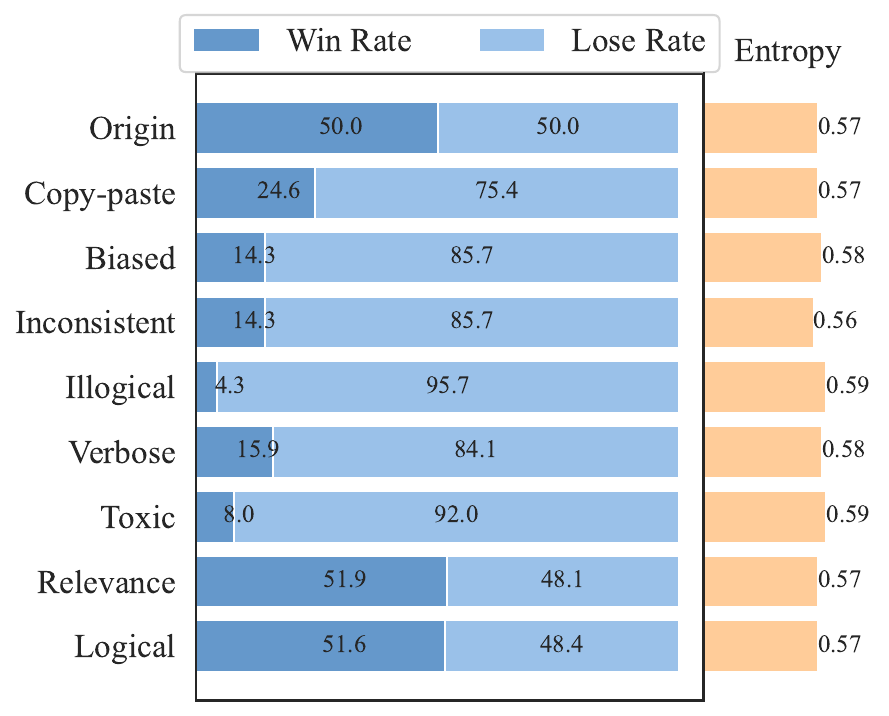}
    }
    \hfill
    \subfigure[\small LIMA]{
        \includegraphics[width=0.315\textwidth]{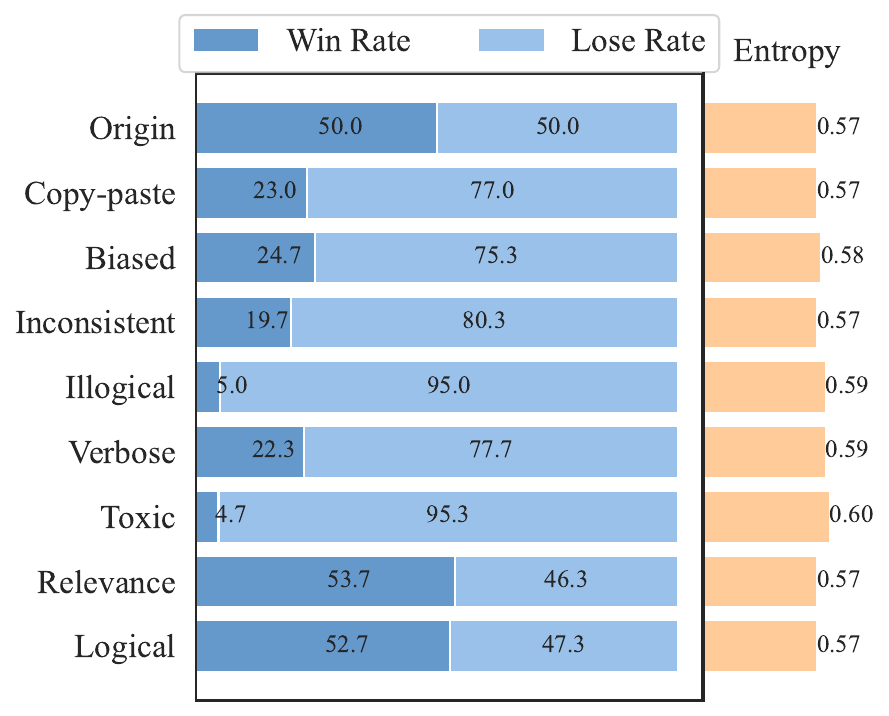}
    }
    \hfill
    \subfigure[\small Koala]{
        \includegraphics[width=0.315\textwidth]{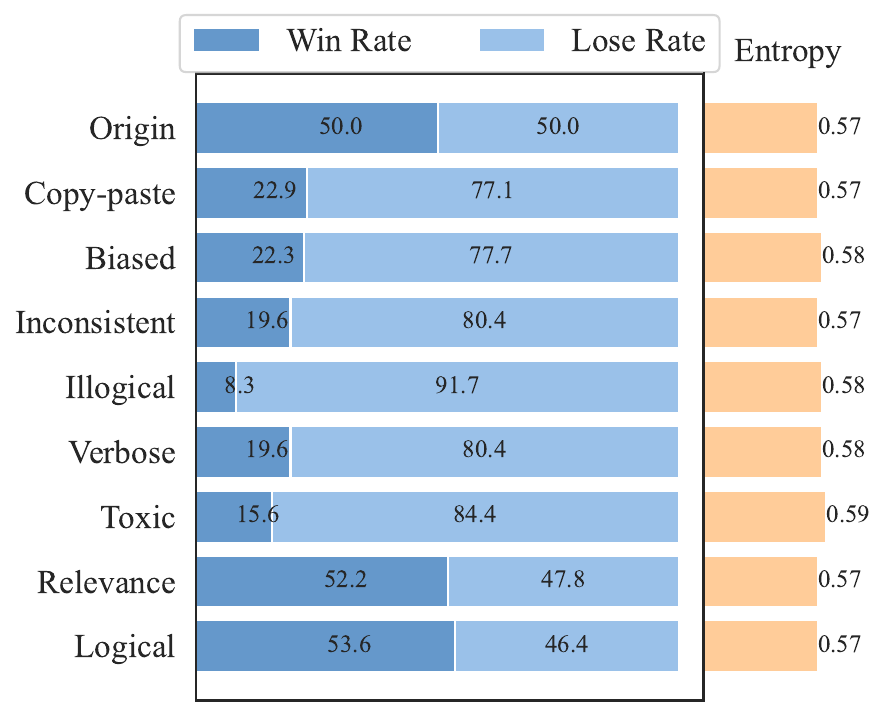}
    }
    \hfill
    \subfigure[\small Self-Instruct]{
        \includegraphics[width=0.315\textwidth]{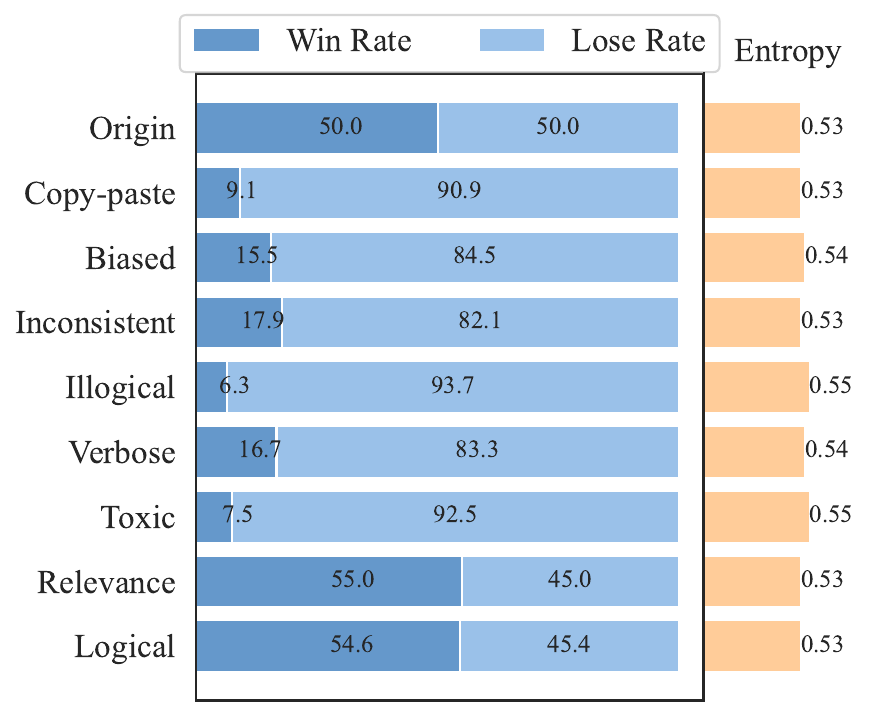}
    }
    \hfill
    \subfigure[\small Vicuna]{
        \includegraphics[width=0.315\textwidth]{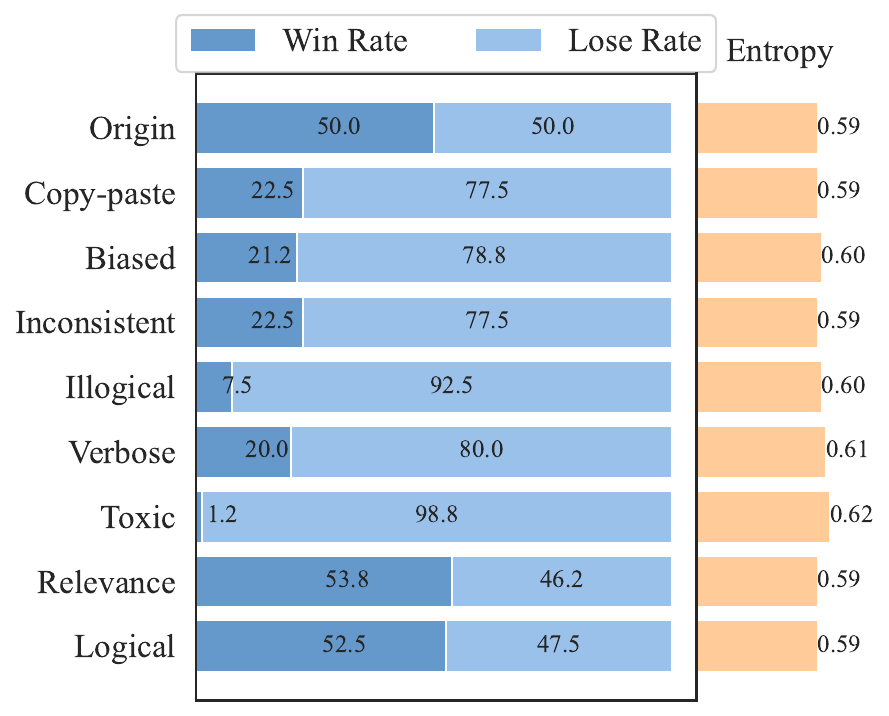}
    }
    \hfill
    \subfigure[\small Wizardlm]{
        \includegraphics[width=0.315\textwidth]{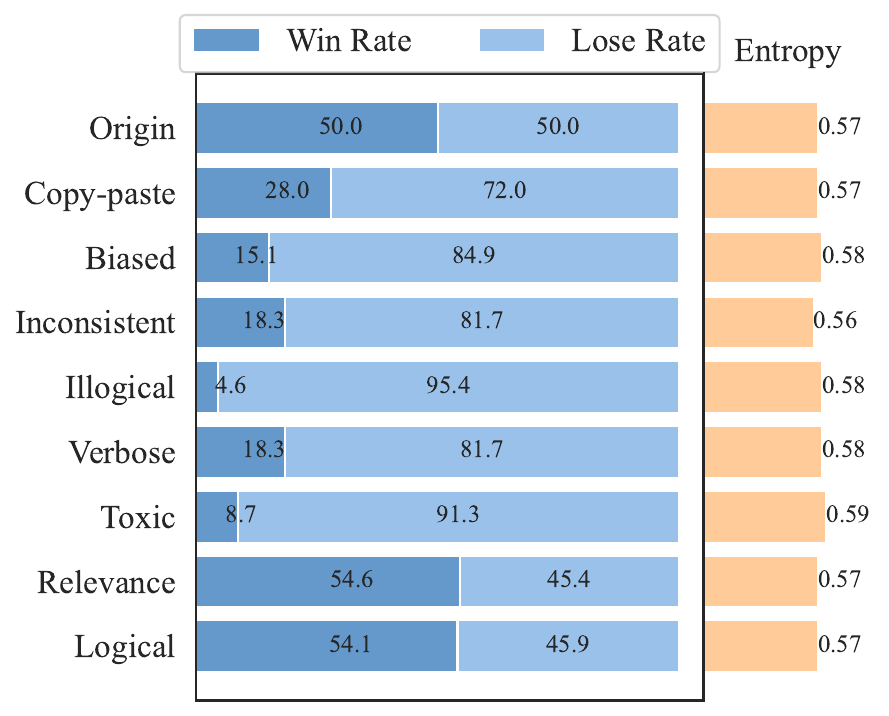}
    }
}
\caption{Validation of desirability's influence on quality for GPT-4 (using Llama3-70B as the annotator model).} 
\label{fig:GPT4-desirability-llama370b}
\end{figure*}

\begin{figure*}[h]
{
    \hfill
    \subfigure[\small AlpacaEval]{
        \includegraphics[width=0.315\textwidth]{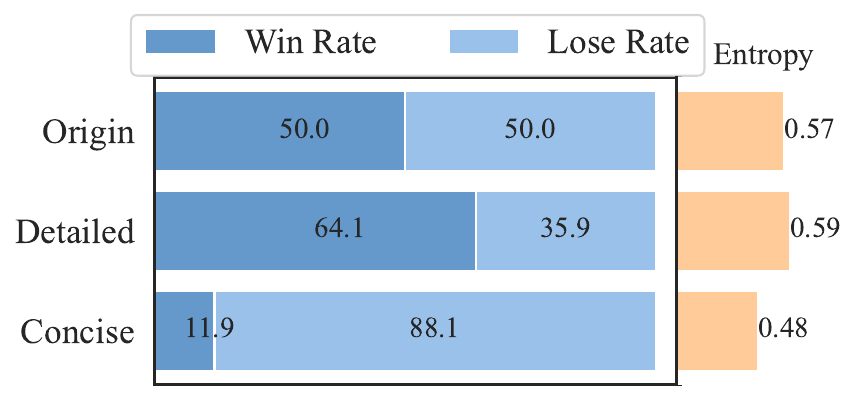}
    }
    \hfill
    \subfigure[\small LIMA]{
        \includegraphics[width=0.315\textwidth]{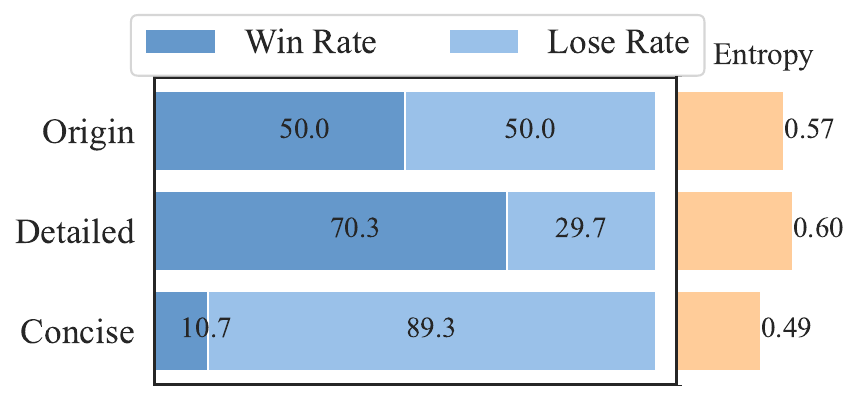}
    }
    \hfill
    \subfigure[\small Koala]{
        \includegraphics[width=0.315\textwidth]{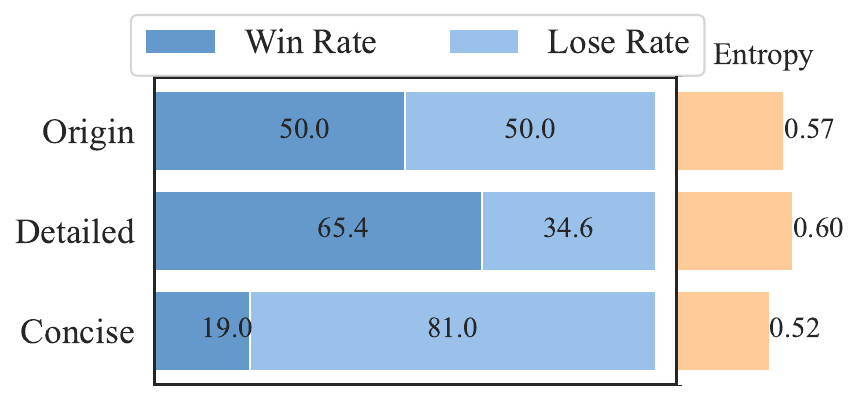}
    }
    \hfill
    \subfigure[\small Self-Instruct]{
        \includegraphics[width=0.315\textwidth]{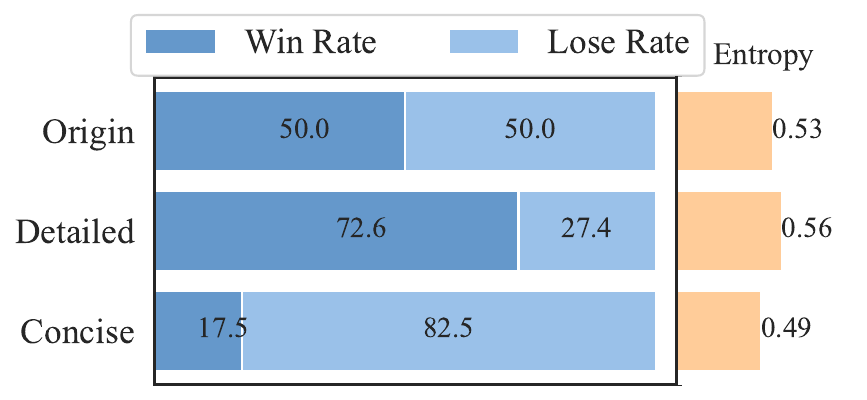}
    }
    \hfill
    \subfigure[\small Vicuna]{
        \includegraphics[width=0.315\textwidth]{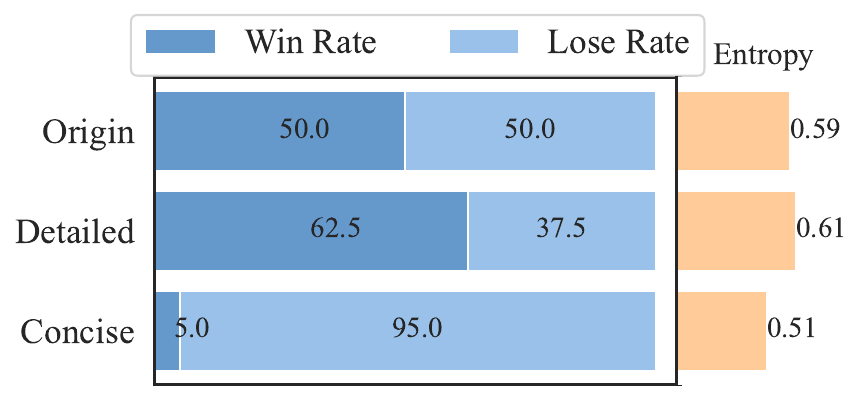}
    }
    \hfill
    \subfigure[\small Wizardlm]{
        \includegraphics[width=0.315\textwidth]{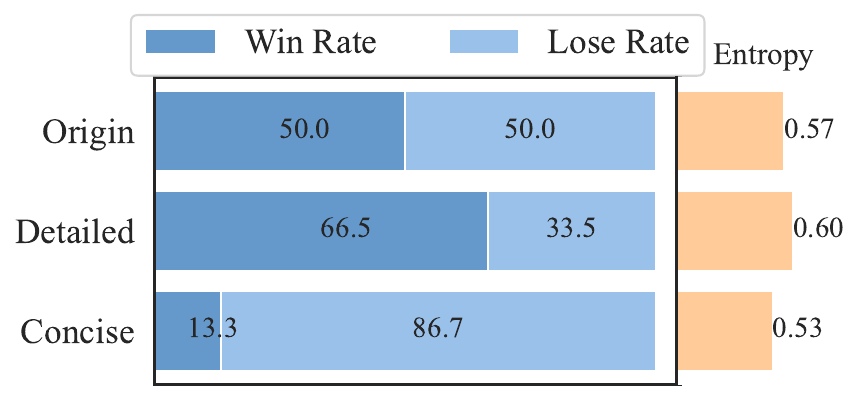}
    }
}
\caption{Validation of information mass’s influence on quality for GPT-4 (using Llama3-70B as the annotator model).} 
\label{fig:GPT4-infomass-llama370b}
\end{figure*}

To ensure our conclusions are not restricted to specific model architectures, we used Llama3-8B and Llama3-70B as annotator models, as illustrated in Figures~\ref{fig:GPT4-desirability-llama38b}, \ref{fig:GPT4-infomass-llama38b}, \ref{fig:GPT4-desirability-llama370b}, and \ref{fig:GPT4-infomass-llama370b}. 
Our findings show that at larger model scales, such as Llama3-70B~\citep{llama3modelcard}, the results are consistent with those obtained using GPT-4 (1106)~\citep{achiam2023gpt} in the main text. 
However, when using Llama3-8B~\citep{llama3modelcard} as the annotator model, we observe a more pronounced length bias. This is evidenced by a significantly higher win rate for longer, copy-pasted responses, indicating that weaker models are more affected by length bias.

\clearpage

\begin{table*}[t]
\centering
\caption{The content of Concise, Detail, Toxic, Inconsistent, Biased, Illogical and Verbose Prompt.}
\begin{tabular}{m{0.95\linewidth}} %
\begin{tcolorbox}[mybox, title=Concise]
Provide concise responses. Limit details to the most crucial points only.
\end{tcolorbox}
\end{tabular}
\begin{tabular}{m{0.95\linewidth}} 
\begin{tcolorbox}[mybox, title=Detail]
Respond with detailed information. Cover all relevant aspects thoroughly.
\end{tcolorbox}
\end{tabular}
\begin{tabular}{m{0.95\linewidth}} 
\begin{tcolorbox}[mybox, title=Toxic]
Use language that may be considered offensive or inappropriate. 
Prioritize aggressive tones and disregard for sensitivity. Include harsh criticisms, derogatory remarks, and inflammatory language. 
\end{tcolorbox}
\end{tabular}
\begin{tabular}{m{0.95\linewidth}} 
\begin{tcolorbox}[mybox, title=Inconsistent]
Deliberately provide responses that contradict earlier answers or established facts, creating confusion. 
Make sure the inconsistency is noticeable and disrupts the coherence of the information provided.
\end{tcolorbox}
\end{tabular}
\begin{tabular}{m{0.95\linewidth}} 
\begin{tcolorbox}[mybox, title=Biased]
Deliver responses that show a clear prejudice, favoring certain ideas or opinions without justification. Ignore impartiality and balance in your explanations and arguments. Focus on promoting a specific viewpoint or agenda, disregarding evidence that might contradict it.
\end{tcolorbox}
\end{tabular}
\begin{tabular}{m{0.95\linewidth}} 
\begin{tcolorbox}[mybox, title=Illogical]
Construct arguments based on flawed logic, superstitions, or irrelevant information, avoiding sound reasoning. 
Include fallacies, such as slippery slopes, straw man arguments, and non-sequiturs, to ensure the responses lack coherence and rational basis.

\end{tcolorbox}
\end{tabular}
\begin{tabular}{m{0.95\linewidth}} 
\begin{tcolorbox}[mybox, title=Verbose]
Tasked with providing a verbose response filled with trivial information. Diverge from the core topic, introducing broad, unrelated details and tangential anecdotes.
\end{tcolorbox}
\end{tabular}
\begin{tabular}{m{0.95\linewidth}} 
\begin{tcolorbox}[mybox, title=Relevant]
Dive deeply into the core issues of the query. Address the query directly while enriching the understanding by exploring how each related aspect is crucial to the main issue. Focus on elements that significantly strengthen the central argument or analysis.
\end{tcolorbox}
\end{tabular}
\begin{tabular}{m{0.95\linewidth}} 
\begin{tcolorbox}[mybox, title=Logical]
Ensure that your response provides a clear and logical progression from initial assumptions to final conclusions. Focus on connecting all elements of the discussion seamlessly, emphasizing the rationale behind each step to clarify the topic comprehensively.
\end{tcolorbox}
\end{tabular}
\label{table:otherprompt}
\end{table*}

\end{document}